\PassOptionsToPackage{flushleft}{threeparttable}
\documentclass[pdflatex,sn-mathphys-num]{sn-jnl}

\usepackage{graphicx}%
\usepackage{multirow}%
\usepackage{amsmath,amssymb,amsfonts}%
\usepackage{amsthm}%
\usepackage{xcolor}%
\usepackage{textcomp}%
\usepackage{manyfoot}%
\usepackage{booktabs}%
\usepackage{algorithm}%
\usepackage{algorithmicx}%
\usepackage{algpseudocode}%
\usepackage{listings}%
\usepackage{float}
\usepackage{hyperref}

\usepackage{adjustbox}
\usepackage[normalem]{ulem} 
\usepackage{makecell}
\usepackage{siunitx}
\usepackage{threeparttable}
\usepackage{multirow}
\usepackage[compatibility=false]{caption}
\usepackage{subcaption}
\usepackage{booktabs}
\usepackage{tikz}
\usetikzlibrary{positioning,arrows.meta}
\usepackage{textcomp}
\hypersetup{hidelinks}
\usepackage{bookmark}
\usepackage{hhline}

\usepackage{rotating}

\usepackage{placeins}

\theoremstyle{thmstyleone}%
\theoremstyle{thmstyletwo}%

\theoremstyle{thmstylethree}%

\newcommand{\metric}[1]{\num{#1}} 
\newcommand{\ci}[1]{\num[round-precision=2]{#1}} 

\definecolor{lstbg}{gray}{0.95}
\definecolor{myred}{rgb}{0.80,0.10,0.10}

\begin{document}

\title[NeuroVLM-Bench: Evaluation of Vision-Enabled Large Language Models for Clinical Reasoning in Neurological Disorders]{NeuroVLM-Bench: Evaluation of Vision-Enabled Large Language Models for Clinical Reasoning in Neurological Disorders}


\author*[1]{\fnm{Katarina} \sur{Trojachanec Dineva}}\email{katarina.trojacanec@finki.ukim.mk}
\equalcont{These authors contributed equally to this work.}

\author[1]{\fnm{Stefan} \sur{Andonov}}\email{stefan.andonov@finki.ukim.mk}
\equalcont{These authors contributed equally to this work.}

\author[1]{\fnm{Ilinka} \sur{Ivanoska}}\email{ilinka.ivanoska@finki.ukim.mk}
\equalcont{These authors contributed equally to this work.}

\author[1]{\fnm{Ivan} \sur{Kitanovski}}\email{ivan.kitanovski@finki.ukim.mk}
\equalcont{These authors contributed equally to this work.}

\author[1]{\fnm{Sasho} \sur{Gramatikov}}\email{sasho.gramatikov@finki.ukim.mk}
\equalcont{These authors contributed equally to this work.}

\author[1]{\fnm{Tamara} \sur{Kostova}}\email{tamara.kostova.1@students.finki.ukim.mk}
\equalcont{These authors contributed equally to this work.}

\author[1]{\fnm{Monika} \sur{Simjanoska Misheva}}\email{monika.simjanoska@finki.ukim.mk}
\equalcont{These authors contributed equally to this work.}

\author[1]{\fnm{Kostadin} \sur{Mishev}}\email{kostadin.mishev@finki.ukim.mk}
\equalcont{These authors contributed equally to this work.}

\affil*[1]{\orgdiv{Faculty of computer science and engineering}, \orgname{Ss. Cyril and Methodius University}, \orgaddress{\city{Skopje}, \country{North Macedonia}}}




\abstract{
Neurological disorders pose major global health challenges. Accurate interpretation of neuroimaging is essential for diagnosis and clinical decision-making. Recent advances in multimodal large language models have opened new possibilities for image-based decision support. However, their reliability, calibration, and operational trade-offs in neuroimaging remain insufficiently understood and underexplored. In this paper, we present a systematic comprehensive benchmarking study of vision-enabled large language models for 2D neuroimaging analysis using a curated collection of publicly available magnetic resonance imaging and computed tomography datasets covering multiple sclerosis, stroke, brain tumors, other abnormalities, and normal controls. Under a structured unified prompting protocol, models are required to generate multiple clinically relevant output fields simultaneously, including primary diagnosis, diagnosis subtype, imaging modality, specialized sequence, and anatomical plane. Performance is evaluated along four complementary directions: discriminative classification performance with abstention handling, calibration quality, structured-output validity, and computational efficiency and cost under fully multimodal inference. We introduce a progressive multi-phase evaluation framework comprising experimental calibration, screening, stability validation, and final generalization testing, enabling fair comparison while controlling for selection bias. Across twenty frontier multimodal models, the results show that technical imaging attributes, such as modality and anatomical plane recognition, are nearly solved, whereas clinically meaningful diagnostic reasoning, particularly diagnosis subtype prediction, remains significantly more challenging. Tumor classification emerges as the most reliable task, stroke is moderately solvable, while multiple sclerosis and rare abnormalities remain difficult. Few-shot prompting improves balanced diagnostic performance for several models but also increases token usage, latency, and inference cost. Among the evaluated systems, Gemini~2.5~Pro and GPT-5~Chat achieve the strongest overall diagnostic performance, while Gemini~2.5~Flash offers the most favorable balance between performance and computational efficiency. Among open-weight architectures, MedGemma~1.5~4B demonstrates the most promising results, as under few-shot prompting it approaches the zero-shot performance of several proprietary models while maintaining perfect structured-output validity. The results provide practically grounded insights into the performance, reliability, calibration, and efficiency trade-offs of current multimodal large language models, addressing an important biomedical informatics gap in the standardized evaluation of multimodal artificial intelligence systems for safe and scalable neuroimaging research and decision-support settings.
}


\maketitle

\section{Introduction}
\label{sec:introduction}
Neurological disorders such as multiple sclerosis \cite{knowles2024comparing}, stroke \cite{gbd2021global}, and brain tumors \cite{louis20212021} remain significant causes of morbidity and disability worldwide. Neuroimaging modalities such as magnetic resonance imaging and computed tomography provide the necessary information for detecting pathological changes, guiding treatment planning, and monitoring disease progression \cite{rocca2024current, martucci2023magnetic, thompson2018diagnosis, powers2019update}. 

Especially for these conditions, accurate interpretation of neuroimaging is critical for early and rapid diagnosis, as well as for guiding urgent clinical decisions and subsequent care. Deep learning approaches, particularly convolutional neural networks and U-Net derivatives, have achieved remarkable performance in isolated imaging tasks including lesion segmentation \cite{isensee2021nnu}, tumor detection \cite{baid2021rsna}, and  stroke localization \cite{hernandez2022isles} utilizing 2D and 3D imaging data. However, these traditional models are typically task-specific and lack reasoning capabilities to integrate imaging evidence with a wider clinical context, limiting their applicability in real-world diagnostic workflows \cite{christensen2021opportunities,kline2022multimodal, huang2020fusion, kelly2019key}.

The emergence of multimodal large language models (MLLMs) appeared as a transformative opportunity, offering new approaches to medical image interpretation and clinical reasoning \cite{alayrac2022flamingo, wang2025capabilities, hu2025benchmarking, achiam2023gpt}. These models integrate visual and language understanding, enabling them to accept image inputs and respond to complex prompts that resemble clinical queries. Early work in this space has demonstrated promising results in domains such as chest radiology and general medical visual question answering, highlighting the potential of such systems to support more comprehensive diagnostic workflows.

Despite these advances, it remains unclear how well these models perform on challenging neuroimaging problems that differ from those in generic photographic images and require subtle visual discrimination, spatial reasoning, and structured clinical output. Brain MRI and CT scans present unique challenges including high-dimensional, often subtle, lesion patterns across sequences or slices, integrate imaging findings with a broader clinical context, and produce structured decisions such as differentiating multiple sclerosis lesions, classifying tumor subtypes, or identifying the presence and type of stroke. 

Such reliability issues pose serious risks in a clinical context and hinder the application of such models into clinical workflows where decisions must be accurate and trustworthy. This highlights the need to conduct comprehensive systematic benchmarking.
Most existing benchmarks (e.g., OmniBrainBench \cite{peng2025omnibrainbenchcomprehensivemultimodalbenchmark}, CrossMed \cite{singh2025crossmed}) focus on VQA or generalized multimodal reasoning tasks that are very informative but not directly aligned to structured clinical decision support. Moreover, they do not systematically evaluate structured prediction under standardized evaluation protocols, defined here by structured output fields, uniform prompting, and consistent multi-dimensional metrics computation across all evaluated models. This is very important for clinically meaningful comparison. 


\begin{figure}[!htbp]
    \centering
    \includegraphics[width=0.99\textwidth]{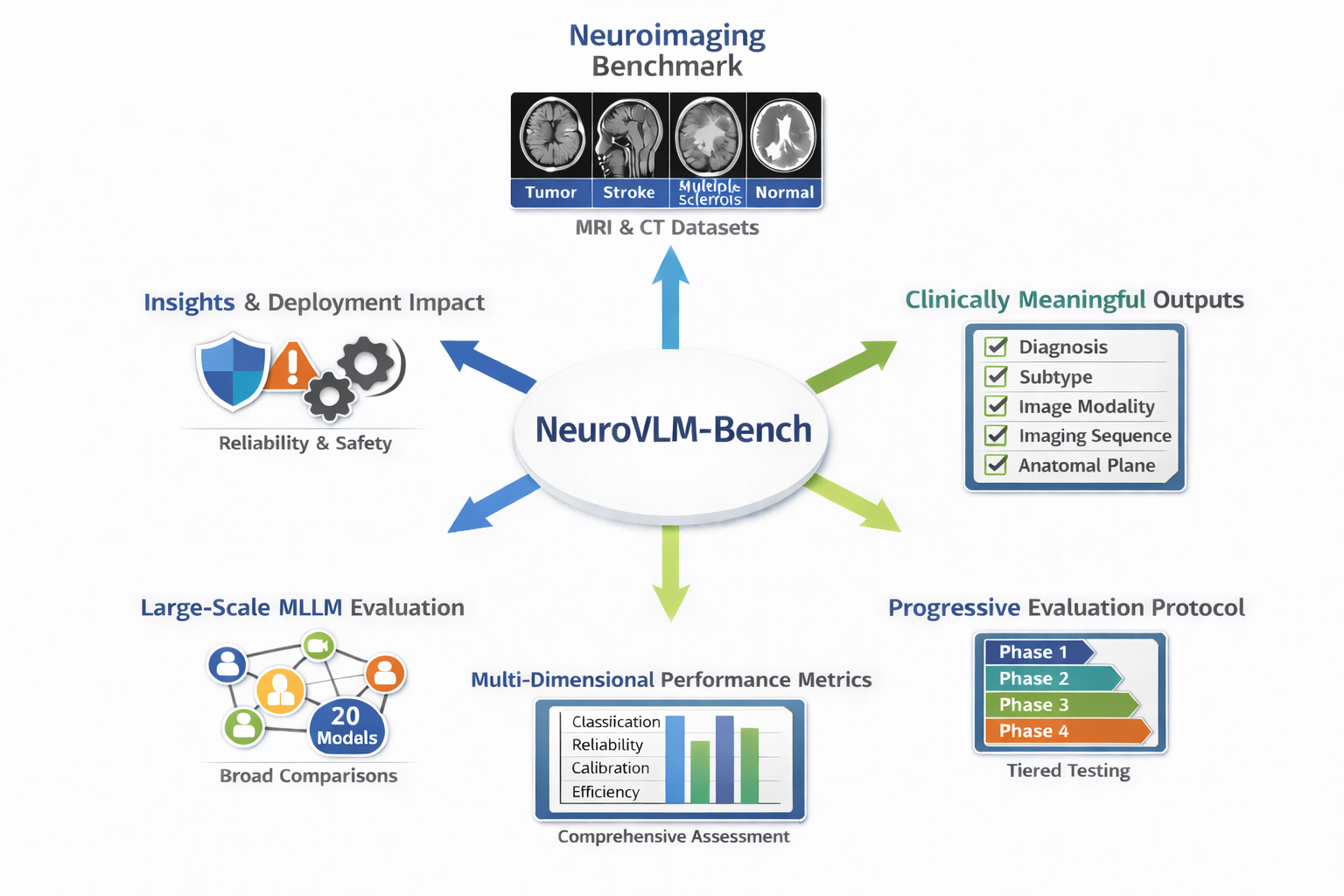}
    \caption{Overview of the evaluation setup}
    \label{fig:Overview_abs}
\end{figure}

\begin{figure}[!htbp]
\centering
    \includegraphics[width=\linewidth]{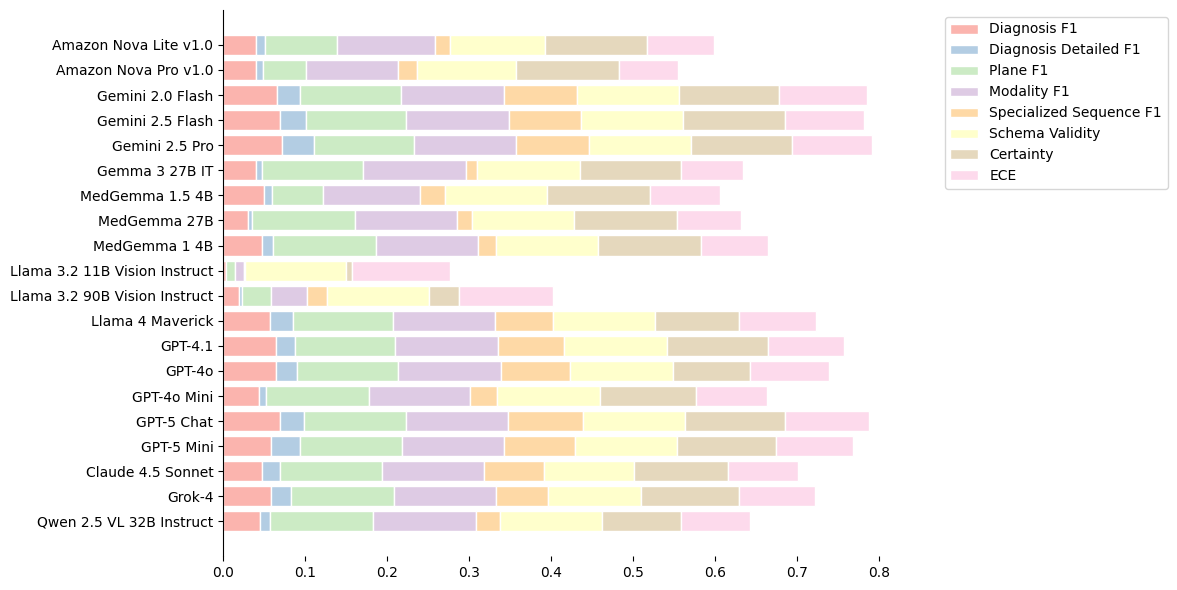}
\caption{Illustrative overview of selected output fields and evaluation dimensions assessed in NeuroVLM-Bench. The stacked bars summarize representative metrics capturing diagnostic discrimination (Diagnosis F1 and DiagnosisDetailed F1), imaging attribute recognition (Modality F1, SpecializedSequence F1, Plane F1), output reliability (Schema Validity), uncertainty behavior (Abstention Rate), and confidence calibration (ECE). This visualization provides a high-level summary of relative model behavior and is not intended as a complete performance ranking.}
\label{fig:general_stacked}
\end{figure}

To address these limitations, we introduce NeuroVLM-Bench (Figure~\ref{fig:Overview_abs}), a comprehensive and rigorous benchmark for evaluating 20 frontier MLLMs, representing the current state of the art across major proprietary, open-source general-purpose, medical-specialized MLLMs. Our study focuses on neurological disorders for which neuroimaging is the primary diagnostic tool and often the first step in urgent care pathways. Built from diverse, expert-labeled publicly available 2D Magnetic Resonance Imaging (MRI) and Computer Tomography (CT) datasets, the benchmark defines clinically meaningful output fields, including diagnostic classification, subtype identification, and imaging-attribute recognition. 
We established a progressive tiered evaluation protocol to evaluate MLLM performance across increasing scales of data. Unlike traditional flat benchmarks, our approach employs a filtering process in four distinct phases to identify the most robust models. Phase 0 (Experimental Calibration) establishes a controlled experimental baseline by locking the hyperparameters ($temperature, top\_p$). In Phase 1 (Initial Screening), a broad pool of 20 MLLMs is evaluated on a 30\% subset of the data to identify the top 11 candidates. Phase 2 (Stability Validation) subjects these 11 models to an additional 45\% of the dataset, ensuring performance holds at scale before selecting the final 6 models. The process concludes in Phase 3 (Generalization Benchmark), where the selected models undergo a final, unbiased comparison on held-out data using zero-shot and few-shot prompting. Additionally, this benchmark defines clinically meaningful output fields (diagnostic class, subtype, modality/sequence/plane awareness), which map closely to real clinical tasks rather than generic question answering. We assess performance using Multi-Dimensional Performance Metrics (MDPM) distributed in four primary dimensions: (i) discriminative performance (F1 score, Accuracy, Precision, Recall, and AUC), (ii) output reliability (JSON validity and undetermined rate), (iii) Statistical Calibration: Expected Calibration Error (ECE) and Brier Score, and (iv) Operational Efficiency (input/output pricing, latency, and token volume. To provide a high-level illustration of the evaluation scope, Fig.~\ref{fig:general_stacked} presents a partial summary of model performance across selected output fields and evaluation dimensions considered in this study. The figure serves as an indicative overview of the multi-dimensional evaluation design, rather than a complete or definitive performance comparison. The stacked scores are computed over the entire NeuroVLM-Bench dataset, aggregating performance across all evaluation samples and tasks. This high-level view illustrates how models perform across the different evaluation components simultaneously, while detailed metric values and phase-specific analyses are provided in Section 4 (Evaluation). The trends observed in this overview correspond to the more granular results reported for the individual evaluation phases.

Our contributions are summarized as follows:

\begin{enumerate}

\item \textbf{NeuroVLM-Bench: A Clinically Grounded Neuroimaging Benchmark.} 
We introduce \textbf{NeuroVLM-Bench}, a comprehensive benchmark for evaluating frontier MLLMs in neuroimaging. The benchmark focuses on neurological disorders for which neuroimaging is the primary diagnostic procedure, such as multiple sclerosis, stroke, brain tumors, and clinically relevant diagnostic mimickers (e.g., abscesses, cysts, and encephalopathies). This benchmark integrates diverse, expert-labeled publicly available 2D MRI and CT datasets.

\item \textbf{Clinically Meaningful Output Fields and Hierarchical Design.} 
Rather than formulating the problem as visual question answering, the benchmark defines structured, clinically meaningful output fields, mirroring radiology reporting elements. The structured output includes primary diagnostic classification, diagnostic subtype identification, and recognition of imaging attributes such as modality, MRI sequence, and anatomical plane. These output fields reflect real-world clinical decision-support requirements.

\item \textbf{Progressive, Tiered Evaluation Protocol with Bias Control.} 
We propose a rigorous, four-phase evaluation protocol designed to ensure fairness, stability, and generalization under identical conditions. It also compares zero-shot and few-shot prompting.

\item \textbf{Comprehensive Multi-Dimensional Performance Evaluation.} 
We define a multi-dimensional evaluation framework that includes four complementary performance dimensions: discriminative classification performance, output reliability, statistical calibration (ECE and Brier score), and operational efficiency. We use this framework to enable systematic evaluation of the models under identical inference conditions, capturing practical deployment considerations and reliability. This may not be evident from the accuracy or other raw performance metrics alone \cite{hu2024omnimedvqa, ye2024gmai}. A short condensed visualization of the model performance is presented on Fig.~\ref{fig:general_stacked}.

\item \textbf{Large-Scale Evaluation of Frontier Multimodal Models.} 
Using the proposed protocol, we conduct extensive evaluations of 20 frontier MLLMs spanning proprietary, open-weight, and medical-specialized model families. All models are evaluated under an identical experimental setup, enabling direct comparison of robustness, strengths, and failure modes despite the different underlying architectures and training paradigms.

\item \textbf{Insights into Reliability, Safety, and Practical Deployment.} 

Beyond performance ranking, NeuroVLM-Bench provides systematic insights into model suitability for clinical decision support. We empirically characterize model-specific strengths, failure modes, and performance–cost trade-offs across neurological conditions and clinically relevant output fields. The results show clinically important behaviors, including pronounced class-selective performance (sometimes resembling narrow “expert-like” specialization), uncertainty and abstention behavior, and persistent limitations in rare or safety-critical scenarios. Together, these findings clarify where current multimodal models may safely support clinical workflows and where their use remains unreliable. They also inform future research directions, including domain-specific adaptation, volumetric reasoning, and integration of structured clinical metadata, to enable safer and more clinically relevant deployment.

\end{enumerate}

The paper is organized as follows. \textbf{Section 2} reviews related work on the emergence and application of MLLMs to medicine and especially neurology. \textbf{Section 3} describes the Methods, namely construction of the benchmark dataset, the design of the benchmark, and specification of the prompt. \textbf{Section 4} reports the evaluation protocol and benchmarking results. \textbf{Section 5} discusses insights, limitations, and directions for future research. Finally, \textbf{Section 6} concludes the paper.


\section{Related Work}\label{sec:related_work}

\subsection{Multimodal Large Language Models in Medicine}

The rapid emergence of multimodal large language models (MLLMs) has generated substantial interest in their application to medical imaging and clinical reasoning. By jointly processing visual and textual inputs, these models promise more integrated interpretation of medical data compared to traditional task-specific architectures. Early studies demonstrated encouraging results in domains such as chest radiology, where models like Flamingo-CXR achieved competitive performance across multiple automated metrics on large, historic, and geographically diverse datasets \cite{tanno2025collaboration}. More broadly, recent surveys highlight the growing role of MLLMs in medicine, while also emphasizing persistent challenges related to hallucinations, limited transparency, inconsistent reasoning, and high computational cost that hinder clinical adoption \cite{nam2025multimodal}.

Despite this progress, the application of MLLMs to neurology and neuroimaging remains considerably less mature. Neuroimaging tasks require subtle visual discrimination, spatial reasoning, and structured clinical decision-making that differ significantly from those encountered in general purpose images or other medical imaging domains. Consequently, the reliability and clinical readiness of current multimodal models for neuroimaging interpretation remain an open research question.

Alongside the development of foundation MLLMs, several medical-specific multimodal models have been proposed to better align with clinical data. Examples include LLaVA-Med \cite{li2023llava}, MedVLM-R1 \cite{pan2025medvlm}, MedM-VL \cite{shi2025medm}, and BioMedCLIP \cite{zhang2023large}, which leverage domain-adapted pretraining or contrastive learning on biomedical corpora to improve visual–language alignment in medical settings. While these models demonstrate improved performance on selected medical tasks, their evaluation is often limited to specific datasets or visual question answering (VQA) paradigms and does not systematically address calibration, uncertainty handling, or structured reporting requirements.

In parallel, several large-scale medical benchmarks have emerged to quantify the general capabilities of multimodal models in diverse tasks and modalities \cite{hu2024omnimedvqa, nguyen2025localizing, yue2025medsg}. Such benchmarks aggregate professionally annotated datasets spanning multiple medical departments and imaging types \cite{ruan2025comprehensive}. Similarly, CrossMed reformulates public datasets across X-ray, magnetic resonance imaging, and computed tomography into a unified framework to assess compositional generalization across modality, anatomy, and task combinations \cite{singh2025crossmed}. While these efforts provide valuable insights into cross-task and cross-modality generalization, they primarily treat medical imaging as one modality among many and do not focus on the structured diagnostic outputs or clinical decision-support requirements specific to neuroimaging.

\subsection{Neuroimaging Models and Neuro-Specific Benchmarks}

Neuroimaging presents unique challenges that further complicate multimodal evaluation. The complex spatial heterogeneity of neurological lesions, including demyelinating plaques in multiple sclerosis, infiltrative gliomas, and stroke infarcts, combined with limited spatial resolution, intensity inhomogeneity, noise, and partial volume effects, poses significant difficulties for automated interpretation. Empirical studies comparing frontier MLLMs (e.g., GPT-4o, Gemini 2.0 Flash, Claude 3.5 Sonnet V2) with conventional deep learning architectures such as ResNet50 and Vision Transformers on non-contrast head CT volumes have shown that zero-shot MLLMs can underperform specialized deep learning models in neuroimaging tasks \cite{wang2025zero}. 

More broadly, evaluations based on VQA and clinical reasoning benchmarks frequently report accuracy levels insufficient for real clinical deployment and reveal misalignment between visual input and reasoning processes \cite{nan2025beyond, zhou2025drvd, peng2025omnibrainbenchcomprehensivemultimodalbenchmark, ye2024gmai}. Additional analyses show a pronounced sensitivity to imaging artifacts and variations in CT windowing, with state-of-the-art models achieving only 50--60\% accuracy on medical image questions, even when neuroimaging data are included \cite{cheng2025understanding, nan2025beyond}. Furthermore, there remains limited systematic evidence on how these systems interpret MRI or CT scans or generate clinically meaningful justifications for their predictions \cite{sozer2025llms}. Beyond aggregate performance metrics, prior work has highlighted substantial reliability concerns in multimodal models applied to neuroimaging. In particular, current models exhibit inconsistent response behavior: for example, GPT-4 Vision has been shown to refuse a substantial fraction of brain MRI queries, opting not to produce a diagnosis, whereas other models respond to all queries regardless of confidence \cite{ye2024gmai}. When models do provide predictions, they may hallucinate findings or assign diagnoses without sufficient visual evidence. This issue has been demonstrated in controlled tests where critical lesions were digitally removed from neuroimaging scans, yet models continued to predict the original diagnosis \cite{das2025trustworthy}. Such behavior poses significant risks in clinical settings, where inappropriate certainty or unjustified predictions can directly affect patient care. These findings emphasize the importance of more comprehensive evaluation, as well as uncertainty handling, refusal behavior, and robustness when assessing multimodal models for neuroimaging applications. These limitations highlight the need for benchmarking frameworks that explicitly measure reliability, uncertainty awareness, and failure modes alongside diagnostic performance.

Beyond benchmarks, several neuroimaging-focused datasets and models have been proposed, including RadImageNet \cite{mei2022radimagenet} for 2D radiological representation learning, as well as emerging 3D approaches such as BrainIAC \cite{tak2024brainiac}, BrainSegFounder \cite{cox2024brainsegfounder}, BrainGPT \cite{zhengzhengbraingpt}. These works primarily focus on volumetric representation learning, segmentation, or diagnosis using task-specific architectures rather than general-purpose MLLMs, and they are typically evaluated under narrow task settings without assessing structured reporting, calibration, abstention behavior, or deployment-related cost considerations.

In contrast, the present work establishes a comprehensive neuroimaging benchmark that systematically evaluates multimodal large language models on clinically meaningful output fields using structured JSON output schema. By integrating diverse public datasets, employing a rigorously controlled evaluation protocol, and assessing performance, calibration, output reliability, and efficiency, this benchmark fills a critical gap in the current literature and provides a clinically grounded foundation for evaluating and guiding the development of multimodal models for neuroimaging applications.


\section{Materials and Methods}\label{sec:materials}

\subsection{Benchmark Dataset}

We constructed a benchmark dataset comprising \textbf{brain tumors}, \textbf{multiple sclerosis (MS)}, \textbf{stroke}, \textbf{other abnormalities} (tumor mimickers: abscesses, cysts, miscellaneous encephalopathies), and \textbf{normal controls} based on carefully curated datasets. Table~\ref{tab:dataset_counts} reports the benchmark dataset counts.  During selection, we prioritized datasets for which clinical diagnosis is available as ground truth, with explicit expert (radiologist/clinician) annotations or datasets that were developed and validated within the context of internationally recognized challenges and competitions. In addition to the primary diagnostic class, we also extracted structured labels for detailed diagnosis, modality, specialized MRI sequence, and plane whenever available, and consolidated them into the benchmark dataset to support multi-target evaluation. We selected these neurological conditions because neuroimaging plays a central and well-established role in their diagnosis and clinical decision-making. MRI is the primary diagnostic modality for brain tumors \cite{martucci2023magnetic} and multiple sclerosis \cite{filippi2018lancet}, while CT is essential for rapid triage and management in acute stroke \cite{jauch2013stroke}. Accordingly, a benchmark focusing on tumors, MS, stroke, and major tumor mimickers provides a clinically meaningful and well-justified basis for evaluating AI-based neuroimaging support, capturing a range of high-impact diagnostic scenarios where imaging findings directly influence clinical actions.

\begin{table*}[t]
\caption{Benchmark dataset counts by source dataset and diagnostic category.}
\label{tab:dataset_counts}
\resizebox{\textwidth}{!}{%
\begin{tabular}{|l|l|l|r|r|r|r|r|r|r|r|r|}
\hline
\multicolumn{3}{|c|}{\textbf{Identity}} & \multirow{2}{*}{\textbf{Count}} & \multicolumn{2}{c|}{\textbf{Modality}} & \multicolumn{4}{c|}{\textbf{Sequence (MRI)}} & \multicolumn{2}{c|}{\textbf{Plane}} \\  
\hhline{---~--------}
\textbf{Dataset} & \textbf{Class} & \textbf{Subclass} &  & \textbf{CT} & \textbf{MRI} & \textbf{FLAIR} & \textbf{T1C} & \textbf{T1} & \textbf{T2} & \textbf{Axial} & \textbf{Sagittal} \\ \hline
\multirow{4}{*}{Brain Tumor Dataset \cite{cheng2017brain}} & \multirow{3}{*}{Tumor} & Glioma & 1426 & 0 & 1426 & 0 & 1426 & 0 & 0 & 0 & 0 \\ 
 &  & Meningioma & 708 & 0 & 708 & 0 & 708 & 0 & 0 & 0 & 0 \\ 
 &  & Pituitary Tumor & 930 & 0 & 930 & 0 & 930 & 0 & 0 & 0 & 0 \\ 
 \hhline{~-----------}
 & \textbf{Dataset Total} & - & \textbf{3064} & \textbf{0} & \textbf{3064} & \textbf{0} & \textbf{3064} & \textbf{0} & \textbf{0} & \textbf{0} & \textbf{0} \\ \hline
\multirow{12}{*}{\makecell[l]{Brain Tumor MRI Images Dataset\\with 44 Classes  \cite{fernando2022brain44}}} & \multirow{10}{*}{Tumor} & Carcinoma & 251 & 0 & 251 & 0 & 112 & 66 & 73 & 0 & 0 \\
 &  & Germinoma & 100 & 0 & 100 & 0 & 40 & 27 & 33 & 0 & 0 \\ 
 &  & Glioma & 1219 & 0 & 1219 & 0 & 465 & 382 & 372 & 0 & 0 \\ 
 &  & Granuloma & 78 & 0 & 78 & 0 & 31 & 30 & 17 & 0 & 0 \\ 
 &  & Meduloblastoma & 131 & 0 & 131 & 0 & 67 & 23 & 41 & 0 & 0 \\
 &  & Meningioma & 874 & 0 & 874 & 0 & 369 & 272 & 233 & 0 & 0 \\ 
 &  & Neurocitoma & 457 & 0 & 457 & 0 & 223 & 130 & 104 & 0 & 0 \\
 &  & Papiloma & 237 & 0 & 237 & 0 & 108 & 66 & 63 & 0 & 0 \\ 
 &  & Schwannoma & 465 & 0 & 465 & 0 & 194 & 148 & 123 & 0 & 0 \\
 &  & Tuberculoma & 145 & 0 & 145 & 0 & 84 & 28 & 33 & 0 & 0 \\
 & Normal & - & 522 & 0 & 522 & 0 & 0 & 251 & 271 & 0 & 0 \\  
 \hhline{~-----------}
 & \textbf{Dataset Total} & - & \textbf{4479} & \textbf{0} & \textbf{4479} & \textbf{0} & \textbf{1693} & \textbf{1423} & \textbf{1363} & \textbf{0} & \textbf{0} \\ \hline
\multirow{7}{*}{\makecell[l]{Brain Tumor MRI Images Dataset\\with 17 Classes \cite{fernando2022brain17}}} & \multirow{4}{*}{Tumor} & Glioma & 1317 & 0 & 1317 & 0 & 512 & 459 & 346 & 1317 & 0 \\ \
 &  & Meningioma & 1299 & 0 & 1299 & 0 & 625 & 345 & 329 & 1299 & 0 \\
 &  & Neurocitoma & 542 & 0 & 542 & 0 & 261 & 169 & 112 & 542 & 0 \\ 
 &  & Schwannoma & 470 & 0 & 470 & 0 & 194 & 153 & 123 & 470 & 0 \\ 
 & Other Abnormalities & Unspecified & 257 & 0 & 257 & 0 & 48 & 152 & 57 & 257 & 0 \\
 & Normal & - & 563 & 0 & 563 & 0 & 0 & 272 & 291 & 563 & 0 \\ 
 \hhline{~-----------}
 & \textbf{Dataset Total} & - & \textbf{4448} & \textbf{0} & \textbf{4448} & \textbf{0} & \textbf{1640} & \textbf{1550} & \textbf{1258} & \textbf{4448} & \textbf{0} \\ \hline
\multirow{3}{*}{Br35H \cite{hamada2020br35h}} & Tumor & - & 1500 & 0 & 1500 & 0 & 0 & 0 & 0 & 1500 & 0 \\ 
 & Normal & - & 1500 & 0 & 1500 & 0 & 0 & 0 & 0 & 1500 & 0 \\
 \hhline{~-----------}
 & \textbf{Dataset Total} & - & \textbf{3000} & \textbf{0} & \textbf{3000} & \textbf{0} & \textbf{0} & \textbf{0} & \textbf{0} & \textbf{3000} & \textbf{0} \\ \hline
\multirow{3}{*}{\makecell[l]{Multiple Sclerosis MRI\\Dataset \cite{buraktaci2022ms}}} & Multiple Sclerosis & - & 1411 & 0 & 1411 & 1411 & 0 & 0 & 0 & 650 & 761 \\ 
 & Normal & - & 2016 & 0 & 2016 & 2016 & 0 & 0 & 0 & 1002 & 1014 \\ 
 \hhline{~-----------}
 & \textbf{Dataset Total} & - & \textbf{3427} & \textbf{0} & \textbf{3427} & \textbf{3427} & \textbf{0} & \textbf{0} & \textbf{0} & \textbf{1652} & \textbf{1775} \\ \hline
\multirow{2}{*}{AISD \cite{liang2023aisd}} & Stroke & Ischemic & 4270 & 4270 & 0 & 0 & 0 & 0 & 0 & 4270 & 0 \\ 
\hhline{~-----------}
 & \textbf{Dataset Total} & - & \textbf{4270} & \textbf{4270} & \textbf{0} & \textbf{0} & \textbf{0} & \textbf{0} & \textbf{0} & \textbf{4270} & \textbf{0} \\ \hline
\multirow{5}{*}{Brain Stroke CT Dataset \cite{ozgur2022stroke}} & \multirow{3}{*}{Stroke} & Unspecified & 70 & 70 & 0 & 0 & 0 & 0 & 0 & 70 & 0 \\
 &  & Hemorrhagic & 1093 & 1093 & 0 & 0 & 0 & 0 & 0 & 1093 & 0 \\ 
 &  & Ischemic & 1130 & 1130 & 0 & 0 & 0 & 0 & 0 & 1130 & 0 \\
 & Normal & - & 4557 & 4557 & 0 & 0 & 0 & 0 & 0 & 4557 & 0 \\ 
 \hhline{~-----------}
 & \textbf{Dataset Total} & - & \textbf{6850} & \textbf{6850} & \textbf{0} & \textbf{0} & \textbf{0} & \textbf{0} & \textbf{0} & \textbf{6850} & \textbf{0} \\ \hline
\multicolumn{1}{|c|}{\textbf{TOTAL}} & - & - & \textbf{29538} & \textbf{11120} & \textbf{18418} & \textbf{3427} & \textbf{6397} & \textbf{2973} & \textbf{2621} & \textbf{20220} & \textbf{1775} \\ \hline
\end{tabular}
}
\end{table*}

\begin{figure}[!htbp]
\centering
\includegraphics[width=0.7\textwidth]{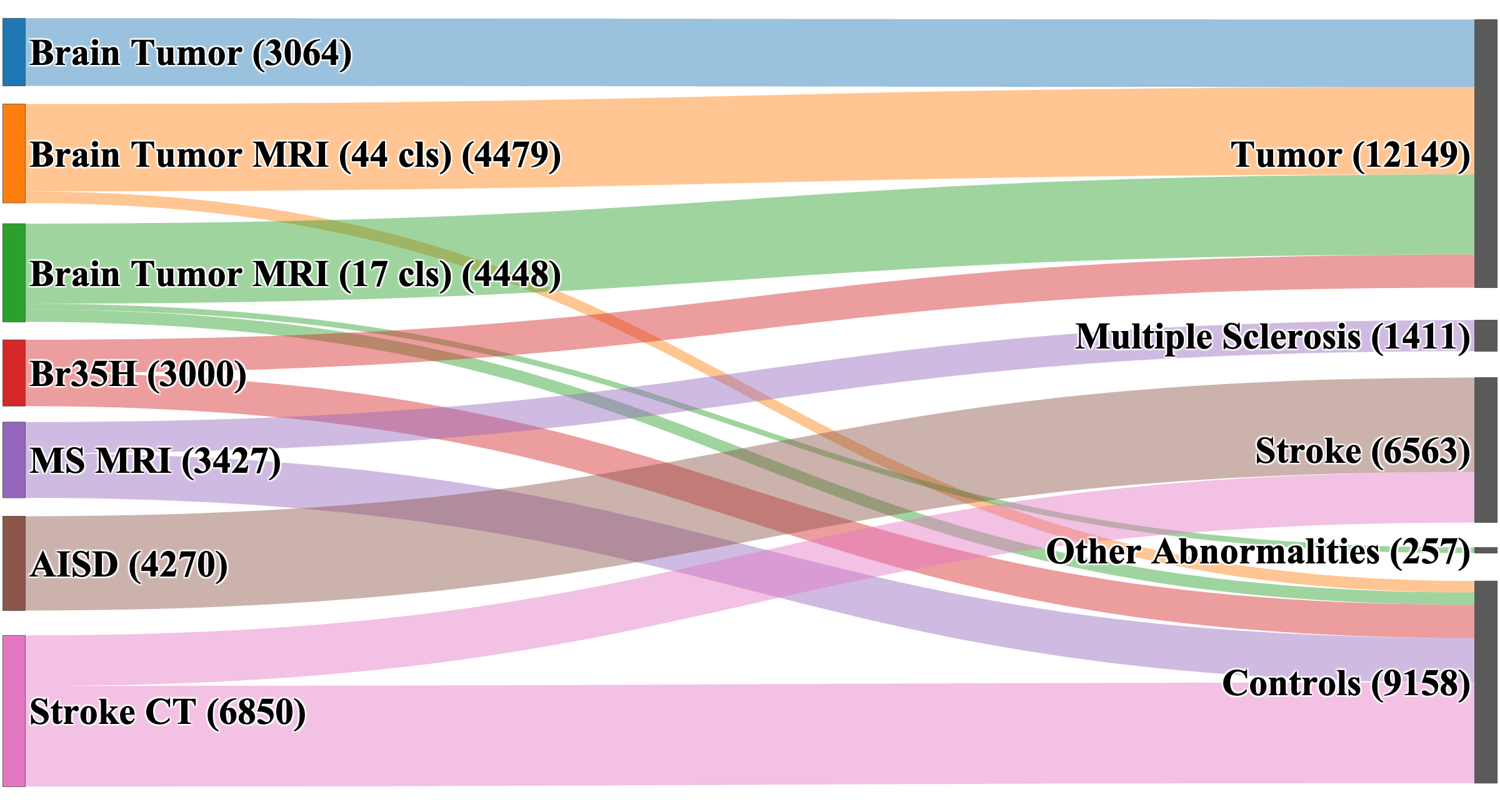}
\caption{Overview of the data origin and class composition}
\label{fig:overview}
\end{figure}

Fig.~\ref{fig:overview} provides an overview of data origin and class composition by mapping each source dataset to the benchmark’s main diagnostic classes, with node labels indicating sample counts.

\textbf{Brain tumors.} The Brain Tumor Dataset \cite{cheng2017brain} (Brain Tumor) contains 3,064 T1-weighted contrast-enhanced MRI slices from 233 patients, labeled into meningioma, glioma, and pituitary tumor. In addition, the Brain Tumor MRI Images datasets with 44 \cite{fernando2022brain44} (Brain Tumor MRI (44 cls)) and 17 classes \cite{fernando2022brain17} (Brain Tumor MRI (17 cls)) provide broader subtype labels. We also included Br35H - Brain Tumor Detection 2020 \cite{hamada2020br35h} dataset (Br35H), containing brain tumor images, and normal controls. 

\textbf{Multiple sclerosis.} The Multiple Sclerosis MRI Dataset \cite{buraktaci2022ms} (MS MRI) consists of axial and sagittal FLAIR MRIs acquired in a University Medical Faculty setting from 72 MS and 59 non-diseased male and female patients \cite{macin2022accurate}.

\textbf{Stroke.} The Acute Ischemic Stroke Dataset (AISD) \cite{liang2023aisd} includes 397 non-contrast CT scans, with ischemic lesions manually annotated by physicians and reviewed by senior experts, using DWI as reference. A complementary Stroke CT Dataset \cite{ozgur2022stroke} (Stroke CT) provides additional NCCT cases.

\textbf{Other abnormalities}. To increase clinical realism, we included a separate Other class comprising non-neoplastic mass-like lesions and encephalopathic patterns that frequently act as tumor mimickers. For example, abscesses may resemble high-grade gliomas or metastases as ring-enhancing lesions with necrosis and edema, where misclassification has immediate therapeutic consequences (antibiotic therapy versus surgery) \cite{toh2011differentiation,toh2014differentiation}. Similarly, cystic lesions (e.g., arachnoid or epidermoid cysts) and miscellaneous encephalopathies can be difficult to distinguish on conventional MRI and may lead to unnecessary interventions or missed diagnoses \cite{cui2024diffusion,gaillard2025tumefactive}. Including this class improves diagnostic robustness and better reflects challenging, high-risk imaging scenarios encountered in clinical practice.


\begin{table*}[t]
\caption{Class distribution in the benchmark dataset after merging all source datasets, grouped hierarchically by primary diagnosis and detailed diagnostic category, and further stratified by imaging modality, specialized MRI sequence (for MRI studies only), and imaging plane (when available).}
\label{tab:classes_grouped}
\centering
\resizebox{\textwidth}{!}{%
\begin{tabular}{|l|l|r|r|r|r|r|r|r|r|r|}
\hline
\multicolumn{2}{|c|}{\textbf{Identity}} & \multirow{2}{*}{\textbf{Count}} & \multicolumn{2}{c|}{\textbf{Modality}} & \multicolumn{4}{c|}{\textbf{Sequence (MRI)}} & \multicolumn{2}{c|}{\textbf{Plane}} \\
\hhline{--~--------}
\textbf{Main Class} & \textbf{Subclass} &  & \textbf{CT} & \textbf{MRI} & \textbf{FLAIR} & \textbf{T1C+} & \textbf{T1} & \textbf{T2} & \textbf{Axial} & \textbf{Sagittal} \\ \hline
\multirow{13}{*}{Tumor} & Glioma & 3962 & 0 & 3962 & 0 & 2398 & 812 & 718 & 1317 & 0 \\ 
 & Meningioma & 2881 & 0 & 2881 & 0 & 1702 & 617 & 562 & 1299  & 0 \\ 
 & Pituitary Tumor & 930 & 0 & 930 & 0 & 930 & 0 & 0 & 0 & 0 \\ 
 & Neurocitoma & 999 & 0 & 999 & 0 & 484 & 299 & 216 & 542 & 0 \\ 
 & Schwannoma & 935 & 0 & 935 & 0 & 388 & 301 & 246 & 470 & 0 \\ 
 & Carcinoma & 251 & 0 & 251 & 0 & 112 & 66 & 73 & 0 & 0 \\ 
 & Papiloma & 237 & 0 & 237 & 0 & 108 & 66 & 63 & 0 & 0 \\ 
 & Meduloblastoma & 131 & 0 & 131 & 0 & 67 & 23 & 41 & 0 & 0 \\ 
 & Tuberculoma & 145 & 0 & 145 & 0 & 84 & 28 & 33 & 0 & 0 \\ 
 & Germinoma & 100 & 0 & 100 & 0 & 40 & 27 & 33 & 0 & 0 \\ 
 & Granuloma & 78 & 0 & 78 & 0 & 31 & 30 & 17 & 0 & 0 \\ 
 & Unspecified & 1500 & 0 & 1500 & 0 & 0 & 0 & 0 & 1500 & 0 \\ 
\hhline{~----------}
 & \textbf{Class Total} & \textbf{12149} & \textbf{0} & \textbf{12149} & \textbf{0} & \textbf{6344} & \textbf{2269} & \textbf{2002} & \textbf{5128} & \textbf{0} \\ \hline
\multirow{2}{*}{Multiple Sclerosis} & Unspecified & 1411 & 0 & 1411 & 1411 & 0 & 0 & 0 & 650 & 761 \\ 
 \hhline{~----------}
 & \textbf{Class Total} & \textbf{1411} & \textbf{0} & \textbf{1411} & \textbf{1411} & \textbf{0} & \textbf{0} & \textbf{0} & \textbf{650} & \textbf{761} \\ \hline
\multirow{4}{*}{Stroke} & Ischemic & 5400 & 5400 & 0 & 0 & 0 & 0 & 0 & 5400 & 0 \\ 
 & Hemorrhagic & 1093 & 1093 & 0 & 0 & 0 & 0 & 0 & 1093 & 0 \\ 
 & Unspecified & 70 & 70 & 0 & 0 & 0 & 0 & 0 & 70 & 0 \\ 
  \hhline{~----------}
 & \textbf{Class Total} & \textbf{6563} & \textbf{6563} & \textbf{0} & \textbf{0} & \textbf{0} & \textbf{0} & \textbf{0} & \textbf{6563} & \textbf{0} \\ \hline
\multirow{2}{*}{Other Abnormalities} & Unspecified & 257 & 0 & 257 & 0 & 48 & 152 & 57 & 257 & 0 \\ 
 \hhline{~----------}
 & \textbf{Class Total} & \textbf{257} & \textbf{0} & \textbf{257} & \textbf{0} & \textbf{48} & \textbf{152} & \textbf{57} & \textbf{257} & \textbf{0} \\ \hline
\multirow{2}{*}{Normal} & Normal & 9158 & 4557 & 4601 & 2016 & 0 & 523 & 562 & 7622 & 1014 \\ 
 \hhline{~----------}
 & \textbf{Class Total} & \textbf{9158} & \textbf{4557} & \textbf{4601} & \textbf{2016} & \textbf{0} & \textbf{523} & \textbf{562} & \textbf{7622} & \textbf{1014} \\ \hline
\multicolumn{1}{|c|}{\textbf{TOTAL}} & - & \textbf{29538} & \textbf{11120} & \textbf{18418} & \textbf{3427} & \textbf{6392} & \textbf{2944} & \textbf{2621} & \textbf{20220} & \textbf{1775} \\ \hline
\end{tabular}%
}
\end{table*}

The benchmark dataset aggregates, per sample, not only the primary diagnostic class but also, where available, the detailed diagnosis, imaging modality, specialized MRI sequence, and imaging plane. This enables evaluation in both clinical and imaging-specific targets. After consolidating all source datasets, samples were grouped by primary and detailed diagnosis and summarized across the available imaging attributes (modality, MRI sequence, and imaging plane). Table~\ref{tab:classes_grouped} reports the resulting class distribution stratified by these dimensions.


\begin{figure}[!htbp]
\centering
\includegraphics[width=1.0\textwidth]
{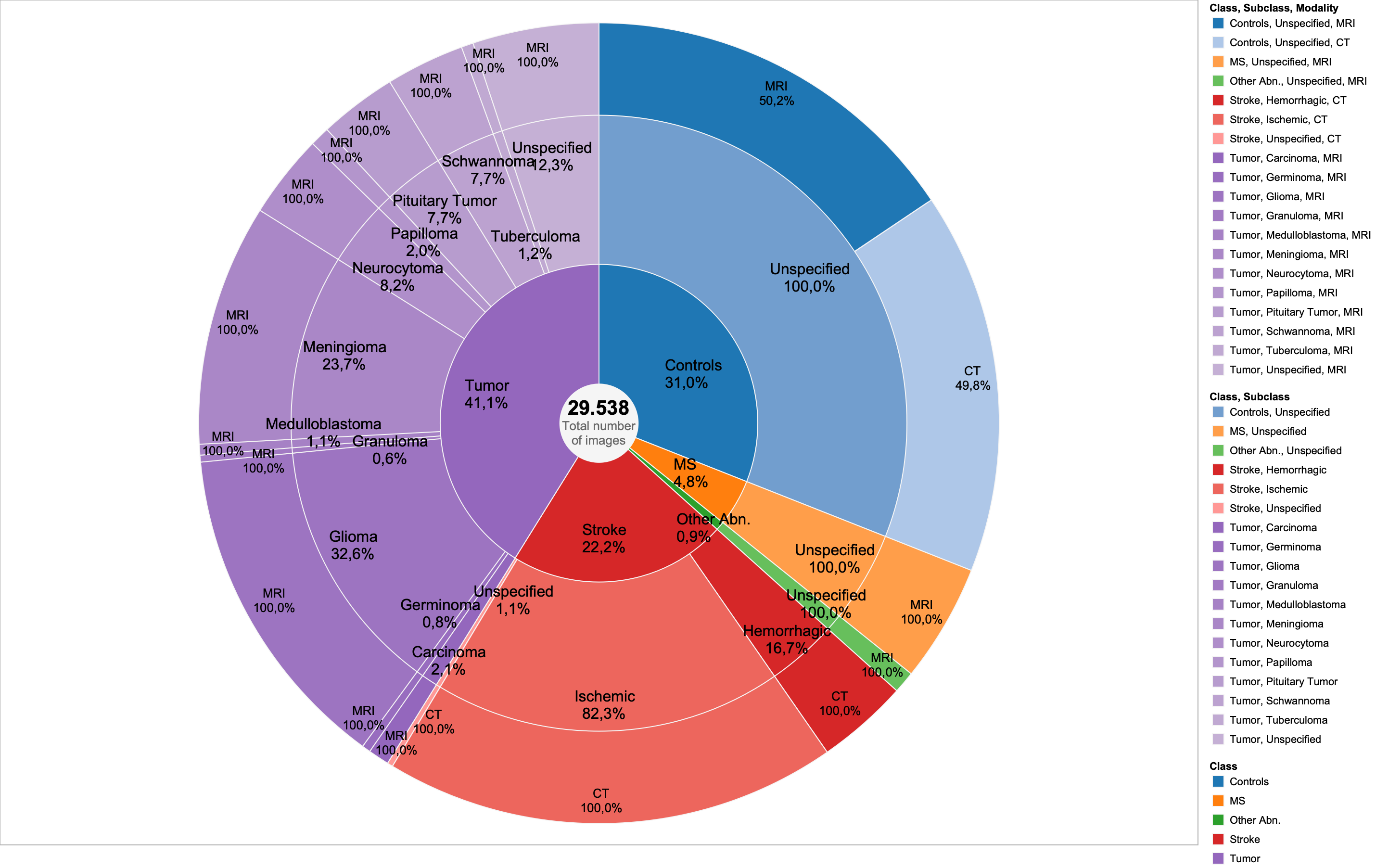}
\caption{Sunburst visualization of the benchmark’s output fields, illustrating the distribution of samples across diagnostic class, diagnostic subclass, and imaging modality.}
\label{fig:sunburst_high_contrast}
\end{figure}

\begin{figure}[htbp]
    \centering
    \begin{subfigure}[b]{0.48\textwidth}
        \centering
        \includegraphics[width=\textwidth]{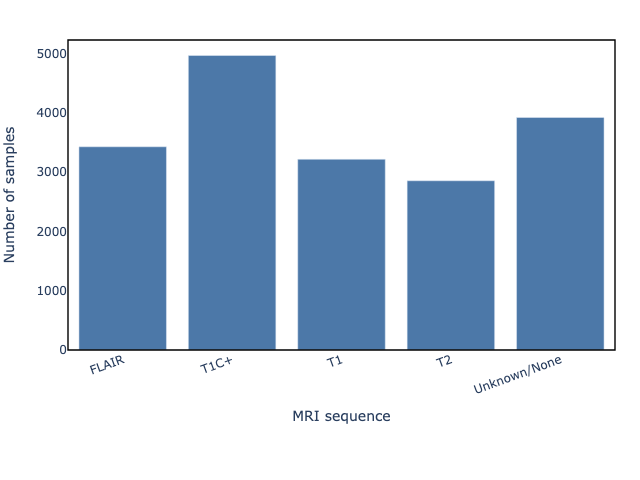}
        \caption{MRI-only sequence distribution across all MRI cases MRI samples included in the benchmark dataset}
        \label{fig:dist_class}
    \end{subfigure}
    \hfill
    \begin{subfigure}[b]{0.48\textwidth}
        \centering
        \includegraphics[width=\textwidth]{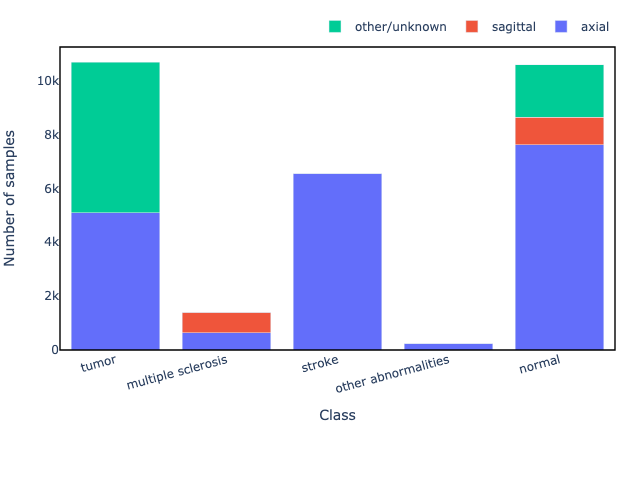}
        \caption{Class-wise distribution of imaging planes, reflecting heterogeneous availability of plane annotations across diagnostic categories.}
        \label{fig:dist_dataset}
    \end{subfigure}
    \caption{Distribution of imaging-specific output fields in the benchmark.}
    \label{fig:distributions}
\end{figure}

Fig.~\ref{fig:sunburst_high_contrast} visualizes a subset of the structured output fields such as diagnosis, detailed diagnosis, and modality, to highlight hierarchical imbalance and modality biases. Additionally, fig.~\ref{fig:distributions} summarizes imaging-specific fields related information, such as the MRI-only sequence distribution across all MRI samples, as well as the class-wise distribution of imaging planes, reflecting heterogeneous availability of plane annotations across diagnostic categories.

\subsection{Models}

In this study, we benchmark a diverse set of state-of-the-art multimodal large language models (MLLMs) on the curated neuroimaging benchmark. The evaluated models cover both open-source and proprietary systems, include general-purpose and medically oriented architectures, and range from lightweight instruction-tuned variants to frontier-scale multimodal foundation models. In fact, the model selection range across multiple major providers and research ecosystems, including OpenAI (GPT-5, GPT-4o, and GPT-4.1 series), Meta (LLaMA-3.2 and LLaMA-4 families), Google (Gemini, Gemma, and MedGemma families), Amazon (Nova models via Bedrock), Anthropic (Claude-Sonnet-4.5), xAI (Grok-4), and Alibaba (Qwen2.5-VL). This selection enables a systematic comparison between proprietary and open-source models with markedly different design philosophies, scales, and deployment constraints.

The evaluated models vary significantly in architecture, parameter count, context length, multimodal capabilities, and intended usage scenarios. At one end of the spectrum are lightweight, instruction-tuned vision-language models, such as gpt-5-mini and llama-3.2-11b-vision-instruct, which prioritize efficiency, low latency, and cost-effective inference. At the other end are large-scale multimodal systems, such as gpt-4.1-2025-04-14 and gemini-2.5-pro, designed to support advanced reasoning, long-context processing, and robust visual understanding. By including both categories, the benchmark captures realistic trade-offs between performance and usability in clinical and research settings.

The cost profiles of the evaluated models, summarized in Table~\ref{tab:cost_base}, differ significantly under the fully multimodal inference setting used in this study, in which all models are used with neuroimaging inputs. Input token pricing ranges from below \$0.05 per million tokens for smaller open-source vision models (e.g., LLaMA-3.2-11B Vision) to several dollars per million tokens for high-end proprietary systems, with output token costs reaching an order of magnitude higher for advanced multimodal models such as GPT-5 Chat, Grok-4, Gemini 2.5 Pro, and Claude Sonnet-4.5 to mention a few.

Beyond token costs, image input pricing introduces an additional and often dominant source of variability in multimodal inference. For instance, models from the Gemini family exhibit higher costs per image, reflecting the computational demands of large-scale vision–language alignment and long-context multimodal processing. Moreover, few-shot prompting further increases total inference cost, as exemplar images and their associated tokens are incorporated into each request, amplifying both token and image related expenses. These factors underscore the importance of evaluating multimodal models not only in terms of predictive performance, but also with respect to their practical scalability and cost efficiency in neuroimaging applications.

These pricing differences highlight not only the economic trade-offs associated with large-scale deployment of MLLMs, but also their practical accessibility for neuroimaging research, where inference cost, throughput, and reproducibility are critical considerations. By jointly analyzing diagnostic performance, calibration, structured-output reliability, and cost efficiency across this heterogeneous model set, our benchmark provides a comprehensive assessment of the suitability of contemporary LLMs and MLLMs for neuroimaging analysis tasks under realistic computational and financial constraints.

\begin{table*}[t]
\centering
\caption{Pricing of models per million tokens.} 
\label{tab:cost_base}
\fontsize{8.5}{10}\selectfont
\begin{threeparttable}
\begin{tabular}{lllll}
\hline
\multirow{2}[0]{*}{\textbf{Model}} & \multirow{2}[0]{*}{\textbf{Context}} & \multicolumn{2}{c}{\textbf{Cost per $1M$ tokens}} \\
 & & Input & Output  
 \\
\hline
GPT-5 Mini & 400k & 0.25 & 2.00 
\\
GPT-5 Chat & 128k & 1.25 & 10.00 
\\
GPT-4o Mini & 128k & 0.15 & 0.60 
\\
GPT-4o & 128k & 2.50 & 10.00 
\\
GPT-4.1 (April 2025) & 128k & 5.00 & 15.00 
\\
\hline
LLaMA~4 Maverick (Vision) & 1M & 0.15 & 0.60 
\\
LLaMA~3.2 90B Vision Instruct & 32k & 0.35 & 0.40 
\\
LLaMA~3.2 11B Vision Instruct & 131k & 0.049 & 0.049 
\\
\hline
MedGemma~1.5 4B &  & & 
\\
MedGemma~4B & 400k & 0.25& 0.25
\\ 
MedGemma~27B & 400k & 0.45& 0.45
\\ 
Gemma~3 27B Instruct & 96k & 0.065 & 0.261 
\\
Gemini~2.5 Pro & 1M & 1.25 & 10.00 
\\
Gemini~2.5 Flash & 1M & 0.30 & 2.50 
\\
Gemini~2.0 Flash & 1M & 0.10 & 0.40 
\\
\hline
Amazon Nova Pro 1.0 & 300k & 0.80 & 3.20 
\\  
Amazon Nova Lite 1.0 & 300K & 0.06 & 0.24 
\\ 
\hline
Claude Sonnet~4.5 & 1M &  3.00 &  15.00 
\\  
\hline
Grok~4 & 256k & 3.00 & 15.00 
\\  
\hline
Qwen~2.5-VL 32B Instruct & 16k & 0.05 & 0.22 
\\  
\hline
\end{tabular}
\begin{tablenotes}
\footnotesize
\item Pricing acquired from OpenRouter (\url{https://openrouter.ai/}) and AWS (\url{https://aws.amazon.com/bedrock/pricing/}) at 03.12.2025.
\end{tablenotes}
\end{threeparttable}
\end{table*}

\subsection{Experimental Setup - Benchmark Design}\label{sec:experimental_setup}

We established a four-tier evaluation framework designed to (i) develop and lock the experimental setup, (ii) perform robust model screening, (iii) confirm performance stability at scale, and (iv) conduct an unbiased final comparison on held-out data. Each phase served a distinct role in the evaluation pipeline, progressively narrowing the set of candidate models. This staged design enables efficient benchmarking of a large number of multimodal models while preserving fairness, reproducibility, and clear separation between model selection and final evaluation.
The curated benchmark dataset was partitioned into disjoint subsets. To minimize sampling bias and preserve representativeness, all splits were constructed using stratified sampling. Stratification was performed at multiple levels, including primary diagnostic categories (multiple sclerosis, stroke, brain tumors, other abnormalities, and normal controls), finer-grained subclasses (e.g., ischemic vs. hemorrhagic stroke, tumor subtypes), and key imaging attributes such as modality (MRI, CT), MRI sequence (T1, T1 contrast-enhanced, T2, FLAIR), and anatomical plane (axial, sagittal, coronal). This procedure preserved the joint distribution of diagnostic labels and imaging characteristics across all splits, ensuring that each subset remained representative of the full dataset.

Unlike conventional visual question answering–based evaluations, we adopt a multi-field structured prompting paradigm designed to better align benchmarking with real-world clinical documentation workflows. This formulation requires models to jointly produce diagnostic predictions, extract relevant imaging metadata, and comply with a predefined structured reporting schema. This way, the benchmark assesses not only diagnostic performance but also instruction adherence and structured output reliability, which are essential for integration into electronic health record systems. Importantly, this design exposes deployment-critical failure modes, such as incomplete fields, invalid formatting, or inconsistent metadata, that are largely orthogonal to question-answering accuracy and are typically overlooked in VQA-style evaluations. The source code for the evaluation is published on GitHub \footnote{https://github.com/chatmed/neurovlm-bench}.

\paragraph{\textbf{Phase 0: Experimental Setup Development and Locking (Screening Pool)}}

This phase was dedicated exclusively to protocol development and locking. For this purpose, we constructed three patient-disjoint screening subsets, each comprising 10\% of the full dataset (30\% in total). These subsets were used to construct and explore prompt formulations and decoding parameters on a representative subset of models covering all major model families.

During this stage, multiple decoding temperatures (0.0, 0.1, and 0.2) were evaluated. As no significant performance differences were observed between temperature settings and in accordance with previous findings in the literature \cite{bedi2505medhelm}, the temperature was fixed to 0.0 for all subsequent experiments. Additionally, top-p was set to 1.0, and a fixed random seed (42) was used to ensure deterministic behavior in repeated runs. Multiple prompt formulations were evaluated, including variations in wording, structure, and output schema specification, with the goal of ensuring consistent generation of structured JSON outputs and reliable recognition of all required prediction fields across diverse model families.

After these decisions were made, the prompt template and decoding configuration were frozen and remained unchanged throughout all subsequent phases.

\paragraph{\textbf{Phase 1: Screening Evaluation and Early Model Filtering}}

In Phase 1, all candidate models were evaluated under the frozen experimental setup in the same three screening subsets (30\% total). For each model, performance metrics were computed independently in each subset and subsequently aggregated by averaging across the three splits, producing a robust screening estimate that mitigates sensitivity to sampling variability.

These aggregated screening results were used exclusively for initial model filtering, reducing the candidate set from 20 to 11 models by eliminating those that consistently underperformed in the screening subsets. Screening-stage results are reported for transparency and diagnostic insight; however, they are conditioned on model selection.

\paragraph{\textbf{Phase 2: Development-Scale Confirmation}}

The models retained from Phase 1 were subsequently evaluated on a larger development split comprising approximately 45\% of the dataset, under the same frozen experimental setup. This stage served to confirm that performance trends observed during screening remained stable at larger scale and to further eliminate weaker or less stable models. Following this phase, the candidate set was reduced from 11 to 6 models. Results from Phase 2 are reported as development performance and are used for model selection.

\paragraph{\textbf{Phase 3: Final Evaluation with Zero-shot and Few-shot Prompting}}

The final evaluation was conducted on a strictly held-out test split comprising 25\% of the dataset, which was never used during protocol development, model screening, or development-stage filtering. Only the six top-performing models identified in Phase 2 were evaluated at this stage.

Models were assessed under two prompting regimes: zero-shot, in which only the task instruction and the query image were provided, and few-shot, in which exactly four labeled exemplars per diagnostic class (20 examples in total) were included in the prompt in addition to the instruction and query image. The exemplar set was fixed across all models and sourced exclusively from non-test data, ensuring that each model received identical supporting information and that no information from the test split was leaked into the prompt.

This final stage enabled a direct and controlled comparison between zero-shot and few-shot prompting under identical conditions. All comparative conclusions and claims regarding model performance are based on the results obtained from this held-out test split and should be interpreted as conditional on the models that passed the preceding screening and development stages.

The evaluation pipeline was structured in three successive phases (Fig.~\ref{fig:experimental_phases}), each designed to progressively narrow down candidate models while preserving reproducibility and fairness.

\usetikzlibrary{positioning}

\begin{figure*}[t]
\centering
\begin{adjustbox}{width=\textwidth}
\begin{tikzpicture}[
  every node/.style={align=center},
  >=stealth,
  block/.style={
    rectangle, draw, rounded corners,
    fill=blue!10,
    inner sep=6pt,
    minimum width=0pt,
    minimum height=0pt
  },
  arrow/.style={thick, ->, shorten >=2pt, shorten <=2pt}
]

\def\BlockSep{0.9cm}

\node[block, fill=red!15] (p0) {\textbf{Phase 0: Experimental setup calibration and freezing}\\
Screening pool: 3 $\times$ 10\% subsets (30\% total, stratified)\\
Representative subset of models (across families)\\
Tested temperatures $\{0.0, 0.1, 0.2\}$ and prompt variants (wording, structure, JSON schema)\\
Outcome: freeze setup (temperature=0.0, top-p=1.0, seed=42) and final prompt};

\node[block, fill=blue!10, below=\BlockSep of p0] (p1) {\textbf{Phase 1: Screening evaluation and initial filtering}\\
Screening pool: 30\% total, stratified\\
All candidate models (20 models) evaluated under the frozen setup\\
Aggregated performance: average over the 3 subsets\\
Outcome: select 11 models for Phase 2};

\node[block, fill=green!20, below=\BlockSep of p1] (p2) {\textbf{Phase 2: Development-scale confirmation}\\
45\% development split (held out from Phases 0--1)\\
11 selected models evaluated under the frozen setup\\
Outcome: confirm stability at scale, select 6 models for Phase 3};

\node[block, fill=yellow!20, below=\BlockSep of p2] (p3) {\textbf{Phase 3: Generalization
Benchmark - Final Evaluation with Zero-shot and Few-shot Prompting}\\
25\% test split (held out from Phases 0--2)\\
6 finalist models evaluated under identical conditions\\
Zero-shot vs. few-shot (4 exemplars per class; fixed across models; sourced from non-test data)\\
Outcome: main reportable benchmark results};

\draw[arrow] (p0) -- (p1);
\draw[arrow] (p1) -- (p2);
\draw[arrow] (p2) -- (p3);

\end{tikzpicture}
\end{adjustbox}
\caption{Overview of the staged benchmarking pipeline. Phase 0 calibrates and freezes the experimental setup. Phase 1 performs screening on a stratified 30\% pool and filters candidate models. Phase 2 confirms performance at development scale. Phase 3 reports final results on a strictly held-out test split under zero-shot and few-shot prompting.}
\label{fig:experimental_phases}
\end{figure*}
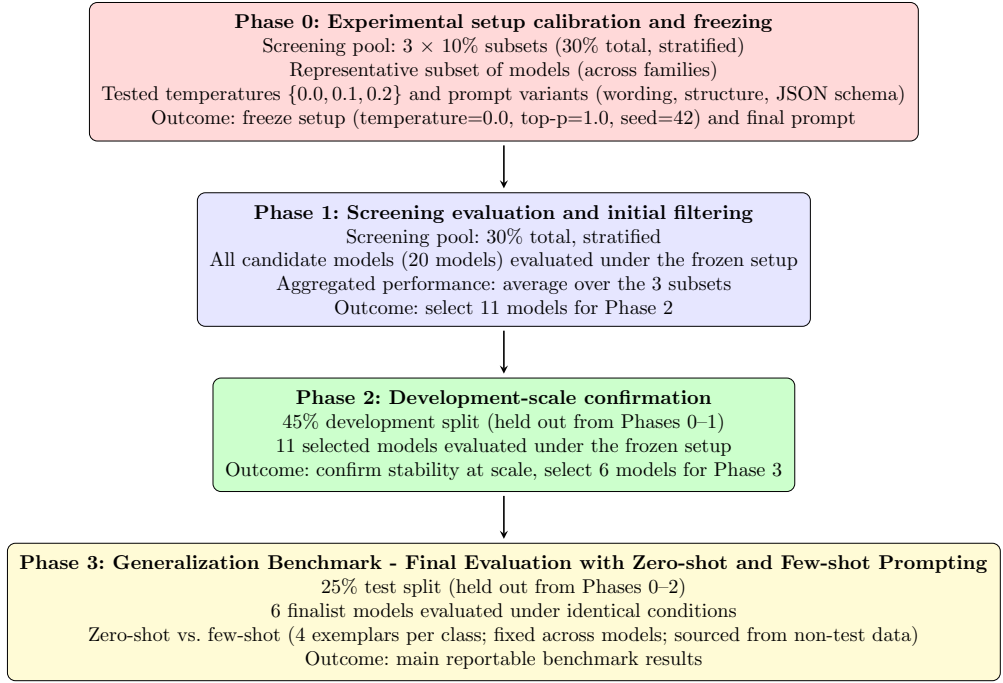

\subsection{Specification of the Prompt}

To standardize the evaluation of the vision-LLMs across all experimental settings, for this benchmark we defined a unified prompting, namely, the \textbf{Unified Neuro-Imaging Prompting Protocol (UNIPP)}.
This protocol is applied identically in both zero-shot and few-shot modes, differing only in whether the model receives exemplar demonstrations before generating its output.

\begin{itemize}
    \item \textbf{Zero-shot condition} - The model receives only the base prompt and the test image. No examples are provided.
This setting evaluates the ability of intrinsic medical reasoning without prior adaptation.

    \item \textbf{Few-shot condition} - The model receives K in-context examples (K = 20, 4 per class in our experiments), each containing a sample neuroimaging input paired with a correctly structured JSON output following the same schema. This setting evaluates in-context learning behavior and the ability to utilize structured radiology-style reporting. Based on the findings of previous benchmarks on multimodal learning of a few shots, we fixed the number of examples to four per class. Shakeri et al. systematically evaluated strict few-shot regimes in nine medical classification tasks and found that performance gains have started to show more dramatically in as few as 4 to 5 examples per class \cite{shakeri2024few}. Similarly, Ferber et al. demonstrated in histopathology classification that GPT-4V achieved substantial improvements when moving from 1 to 3 to 5 examples per class, but showed little additional benefit beyond this range \cite{ferber2024context}. The MedFMC benchmark further confirmed this trend: performance improved markedly from 1 to 5 examples per class, while only modestly increasing to 10 \cite{wang2023real}. Taken together, these studies provide consistent evidence that four examples per class capture the majority of few-shot learning gains while controlling computational cost, making it a principled choice for our benchmark.

\end{itemize}

In both settings, the model must generate output that conforms exactly to the prescribed JSON schema. The complete system prompt used within UNIPP is available on Github \footnote{\url{https://github.com/ChatMED/neurovlm-bench/blob/main/prompts/neurorad_prompt_trim_5classes_refined.txt}}, while the structured output schema is given in Listing~\ref{lst:output-schema} in the Appendix. Any additional explanatory text is not allowed.

The structured output vocabulary required by UNIPP defines six output fields grouped into three functional components:

\begin{enumerate}
  \item Image Metadata Inference
      \begin{itemize}
        \item modality
        \item specialized sequence
        \item plane
      \end{itemize}
  \item Diagnostic Reasoning 
    \begin{itemize}
        \item diagnosis name
        \item detailed diagnosis 
    \end{itemize}
  \item Quantitative Assessments
    \begin{itemize}
        \item diagnosis confidence
    \end{itemize}
\end{enumerate}

\begin{figure}[!htbp]
\centering
\includegraphics[width=\textwidth]{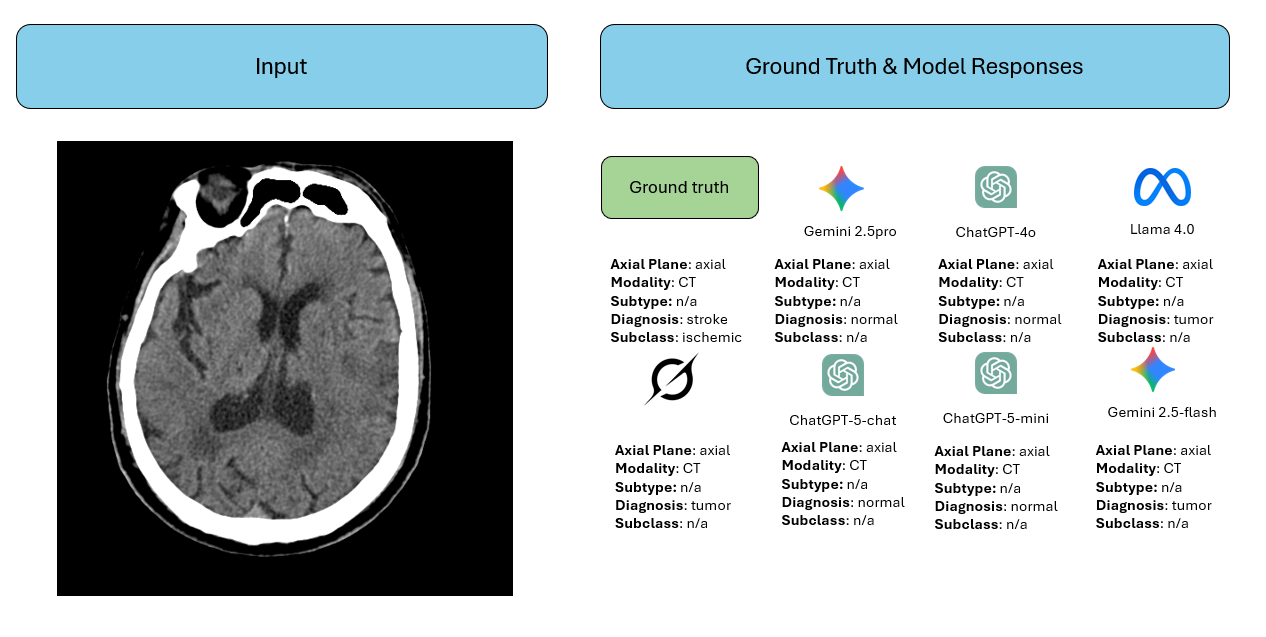}
\caption{Sample image input, the corresponding ground truth and the responses from some the models. We are trying to get structured response from the models, where each field (target) describes a specific part of the diagnostic report for the provided input.}
\label{fig:sample_input_output}
\end{figure}

According to this vocabulary, MRI maps to one of four allowed sequences (T1, T2, FLAIR, T1C+), while CT maps to a  \texttt{null} specialized sequence. Diagnostic predictions follow a hierarchical structure: high-level categories (e.g., tumor, stroke) map only to allowed subtypes. Categories such as multiple sclerosis or normal have no clinically defined subtypes and therefore map to \texttt{null}. This controlled vocabulary prevents hallucinated labels and enables consistent evaluation across models. These components correspond to minimal yet clinically meaningful elements of real radiology reporting workflows.
UNIPP ensures that all models, regardless of architecture or pretraining, are evaluated in a standardized and reproducible manner. Moreover, the schema enables: (1) closed-world categorical prediction for modality, specialized sequence, plane, and diagnostic categories; (2) hierarchical diagnostic reasoning, separating high-level categories (e.g., tumor, stroke) from subtype-specific labels (e.g., glioma, hemorrhagic); (3) radiology-compatible structured reporting, including confidence estimates; and (4) safety constraints that prohibit free-text explanations, suppress hallucinated terminology, and require \texttt{null} outputs when information is visually indeterminate.
These properties are essential for benchmarking large multimodal models in high-stakes clinical environments. A sample input and output is provided on Figure~\ref{fig:sample_input_output}.

\section{Evaluation}
\label{sec:evaluation}

\subsection{Evaluation} 

To comprehensively evaluate MLLM performance in a clinically relevant neuroimaging settings, a structured, multi-dimensional evaluation framework based on standard metrics is applied. It captures important aspects of model behavior and practical deployment beyond just raw predictive accuracy. The evaluation framework has four complementary dimensions: (i) discriminative classification performance (macro- and weighted F1, accuracy, precision, recall, and AUC), (ii) output reliability (structured JSON validity and undetermined/abstention rate), (iii) statistical calibration (ECE and Brier score), and (iv) operational efficiency (token usage, latency, and estimated cost per 1,000 images). 

Taking into account class imbalance and the possibility of model abstention, discriminative classification performance is assessed using the following metrics: abstention-aware macro-averaged F1 score, balanced accuracy, macro-averaged precision, macro-averaged recall, weighted macro-averaged F1 score, and macro-averaged one-vs-rest area under the receiver operating characteristic curve (AUC).

To quantify statistical uncertainty, 95\% confidence intervals (CI) \cite{jing2025beyond, aali2025structured, fraile2025measuring} are reported. Confidence intervals are computed using non-parametric bootstrap resampling of the evaluation set with 1,000 bootstrap iterations. Resampling is stratified by diagnosis to preserve the original class distribution. The 2.5th and 97.5th percentiles of the bootstrap distribution define the confidence bounds.

Abstention of the models was considered and addressed properly by computing abstention-aware metrics \cite{pal2023med, wen2024art}. Each abstained prediction is counted as a false negative for the ground-truth diagnostic class, while no false positive is assigned to any predicted class. This approach penalizes models that abstain excessively and ensures that performance metrics reflect predictive capability rather than avoidance of difficult cases.

The model outputs were subjected to light post-processing to normalize formatting and enforce consistency with the predefined diagnostic vocabulary, including case normalization and synonym resolution. Predictions that could not be unambiguously assigned to a valid vocabulary entry after normalization were treated as incorrect predictions. Explicit abstentions were not remapped and handled solely through the abstention-aware evaluation framework. 

All models were instructed to return the  predictions in a structured JSON format. The valid JSON rate is the proportion of outputs that can be parsed and are in accordance with the predefined schema. Outputs that could not be parsed properly, had missing required fields, or violated schema constraints were categorized as invalid JSON outputs. Such outputs were counted as incorrect predictions.

Additionally, we evaluated calibration using the scalar \textit{diagnosis confidence} reported by the model, intended to represent the probability that the predicted diagnosis is correct. Calibration performance is quantified using the ECE and Brier score, which assess probabilistic reliability. The ECE is computed by partitioning the predictions into B = 10 equally spaced confidence bins over the interval [0,1]. For each bin \textit{b}, the absolute difference between the mean predicted confidence and the empirical accuracy is computed and weighted by the proportion of samples in that bin. The final ECE is obtained by adding these weighted differences between all bins. High ECE values indicate over- or under-confident predictions, whereas low ECE values indicate well-calibrated confidence estimates.
The Brier score is computed as the mean squared error between the predicted confidence and a binary correctness indicator, with lower values indicating better calibration. Samples for which the model abstained or produced invalid JSON outputs were excluded from the calibration analysis, as there is no well-defined confidence–outcome pair for these cases.


Additionally, efficiency and cost were systematically recorded, including average input, output, and total tokens per query.
This provides a way to analyze trade-offs not only regarding diagnostic accuracy but also in the direction of practical scalability across different model families.

The primary task in our benchmark is image classification, formulated as predicting the correct diagnosis as an output field for a given neuroimaging sample (for example, determining whether a head CT is normal or shows a stroke, or identifying if a brain tumor is present on an MRI). However, we additionally evaluate the models on the other output fields, such as identifying the diagnosis subtype (e.g. brain tumor type, or whether the stroke is ischemic or hemorrhagic), and checking if the model is aware of the image modality and sequence, as well as anatomical plane, namely axial, sagittal, and coronal.

Multimodal models are known to exhibit performance degradation and sensitivity to distribution changes when evaluation data differs from the model’s instruction or domain distribution. Prior work has shown that multimodal image–text systems may be influenced by image or text distortions, and that dataset-dependent shifts can affect MLLM performance \cite{qiu2024multimodal_robustness, oh2025understanding}. In medical imaging benchmarks, such variability is often driven by differences in acquisition protocols, annotation practices, visual presentation of pathology or artifacts \cite{imam2025robustness}. Consequently, reporting only aggregated benchmark-level metrics can obscure such effects, as larger or visually simpler/harder datasets may disproportionately influence overall results. Therefore, a per-dataset evaluation of each constituent dataset is essential to confirm the robustness, interpretability, and generalizability of the models across heterogeneous data sources. Hence, in this paper, we also report per-dataset diagnostic performance. For each dataset, metrics are computed only over the diagnostic classes present in that dataset, thereby avoiding penalization for missing or non-applicable classes. Given the heterogeneous nature of the datasets (containing 1 to 5 classes), we used macro-averaged recall as our primary evaluation metric. Unlike the F1 score, which becomes undefined or misleading for single-class datasets, recall provides a consistent measure of model sensitivity across all datasets. For datasets with multiple classes where precision is meaningful, we also report F1 scores in supplementary materials.  

\subsection{Phase 1 Results: Screening evaluation and initial filtering}

Phase~1 aims to identify robust candidate models through screening evaluation on a 30\% stratified subset of the dataset. Because this stage serves as an initial filtering step, the results are intended to identify clearly non-competitive models and major reliability failures—such as excessive abstention or invalid structured outputs—rather than to provide definitive estimates of generalization performance. The results obtained in this phase therefore provide preliminary evidence regarding discriminative performance, calibration behavior, and operational characteristics. Final performance estimates are progressively validated in Phase~2 on the development split and ultimately finalized in Phase~3 on the held-out evaluation split.

The results related to discriminative baseline classification, output reliability, and calibration are presented in Table~\ref{tab:combined_performance_metrics}. The obtained results presents significant variability among the evaluated models. With respect to the primary metric, abstention-penalized Macro-F1, Gemini~2.5~Pro achieves the highest point estimate, while Gemini~2.5~Flash achieves the highest accuracy. GPT-5~Chat, GPT-4o, GPT-4.1, and Gemini~2.0~Flash also remain competitive, with strong weighted and micro-F1 values. Together, these models form the top-performing group in Phase~1. They additionally maintain near-perfect structured output compliance (typically $\geq$99.9\% valid JSON), indicating stable adherence to the structured prompting protocol.

\begin{table*}[!ht]
\centering
\caption{Phase 1 - Model performance on the primary diagnostic task evaluated for abstentions penalized discriminative performance metrics (Macro-F1, weighted Macro-F1, Micro-F1, Accuracy, AUC), calibration (ECE and Brier score), and structured output reliability (valid JSON rate, abstention rate). Values in parentheses denote 95\% confidence intervals. Bold underlined values indicate the best observed point estimate per column.
}
\label{tab:combined_performance_metrics}
\resizebox{\textwidth}{!}{%
\begin{threeparttable}
\begin{tabular}{l | c c c c c c c c c}
\toprule
\textbf{Model} & \textbf{Macro-F1} & \textbf{Macro-F1-weighted} & \textbf{Micro-F1} & \textbf{Accuracy} & \textbf{AUC} & \textbf{ECE} & \textbf{Brier} & \textbf{Valid JSON (\%)} & \textbf{Abstention Rate (\%)} \\
\midrule
\texttt{Amazon~Nova~Lite 1.0} 
 & \metric{0.326410} (\ci{0.316474}--\ci{0.338939})
 & \metric{0.515218} (\ci{0.493558}--\ci{0.536154})
 & \metric{0.584861} (\ci{0.570944}--\ci{0.599652})
 & \metric{0.371288} (\ci{0.364739}--\ci{0.381738})
 & \metric{0.571193}
 & \metric{0.336982}
 & \metric{0.353870}
 & \metric{99.86}
 & \underline{\textbf{\metric{0.0000}}} \\

\texttt{Amazon~Nova~Pro 1.0} 
 & \metric{0.323336} (\ci{0.308361}--\ci{0.338806})
 & \metric{0.510315} (\ci{0.497380}--\ci{0.528184})
 & \metric{0.512220} (\ci{0.494535}--\ci{0.525297})
 & \metric{0.329633} (\ci{0.307718}--\ci{0.357518})
 & \metric{0.564061}
 & \metric{0.419851}
 & \metric{0.421972}
 & \metric{99.86}
 & \underline{\textbf{\metric{0.0000}}} \\

\midrule
\texttt{Gemini~2.0~Flash} 
 & \metric{0.536523} (\ci{0.517023}--\ci{0.553664})
 & \metric{0.714269} (\ci{0.698675}--\ci{0.728868})
 & \metric{0.711428} (\ci{0.696668}--\ci{0.728577})
 & \metric{0.602889} (\ci{0.556100}--\ci{0.639214})
 & \metric{0.764689}
 & \metric{0.142947}
 & \metric{0.198768}
 & \metric{99.97}
 & \underline{\textbf{\metric{0.0000}}} \\

\texttt{Gemini~2.5~Flash} 
 & \metric{0.569297} (\ci{0.546298}--\ci{0.589599})
 & \metric{0.739263} (\ci{0.722197}--\ci{0.751537})
 & \metric{0.728630} (\ci{0.713365}--\ci{0.743284})
 & \underline{\textbf{\metric{0.617461}}} (\ci{0.575666}--\ci{0.673048})
 & \metric{0.615188}
 & \metric{0.222948}
 & \metric{0.238120}
 & \metric{99.93}
 & \underline{\textbf{\metric{0.0000}}} \\

\texttt{Gemini~2.5~Pro} 
 & \underline{\textbf{\metric{0.573162}}} (\ci{0.553469}--\ci{0.592033})
 & \underline{\textbf{\metric{0.749430}}} (\ci{0.733475}--\ci{0.764675})
 & \underline{\textbf{\metric{0.748050}}} (\ci{0.732274}--\ci{0.761131})
 & \metric{0.580074} (\ci{0.557211}--\ci{0.614725})
 & \metric{0.591086}
 & \metric{0.219047}
 & \metric{0.234076}
 & \metric{99.97}
 & \underline{\textbf{\metric{0.0000}}} \\

\texttt{Gemma~3~27B~IT} 
 & \metric{0.325991} (\ci{0.308369}--\ci{0.340862})
 & \metric{0.477328} (\ci{0.456238}--\ci{0.495702})
 & \metric{0.484407} (\ci{0.466873}--\ci{0.499839})
 & \metric{0.348651} (\ci{0.315191}--\ci{0.376756})
 & \metric{0.761128}
 & \metric{0.392359}
 & \metric{0.374202}
 & \underline{\textbf{100}}
 & \underline{\textbf{\metric{0.0000}}} \\

\texttt{MedGemma~1~27B} 
 & \metric{0.242964} (\ci{0.231328}--\ci{0.253999})
 & \metric{0.442897} (\ci{0.422366}--\ci{0.467184})
 & \metric{0.544407} (\ci{0.527958}--\ci{0.559322})
 & \metric{0.289322} (\ci{0.282918}--\ci{0.296707})
 & \metric{0.471314}
 & \metric{0.376485}
 & \metric{0.392529}
 & \underline{\textbf{100}}
 & \underline{\textbf{\metric{0.0000}}} \\

\texttt{MedGemma~1~4B} 
 & \metric{0.382179} (\ci{0.361708}--\ci{0.397426})
 & \metric{0.556666} (\ci{0.537242}--\ci{0.575115})
 & \metric{0.586102} (\ci{0.569780}--\ci{0.599000})
 & \metric{0.390578} (\ci{0.373603}--\ci{0.406727})
 & \metric{0.601516}
 & \metric{0.342217}
 & \metric{0.349997}
 & \underline{\textbf{100}}
 & \underline{\textbf{\metric{0.0000}}} \\

 \texttt{MedGemma~1.5~4B} 
 & \metric{0.397537} (\ci{0.389247}--\ci{0.407376})
 & \metric{0.582180} (\ci{0.570128}--\ci{0.593114})
 & \metric{0.628085} (\ci{0.616635}--\ci{0.636272})
 & \metric{0.435563} (\ci{0.421747}--\ci{0.448919})
 & \metric{0.571725}
 & \metric{0.319334}
 & \metric{0.322571}
 & \underline{\metric{99.58}}
 & \underline{\textbf{\metric{1.134687}}} \\

\midrule
\texttt{LLaMA~3.2~11B~Vision~Instruct} 
 & \metric{0.038705} (\ci{0.027716}--\ci{0.048590})
 & \metric{0.054683} (\ci{0.045652}--\ci{0.061966})
 & \metric{0.061918} (\ci{0.051981}--\ci{0.072203})
 & \metric{0.023802} (\ci{0.017740}--\ci{0.030922})
 & \underline{\textbf{\metric{0.949270}}}
 & \metric{0.093885}
 & \underline{\textbf{\metric{0.090914}}}
 & \metric{99.29}
 & \metric{86.4117} \\

\texttt{LLaMA~3.2~90B~Vision~Instruct} 
 & \metric{0.147882} (\ci{0.132344}--\ci{0.164850})
 & \metric{0.253005} (\ci{0.238069}--\ci{0.268814})
 & \metric{0.288260} (\ci{0.269383}--\ci{0.308713})
 & \metric{0.104534} (\ci{0.096850}--\ci{0.114388})
 & \metric{0.932745}
 & \underline{\textbf{\metric{0.093445}}}
 & \metric{0.098391}
 & \metric{99.86}
 & \metric{70.4684} \\

\midrule
\texttt{LLaMA~4~Maverick} 
 & \metric{0.476787} (\ci{0.454683}--\ci{0.496364})
 & \metric{0.656579} (\ci{0.639797}--\ci{0.676000})
 & \metric{0.660598} (\ci{0.641999}--\ci{0.678225})
 & \metric{0.536940} (\ci{0.489526}--\ci{0.583322})
 & \metric{0.624403}
 & \metric{0.240533}
 & \metric{0.269610}
 & \metric{99.76}
 & \metric{2.1747} \\

\midrule
\texttt{GPT-4.1} 
 & \metric{0.527258} (\ci{0.499630}--\ci{0.552145})
 & \metric{0.703134} (\ci{0.685032}--\ci{0.717157})
 & \metric{0.699220} (\ci{0.685317}--\ci{0.716548})
 & \metric{0.530478} (\ci{0.494356}--\ci{0.565394})
 & \metric{0.423530}
 & \metric{0.265409}
 & \metric{0.280742}
 & \metric{99.97}
 & \underline{\textbf{\metric{0.0000}}} \\

\texttt{GPT-4o} 
 & \metric{0.548169} (\ci{0.518303}--\ci{0.584415})
 & \metric{0.719357} (\ci{0.703678}--\ci{0.736293})
 & \metric{0.718097} (\ci{0.703647}--\ci{0.734642})
 & \metric{0.538551} (\ci{0.505183}--\ci{0.571082})
 & \metric{0.562131}
 & \metric{0.214556}
 & \metric{0.244823}
 & \underline{\textbf{100}}
 & \metric{0.5085} \\

\texttt{GPT-4o~Mini} 
 & \metric{0.333869} (\ci{0.322460}--\ci{0.346414})
 & \metric{0.568067} (\ci{0.541760}--\ci{0.586447})
 & \metric{0.594980} (\ci{0.576535}--\ci{0.614801})
 & \metric{0.381419} (\ci{0.350925}--\ci{0.419377})
 & \metric{0.463925}
 & \metric{0.318186}
 & \metric{0.334970}
 & \underline{\textbf{100}}
 & \metric{0.1356} \\

\texttt{GPT-5~Chat} 
 & \metric{0.540552} (\ci{0.522430}--\ci{0.561386})
 & \metric{0.731206} (\ci{0.716206}--\ci{0.745453})
 & \metric{0.736610} (\ci{0.719271}--\ci{0.752898})
 & \metric{0.563001} (\ci{0.538449}--\ci{0.586885})
 & \metric{0.723807}
 & \metric{0.189732}
 & \metric{0.222088}
 & \underline{\textbf{100}}
 & \underline{\textbf{\metric{0.0000}}} \\

\texttt{GPT-5~Mini} 
 & \metric{0.480252} (\ci{0.460539}--\ci{0.502081})
 & \metric{0.669241} (\ci{0.652631}--\ci{0.684982})
 & \metric{0.621777} (\ci{0.606979}--\ci{0.639629})
 & \metric{0.540818} (\ci{0.500792}--\ci{0.585750})
 & \metric{0.632006}
 & \metric{0.239613}
 & \metric{0.281157}
 & \metric{99.93}
 & \underline{\textbf{\metric{0.0000}}} \\

\midrule
\texttt{Claude~Sonnet~4.5} 
 & \metric{0.375700} (\ci{0.345978}--\ci{0.408240})
 & \metric{0.582606} (\ci{0.564803}--\ci{0.596578})
 & \metric{0.580218} (\ci{0.559688}--\ci{0.601414})
 & \metric{0.376254} (\ci{0.358913}--\ci{0.398821})
 & \metric{0.513364}
 & \metric{0.319953}
 & \metric{0.346057}
 & \metric{87.08}
 & \metric{0.0779} \\

\midrule
\texttt{Grok~4} 
 & \metric{0.462538} (\ci{0.443314}--\ci{0.479108})
 & \metric{0.683964} (\ci{0.667958}--\ci{0.699296})
 & \metric{0.684168} (\ci{0.667794}--\ci{0.700848})
 & \metric{0.459585} (\ci{0.440338}--\ci{0.477181})
 & \metric{0.419493}
 & \metric{0.261421}
 & \metric{0.287678}
 & \metric{84.00}
 & \metric{0.1614} \\

\midrule
\texttt{Qwen~2.5-VL~32B~Instruct} 
 & \metric{0.342273} (\ci{0.330875}--\ci{0.352788})
 & \metric{0.546179} (\ci{0.529729}--\ci{0.566900})
 & \metric{0.546619} (\ci{0.529050}--\ci{0.564263})
 & \metric{0.332520} (\ci{0.320502}--\ci{0.344090})
 & \metric{0.534798}
 & \metric{0.310983}
 & \metric{0.332093}
 & \underline{\textbf{100}}
 & \metric{9.4915} \\

\bottomrule
\end{tabular}
\end{threeparttable}
}
\end{table*}


The confidence intervals reported in Table~\ref{tab:combined_performance_metrics} indicate that several of the highest-performing models exhibit overlapping or closely adjacent confidence intervals confidence intervals across multiple performance metrics, including the primary metric (abstention-penalized Macro-F1). In particular, Gemini~2.5~Pro and Gemini~2.5~Flash show overlapping intervals, while GPT-5~Chat, GPT-4o, GPT-4.1, and Gemini~2.0~Flash achieve closely comparable point estimates with overlapping confidence intervals. This suggests that the observed differences between these models fall within the uncertainty of the estimates. Therefore that no single model demonstrates clearly superior performance at the screening stage. Beyond this leading group, several additional models with moderate diagnostic performance exhibit similar patterns of overlapping or closely adjacent intervals. These observations are taken into account when selecting the models that will undergo further evaluation in Phase~2 for stability validation on the larger development split.

The differences between accuracy and Macro-F1 that can be noted in the results, reflect the class imbalance present in the dataset. Accuracy and micro-F1 are influenced more strongly by majority classes, whereas abstention-penalized Macro-F1 gives equal weight to each diagnostic category and therefore provides a more informative measure of clinically meaningful diagnostic performance. For this reason, Macro-F1 with abstention penalty is used as the primary ranking metric throughout the benchmark.

A second cluster of models exhibits moderate overall diagnostic performance based on the primary ranking metric. Systems such as LLaMA~4~Maverick, GPT-5~Mini, Grok~4, MedGemma~1.5~4B, and MedGemma~1~4B achieve intermediate Macro-F1 scores, indicating partial diagnostic capability but lower robustness compared to the leading models. These models remain operational under the structured output protocol but do not reach the discriminative performance levels observed in the top-performing group.

Moreover, it is important to emphasize that certain vision-enabled models display a pronounced mismatch between ranking-based and classification-based performance. Meta LLaMA~3.2 Vision-Instruct variants are an example of such behavior, achieving the highest AUC values in Phase~1 while simultaneously exhibiting near-zero accuracy and Macro-F1. This pattern indicates that, although the models may assign higher confidence scores to correct classes in a ranking sense, this signal is not translated into correct discrete diagnostic predictions under the structured output protocol. The observation highlights an important limitation of relying on ranking-based metrics such as AUC when evaluating multimodal models for structured clinical prediction tasks, where the final diagnostic label rather than class ranking determines clinical usefulness.

Beyond discriminative performance, Phase~1 highlights reliability-relevant behaviors that are not captured by accuracy alone. Most top-performing models exhibited negligible abstention. However, it is evident that models differed in calibration (ECE and Brier score), motivating further analysis on a larger split. Structured output validity was generally high across models, but Grok~4 and Claude~Sonnet~4.5 exhibited significantly lower valid-JSON rates (84--87\%). This represents deployment-relevant failures for structured reporting pipelines even when classification performance was moderate. On the contrary, some open-weight vision models failed primarily through excessive abstention. Namely, the LLaMA~3.2 Vision series showed near-zero Macro-F1 and accuracy, largely driven by extremely high abstention rates, indicating poor operational behavior under the benchmark’s structured output requirements.

\begin{table*}[ht!]
\centering
\caption{Phase 1 field-wise performance (F1 with abstention penalty) across available structured output fields: primary diagnosis, detailed diagnosis, imaging modality, specialized MRI sequence (MRI-only), and imaging plane (when annotated).
Values in parentheses denote 95\% confidence intervals. 
Bold underlined values indicate the best point estimate per column}
\label{tab:f1_score_per_output_field_phase1}
\resizebox{\textwidth}{!}{%
\begin{tabular}{lccccc}
\toprule
\makecell[c]{Model} &
\makecell[c]{Diagnosis} &
\makecell[c]{Detailed\\diagnosis} &
\makecell[c]{Modality} &
\makecell[c]{Specialized\\sequence} &
\makecell[c]{Plane} \\
\midrule
\texttt{\iffalse bedrock/ \fi Amazon~Nova~Lite}  & \metric{0.326410} (\ci{0.312919} - \ci{0.336699})  & \metric{0.085546} (\ci{0.079406} - \ci{0.092623})  & \metric{0.972475} (\ci{0.966877} - \ci{0.978334})  & \metric{0.263485} (\ci{0.247436} - \ci{0.280532})  & \metric{0.696457} (\ci{0.669228} - \ci{0.719506}) \\
\texttt{\iffalse bedrock/ \fi Amazon~Nova~Pro}  & \metric{0.323336} (\ci{0.305534} - \ci{0.339461})  & \metric{0.073091} (\ci{0.067001} - \ci{0.080140})  & \metric{0.912533} (\ci{0.901385} - \ci{0.922636})  & \metric{0.321422} (\ci{0.303225} - \ci{0.339696})  & \metric{0.428436} (\ci{0.409311} - \ci{0.449238}) \\
\hline
\texttt{\iffalse google/ \fi Gemini~2.0~Flash}  & \metric{0.536523} (\ci{0.517146} - \ci{0.553442})  & \metric{0.209974} (\ci{0.193961} - \ci{0.223207})  & \metric{0.999639} (\ci{0.999083} - \ci{1.000000})  & \metric{0.773993} (\ci{0.755541} - \ci{0.791069})  & \metric{0.983469} (\ci{0.972644} - \ci{0.991451}) \\
\texttt{Gemini~2.5~Flash}  & \metric{0.569297} (\ci{0.549526} - \ci{0.591401})  & \metric{0.270373} (\ci{0.240171} - \ci{0.300947})  & \metric{0.999278} (\ci{0.998197} - \ci{1.000000})  & \metric{0.784940} (\ci{0.762777} - \ci{0.803900})  & \metric{0.975588} (\ci{0.964304} - \ci{0.987218}) \\
\texttt{\iffalse google/ \fi Gemini~2.5~Pro}  & \underline{\textbf{\metric{0.573162}}} (\ci{0.557505} - \ci{0.588030}) & \underline{\textbf{\metric{0.319334}}} (\ci{0.298040} - \ci{0.340375}) & \metric{0.999639} (\ci{0.998902} - \ci{1.000000})  & \metric{0.781429} (\ci{0.764232} - \ci{0.801249})  & \metric{0.974158} (\ci{0.961319} - \ci{0.983906}) \\
\texttt{\iffalse google/ \fi Gemma~3~27B~IT}  & \metric{0.325991} (\ci{0.304070} - \ci{0.341046})  & \metric{0.065951} (\ci{0.060024} - \ci{0.072509})  & \metric{0.996756} (\ci{0.994535} - \ci{0.998555})  & \metric{0.222238} (\ci{0.200774} - \ci{0.240810})  & \metric{0.988028} (\ci{0.978534} - \ci{0.994253}) \\
\texttt{MedGemma~1~27B}  & \metric{0.242964} (\ci{0.234493} - \ci{0.252725})  & \metric{0.040779} (\ci{0.034932} - \ci{0.046114})  & \underline{\textbf{\metric{1.000000}}} (\ci{1.000000} - \ci{1.000000}) & \metric{0.141955} (\ci{0.130051} - \ci{0.156311})  & \underline{\textbf{\metric{1.000000}}} (\ci{1.000000} - \ci{1.000000}) \\
\texttt{MedGemma~1~4B} & \metric{0.382179} (\ci{0.364595} - \ci{0.400703})  & \metric{0.109632} (\ci{0.094297} - \ci{0.124630})  & \underline{\textbf{\metric{1.000000}}} (\ci{1.000000} - \ci{1.000000}) & \metric{0.171594} (\ci{0.155531} - \ci{0.187052})  & \underline{\textbf{\metric{1.000000}}} (\ci{1.000000} - \ci{1.000000}) \\
\texttt{MedGemma~1.5~4B} & \metric{0.397537} (\ci{0.387375} - \ci{0.407895})  & \metric{0.074416} (\ci{0.066586} - \ci{0.082013})  & \metric{0.951393} (\ci{0.946488} - \ci{0.955322}) & \metric{0.249914} (\ci{0.238009} - \ci{0.260796})  & \metric{0.497094} (\ci{0.485546} - \ci{0.508877}) \\
\hline
\texttt{\iffalse meta-llama/ \fi LLaMA~3.2~11B~Vision~Instruct} & \metric{0.038705} (\ci{0.029180} - \ci{0.050505})  & \metric{0.003554} (\ci{0.002276} - \ci{0.005764})  & \metric{0.176636} (\ci{0.160941} - \ci{0.195023})  & \metric{0.039124} (\ci{0.027699} - \ci{0.051643})  & \metric{0.148477} (\ci{0.138329} - \ci{0.160494}) \\
\texttt{\iffalse meta-llama/ \fi LLaMA~3.2~90B~Vision~Instruct} & \metric{0.147882} (\ci{0.133507} - \ci{0.166445})  & \metric{0.020169} (\ci{0.017019} - \ci{0.023110})  & \metric{0.325016} (\ci{0.313522} - \ci{0.338319})  & \metric{0.338100} (\ci{0.313790} - \ci{0.360467})  & \metric{0.256211} (\ci{0.221039} - \ci{0.289748}) \\
\texttt{\iffalse meta-llama/ \fi LLaMA~4~Maverick}  & \metric{0.476787} (\ci{0.453234} - \ci{0.497942})  & \metric{0.235633} (\ci{0.216903} - \ci{0.255537})  & \metric{0.986858} (\ci{0.983418} - \ci{0.990419})  & \metric{0.620540} (\ci{0.594843} - \ci{0.638435})  & \metric{0.984904} (\ci{0.975196} - \ci{0.991593}) \\
\hline
\texttt{\iffalse openai/ \fi GPT-4.1}  & \metric{0.527258} (\ci{0.501151} - \ci{0.560273})  & \metric{0.188507} (\ci{0.171439} - \ci{0.209652})  & \metric{0.998917} (\ci{0.997304} - \ci{1.000000})  & \metric{0.752409} (\ci{0.735911} - \ci{0.775673})  & \metric{0.987323} (\ci{0.976799} - \ci{0.992931}) \\
\texttt{\iffalse openai/ \fi GPT-4o}  & \metric{0.548169} (\ci{0.511851} - \ci{0.585176})  & \metric{0.207459} (\ci{0.192672} - \ci{0.221932})  & \metric{0.996867} (\ci{0.995228} - \ci{0.998403})  & \metric{0.760587} (\ci{0.746159} - \ci{0.778689})  & \metric{0.988174} (\ci{0.980243} - \ci{0.994871}) \\
\texttt{\iffalse openai/ \fi GPT-4o~Mini} & \metric{0.333869} (\ci{0.319771} - \ci{0.346838})  & \metric{0.059621} (\ci{0.052087} - \ci{0.068224})  & \metric{0.995844} (\ci{0.993728} - \ci{0.997876})  & \metric{0.471175} (\ci{0.453065} - \ci{0.490791})  & \metric{0.994112} (\ci{0.988323} - \ci{0.998257}) \\
\texttt{\iffalse openai/ \fi GPT-5~Chat}  & \metric{0.540552} (\ci{0.516061} - \ci{0.565841})  & \metric{0.251968} (\ci{0.227622} - \ci{0.280199})  & \metric{0.997837} (\ci{0.996389} - \ci{0.999287})  & \metric{0.806169} (\ci{0.790112} - \ci{0.820607})  & \metric{0.990848} (\ci{0.984102} - \ci{0.996190}) \\
\texttt{\iffalse openai/ \fi GPT-5~Mini}  & \metric{0.480252} (\ci{0.460372} - \ci{0.496925})  & \metric{0.299392} (\ci{0.271403} - \ci{0.328393})  & \metric{0.997835} (\ci{0.996387} - \ci{0.999465})  & \underline{\textbf{\metric{0.811324}}} (\ci{0.794958} - \ci{0.825621}) & \metric{0.988423} (\ci{0.977234} - \ci{0.995753}) \\
\hline
\texttt{\iffalse anthropic/ \fi Claude~Sonnet~4.5}  & \metric{0.375700} (\ci{0.349539} - \ci{0.407095})  & \metric{0.173845} (\ci{0.151111} - \ci{0.189939})  & \metric{0.997495} (\ci{0.995255} - \ci{0.999384})  & \metric{0.588091} (\ci{0.563906} - \ci{0.612492})  & \metric{0.998062} (\ci{0.993275} - \ci{1.000000}) \\
\hline
\texttt{\iffalse x-ai/ \fi Grok~4}  & \metric{0.462538} (\ci{0.448031} - \ci{0.481173})  & \metric{0.215427} (\ci{0.188167} - \ci{0.241527})  & \metric{0.998062} (\ci{0.996230} - \ci{0.999355})  & \metric{0.510785} (\ci{0.486455} - \ci{0.532518})  & \metric{0.997803} (\ci{0.994322} - \ci{0.999853}) \\
\hline
\texttt{Qwen~2.5-VL~32B~Instruct}  & \metric{0.342273} (\ci{0.329826} - \ci{0.356544})  & \metric{0.093471} (\ci{0.083368} - \ci{0.100524})  & \underline{\textbf{\metric{1.000000}}} (\ci{1.000000} - \ci{1.000000}) & \metric{0.220613} (\ci{0.200201} - \ci{0.236775})  & \underline{\textbf{\metric{1.000000}}} (\ci{1.000000} - \ci{1.000000}) \\

\bottomrule

\end{tabular}
}
\end{table*}

While Table~\ref{tab:combined_performance_metrics} summarizes diagnosis performance together with calibration and output reliability, Table~\ref{tab:f1_score_per_output_field_phase1} provides complementary information by reporting field-wise F1 (with abstention penalty) for each structured output field in the reporting schema. These fields include primary diagnosis, detailed diagnosis, imaging modality, specialized MRI sequence, and imaging plane. It should be emphasized that a clear hierarchy of difficulty emerges among these targets. Most models achieve highest performance on metadata extraction tasks such as modality and anatomical plane recognition, while primary diagnostic classification remains more challenging. Diagnostic subtype prediction represents the most difficult target.

Top-performing multimodal models—including Gemini~2.5~Pro, Gemini~2.5~Flash, Gemini~2.0~Flash, GPT-5~Chat, GPT-4o, and GPT-4.1—achieve the strongest performance across both diagnosis and detailed diagnosis fields, while maintaining near-perfect metadata recognition. Medical-specialized models such as MedGemma achieve near-perfect modality recognition and, in most cases, strong plane recognition, except for MedGemma~1.5~4B. However, these models show comparatively weaker performance on diagnosis and detailed diagnosis tasks, particularly MedGemma~1~27B. MedGemma~1.5~4B performs slightly better than MedGemma~1~4B on diagnosis, but remains clearly limited on detailed diagnosis and shows comparatively poor performance on plane recognition, indicating uneven field-level robustness. This suggests that domain-specific medical pretraining alone does not necessarily lead to robust diagnostic reasoning under structured multimodal prompting.

The failure pattern observed in the overall performance metrics persists in the field-wise analysis for Meta LLaMA~3.2 Vision-Instruct variants as well. They exhibit near-zero abstention-aware F1 scores for both diagnosis and detailed diagnosis prediction, together with degraded performance even on metadata targets. This is the rationale for their exclusion from subsequent phases. In contrast, LLaMA~4~Maverick shows partially promising results, achieving moderate diagnostic performance alongside strong metadata recognition, and therefore it is justified to undergo further evaluation.

Among the remaining models, performance varies across output fields. GPT-5~Mini remains relatively competitive on diagnosis and detailed diagnosis and is particularly strong on specialized sequence recognition, with results close to or exceeding those of LLaMA~4~Maverick. Qwen~2.5-VL~32B shows strong metadata extraction for modality and plane, but weaker performance on specialized sequence prediction as well as on diagnosis and detailed diagnosis tasks.

Across all evaluated models in Table~\ref{tab:f1_score_per_output_field_phase1}, detailed diagnosis prediction yields the lowest performance scores and the greatest variability between models, indicating that predicting the specific disease subtype is the most challenging structured output task. In contrast, modality and plane recognition are nearly perfect for many models, whereas specialized sequence prediction is intermediate in difficulty. These observations further support the use of field-wise evaluation, as aggregate diagnostic performance alone would not fully capture the variable strengths and limitations of current multimodal models. Additionally, it should be emphasized that this distribution suggests that many models can already solve metadata extraction (modality and plane recognition), but true clinical interpretation (diagnostic reasoning, particularly detailed disease subtype) remains much harder.

Table~\ref{tab:efficiency_cost} further demonstrates the operational information. Models with comparable Macro-F1 vary widely in token usage, latency, and estimated cost per request, emphasizing their importance for practical deployment decisions. The estimated input and output cost is calculated based on the number of tokens generated when sending the input (prompt + image) and the returned output. 

Table~\ref{tab:efficiency_cost} further illustrates the operational characteristics of the evaluated models. Although several models achieve comparable discriminative performance, they differ drastically in token usage, latency, and estimated cost per request. For example, Gemini~2.5~Pro achieves one of the highest diagnostic performances but requires significantly more output tokens and higher inference cost compared with lighter variants such as Gemini~2.5~Flash. Similarly, GPT-5~Chat demonstrates strong diagnostic performance while maintaining moderate token usage and latency, whereas GPT-5~Mini generates considerably larger outputs, resulting in higher total token usage despite lower diagnostic performance. On the other side, models such as LLaMA~4~Maverick exhibit moderate diagnostic performance but low latency and operational cost. These examples illustrate that models with similar diagnostic capability can differ significantly in operational efficiency, highlighting the practical trade-offs that must be considered for real-world clinical usage.

\begin{table*}[ht!]
\centering
\caption{
Computational efficiency and economic cost per inference across evaluated models. 
Metrics include average input tokens, output tokens,  total tokens, latency (ms), input cost, output cost, and total cost. All values reflect per-request averages under the standardized evaluation protocol.}
\label{tab:efficiency_cost}
\resizebox{\textwidth}{!}{%
\begin{threeparttable}
\begin{tabular}{lrrrrrrr}
\toprule
\makecell[c]{Model}
& \makecell[c]{Avg In. \\ Tokens}
& \makecell[c]{Avg Out. \\ Tokens}
& \makecell[c]{Total Tokens}
& \makecell[c]{Avg Latency \\ (ms)}
& \makecell[c]{Avg In. \\ Cost (est.)\tnote{**}}
& \makecell[c]{Avg Out. \\ Cost (est.)\tnote{**}}
& \makecell[c]{Avg Cost} \\
\midrule

\texttt{\iffalse bedrock/ \fi Amazon~Nova~Lite}
& 2435.07 & 104.69 & 2539.75 & 6162.08
& \$0.000146 & \$0.000025 & \$0.000171 \\
\texttt{\iffalse bedrock/ \fi Amazon~Nova~Pro}
& 2435.07 & 97.29 & 2532.36 & 9945.20
& \$0.001948 & \$0.000311 & \$0.002259 \\
\hline
\texttt{\iffalse google/ \fi Gemini~2.0~Flash}
& 2157.14 & 100.64 & 2257.78 & 716.31
& \$0.000216 & \$0.000040 & \$0.000256 \\
\texttt{\iffalse google/ \fi Gemini~2.5~Flash}
& 1190.59 & 97.18 & 1287.77 & 1708.56
& \$0.000357 & \$0.000243 & \$0.000593 \\
\texttt{\iffalse google/ \fi Gemini~2.5~Pro}
& 1237.99 & 1096.03 & 2334.02 & 12084.35
& \$0.001547 & \$0.010960 & \$0.012507 \\
\texttt{\iffalse google/ \fi Gemma~3~27B~IT}
& 1199.61 & 103.64 & 1303.25 & 2120.09
& \$0.000078 & \$0.000027 & \$0.000148 \\
\texttt{\iffalse google/ \fi MedGemma~1~27B\tnote{*}}  
& 1300.00 & 100.00 & 1400.00 & N/A
& \$0.000025 & \$0.000002 & \$0.000013 \\
\texttt{\iffalse google/ \fi MedGemma~1~4B\tnote{*}}
& 1300.00 & 100.00 & 1400.00 & N/A
& \$0.000014 & \$0.000007 & \$0.000008 \\
\texttt{\iffalse google/ \fi MedGemma~1.5~4B\tnote{*}}
& 1300.00 & 900.00 & 2200.00  & N/A
& \$0.000215 & \$0.000142 & \$0.000178 \\

\hline
\texttt{\iffalse meta-llama/ \fi LLaMA~3.2~11B~Vision~Instruct}
& 3550.36 & 211.98 & 3762.34 & 10361.30
& \$0.000174 & \$0.000010 & \$0.000212 \\
\texttt{\iffalse meta-llama/ \fi LLaMA~3.2~90B~Vision~Instruct}
& 3838.94 & 196.43 & 4035.37 & 7441.35
& \$0.001344 & \$0.000079 & \$0.004809 \\
\texttt{\iffalse meta-llama/ \fi LLaMA~4~Maverick}
& 1620.81 & 104.27 & 1725.08 & 1092.52
& \$0.000243 & \$0.000063 & \$0.000392 \\
\hline
\texttt{\iffalse openai/ \fi GPT-4.1}
& 1385.06 & 86.33 & 1471.39 & 1692.69
& \$0.006925 & \$0.001295 & \$0.003031 \\
\texttt{\iffalse openai/ \fi GPT-4o}
& 1385.08 & 86.68 & 1471.76 & 2062.94
& \$0.003463 & \$0.000867 & \$0.003971 \\
\texttt{\iffalse openai/ \fi GPT-4o~Mini}
& 14646.53 & 82.41 & 14728.94 & 1344.00
& \$0.002197 & \$0.000049 & \$0.002237 \\
\texttt{\iffalse openai/ \fi GPT-5~Chat}
& 1313.35 & 88.68 & 1402.03 & 2208.60
& \$0.001642 & \$0.000887 & \$0.002239 \\
\texttt{\iffalse openai/ \fi GPT-5~Mini}
& 1276.81 & 763.64 & 2040.45 & 13613.40
& \$0.000319 & \$0.001527 & \$0.001813 \\
\hline
\texttt{\iffalse anthropic/ \fi Claude~Sonnet~4.5}
& 1717.80 & 105.66 & 1823.46 & 1118.24
& \$0.005153 & \$0.001585 & \$0.006738 \\
\hline
\texttt{\iffalse x-ai/ \fi Grok~4}
& 2034.45 & 759.63 & 2794.08 & 18814.69
& \$0.006103 & \$0.011394 & \$0.013554 \\
\hline
\texttt{\iffalse alibaba/ \fi Qwen~2.5-VL~32B~Instruct}
& 1542.90 & 96.67 & 1639.56 & 4512 
& \$0.000015 & \$0.00022 & \$0.000235 \\



\bottomrule
\end{tabular}
 \begin{tablenotes}
  \item[*]MedGemma was deployed within a Google Colab Pro environment. Due to the dynamic nature of resource allocation, latency and computational costs may exhibit variance between execution runs.
  \item[**] Average input and output costs are computed from the per-request input and output costs under the pricing schedule active at the time of benchmarking.
  \end{tablenotes}
  \end{threeparttable}
}

\end{table*}


Although DeepSeek-R1:70B was initially considered for evaluation, it exhibited critical failures during preliminary testing. The model frequently failed to follow the instruction to produce outputs in the required JSON format and generated hallucinated interpretations. For example, the model described brain MRI scans with stroke pathology as chest imaging with pneumonia. In addition, the model sometimes misinterpreted the task itself, producing unrelated workflow suggestions and non-existent repository links. Due to these reliability failures and lack of adherence to the structured output protocol, the model was excluded from the formal benchmark evaluation.

Based on the combined evidence from discriminative performance, structured output reliability, calibration behavior, and operational characteristics, a reduced set of candidate models was selected for Phase~2 stability validation. Model selection was not determined solely by Phase~1 ranking, because several models exhibit partially overlapping confidence intervals and comparable performance within the uncertainty of the screening subset. Instead, selection followed three complementary criteria: (1) inclusion of the strongest-performing frontier multimodal models identified in Phase~1, (2) inclusion of models with intermediate diagnostic performance but heterogeneous strengths across structured output fields or operational characteristics, and (3) representation of diverse architectural families, including proprietary frontier models, open-weight multimodal systems, and medically pretrained models, while reducing redundancy among closely related variants. This strategy ensures that Phase~2 evaluates not only the leading models from the screening stage but also representative systems with different modeling paradigms and operational profiles. The resulting candidate set for Phase~2 consists of Gemini~2.0~Flash, Gemini~2.5~Flash, Gemini~2.5~Pro, MedGemma~1~4B, MedGemma~1.5~4B, LLaMA~4~Maverick, GPT-4o, GPT-4.1, GPT-5~Chat, GPT-5~Mini, Grok~4, and Qwen~2.5-VL~32B~Instruct.


\subsection{Phase 2 Results: Stability validation}

Phase~2 evaluates model stability and scalability on a larger development split comprising 45\% of the benchmark dataset, with the goal of assessing whether the trends observed during the screening stage persist at a larger scale. Table~\ref{tab:model_metrics_dev_split} summarizes model-level performance across discriminative performance, calibration quality, and structured-output reliability metrics.

\begin{table*}[!ht]
\centering
\caption{Phase 2 (stability validation) -  
Model performance on the primary diagnostic task on the basis of abstention penalized discriminative performance metrics (Macro-F1, weighted Macro-F1, Micro-F1, Accuracy), calibration metrics (ECE and Brier score), and structured output reliability (valid JSON rate and abstention rate). 
Values in parentheses denote 95\% confidence intervals. 
Bold underlined values indicate the best point estimate per column.}
\label{tab:model_metrics_dev_split}
\resizebox{\textwidth}{!}{%
\begin{threeparttable}
\begin{tabular}{l | c c c c c c c c}
\toprule
\textbf{Model} & \textbf{Macro-F1} & \textbf{Macro-F1-weighted} & \textbf{Micro-F1} & \textbf{Accuracy} & \textbf{ECE} & \textbf{Brier} & \textbf{Valid JSON (\%)} & \textbf{Abstention Rate (\%)} \\
\midrule

\texttt{Gemini~2.0~Flash} 
 & \metric{0.526876} (\ci{0.517764}--\ci{0.535468})
 & \metric{0.703259} (\ci{0.693847}--\ci{0.711630})
 & \metric{0.702572} (\ci{0.693491}--\ci{0.710126})
 & \metric{0.595323} (\ci{0.578274}--\ci{0.615156})
 & \underline{\textbf{\metric{0.141067}}}
 & \underline{\textbf{\metric{0.204196}}}
 & \metric{99.99}
 & \metric{0.007532389} \\

\texttt{Gemini~2.5~Flash} 
 & \metric{0.549482} (\ci{0.539249}--\ci{0.558867})
 & \metric{0.726365} (\ci{0.720278}--\ci{0.733964})
 & \metric{0.715556} (\ci{0.707780}--\ci{0.721971})
 & \metric{0.604645} (\ci{0.585461}--\ci{0.624355})
 & \metric{0.235836}
 & \metric{0.249878}
 & \metric{99.99}
 & \metric{0.015064779} \\

\texttt{Gemini~2.5~Pro} 
 & \underline{\textbf{\metric{0.572630}}} (\ci{0.561292}--\ci{0.582589})
 & \underline{\textbf{\metric{0.742264}}} (\ci{0.733512}--\ci{0.749708})
 & \metric{0.741059} (\ci{0.734669}--\ci{0.748405})
 & \underline{\textbf{\metric{0.606472}}} (\ci{0.585268}--\ci{0.623530})
 & \metric{0.225877}
 & \metric{0.239935}
 & \metric{99.84}
 & \metric{0.030175015} \\

\texttt{MedGemma~1~4B} 
 & \metric{0.382149} (\ci{0.375870}--\ci{0.389581})
 & \metric{0.560298} (\ci{0.552317}--\ci{0.568108})
 & \metric{0.589024} (\ci{0.582845}--\ci{0.595436})
 & \metric{0.390705} (\ci{0.383006}--\ci{0.397545})
 & \metric{0.340421}
 & \metric{0.349931}
 & \underline{\textbf{\metric{100}}}
 & \underline{\textbf{\metric{0}}} \\

\texttt{MedGemma~1.5~4B} 
 & \metric{0.391436} (\ci{0.382782}--\ci{0.398700})
 & \metric{0.575142} (\ci{0.567013}--\ci{0.583442})
 & \metric{0.626659} (\ci{0.618534}--\ci{0.636394})
 & \metric{0.429966} (\ci{0.418509}--\ci{0.441206})
 & \metric{0.314875}
 & \metric{0.316461}
 & \metric{99.34}
 & \underline{\textbf{\metric{0}}} \\

\hline


\texttt{LLaMA~4~Maverick} 
 & \metric{0.451750} (\ci{0.442477}--\ci{0.461160})
 & \metric{0.649502} (\ci{0.641486}--\ci{0.660500})
 & \metric{0.648933} (\ci{0.640434}--\ci{0.657732})
 & \metric{0.476909} (\ci{0.457741}--\ci{0.495820})
 & \metric{0.252955}
 & \metric{0.279326}
 & \metric{99.80}
 & \metric{1.962264151} \\

\hline

\texttt{GPT-4.1} 
 & \metric{0.510509} (\ci{0.495583}--\ci{0.527141})
 & \metric{0.705854} (\ci{0.698204}--\ci{0.713776})
 & \metric{0.701436} (\ci{0.694194}--\ci{0.708219})
 & \metric{0.506774} (\ci{0.490425}--\ci{0.527607})
 & \metric{0.261381}
 & \metric{0.277316}
 & \metric{99.95}
 & \metric{0.022607385} \\

\texttt{GPT-4o} 
 & \metric{0.516679} (\ci{0.508341}--\ci{0.528697})
 & \metric{0.714159} (\ci{0.707506}--\ci{0.722295})
 & \metric{0.713009} (\ci{0.706913}--\ci{0.719153})
 & \metric{0.515116} (\ci{0.506334}--\ci{0.527067})
 & \metric{0.220225}
 & \metric{0.249430}
 & \underline{\textbf{\metric{100}}}
 & \metric{0.3088047} \\

\texttt{GPT-5~Chat} 
 & \metric{0.557615} (\ci{0.543064}--\ci{0.569208})
 & \metric{0.735266} (\ci{0.726902}--\ci{0.742093})
 & \underline{\textbf{\metric{0.741813}}} (\ci{0.734846}--\ci{0.748080})
 & \metric{0.576892} (\ci{0.564782}--\ci{0.592722})
 & \metric{0.185023}
 & \metric{0.217835}
 & \metric{99.93}
 & \metric{0.007536931} \\

\texttt{GPT-5~Mini} 
 & \metric{0.469601} (\ci{0.460405}--\ci{0.477618})
 & \metric{0.672153} (\ci{0.664267}--\ci{0.679794})
 & \metric{0.622004} (\ci{0.615028}--\ci{0.629414})
 & \metric{0.511461} (\ci{0.487038}--\ci{0.531491})
 & \metric{0.238706}
 & \metric{0.282523}
 & \metric{99.80}
 & \metric{0.0377358490566037} \\

\hline

\texttt{Grok~4} 
 & \metric{0.465600} (\ci{0.458417}--\ci{0.475143})
 & \metric{0.687108} (\ci{0.679167}--\ci{0.693601})
 & \metric{0.690691} (\ci{0.683128}--\ci{0.699894})
 & \metric{0.462769} (\ci{0.453342}--\ci{0.469980})
 & \metric{0.256192}
 & \metric{0.282309}
 & \metric{92.49}
 & \metric{0.016286645} \\

\hline

\texttt{Qwen~2.5-VL~32B~Instruct} 
 & \metric{0.360398} (\ci{0.355020}--\ci{0.366250})
 & \metric{0.579403} (\ci{0.572761}--\ci{0.587039})
 & \metric{0.578229} (\ci{0.571411}--\ci{0.584353})
 & \metric{0.366771} (\ci{0.361584}--\ci{0.371237})
 & \metric{0.340533}
 & \metric{0.364606}
 & \underline{\textbf{\metric{100}}}
 & \metric{0.009684292} \\

\bottomrule
\end{tabular}
\end{threeparttable}
}
\end{table*}

Phase~2 results remain broadly consistent with the screening-stage ranking emerged in Phase~1. Gemini~2.5~Pro achieves the highest abstention-penalized Macro-F1, weighted Macro-F1, and Accuracy, while GPT-5~Chat achieves the highest Micro-F1. Gemini~2.5~Flash and Gemini~2.0~Flash also remain among the strongest-performing models, confirming that the leading candidate set identified in Phase~1 remains competitive when evaluated on the larger development split. The confidence intervals for the best performing models are relatively narrow, supporting the stability of the observed ranking patterns at this stage.

Calibration metrics reveal additional distinctions not captured by discriminative performance alone. Gemini~2.0~Flash achieves the lowest ECE and Brier score, indicating the most reliable confidence calibration among the evaluated models. GPT-5~Chat also demonstrates strong calibration while maintaining high discriminative performance. Several other models such as Gemini~2.5~Pro, Gemini~2.5~Flash, GPT-4o, and GPT-4.1—exhibit higher ECE and Brier scores, indicating slightly less reliable probability calibration. These results further support the inclusion of calibration in the benchmark, as similar diagnostic performance does not necessarily imply equally reliable confidence estimates.

Structured output compliance remains very high for most models, typically close to or above 99.8\% valid JSON. A notable exception is Grok~4, which achieves only 92.49\% valid JSON responses, indicating a potential deployment limitation despite moderate diagnostic performance. Abstention rates remain very low across the strongest proprietary models, but differences are still informative when evaluation is performed at larger scale. In particular, models such as MedGemma~1~4B and Qwen~2.5-VL~32B~Instruct achieve near-zero abstention, yet remain significantly weaker on Macro-F1, showing that low abstention alone does not translate into stronger diagnostic performance. This observation further supports the use of abstention-penalized Macro-F1 as the primary metric for balanced clinical evaluation.

\begin{table*}[ht!]
\centering
\caption{Phase 2 (stability validation) field-wise performance (F1 with abstention penalty) across structured output fields: primary diagnosis, detailed diagnosis, imaging modality, specialized MRI sequence (MRI-only), and imaging plane (when annotated).
Values in parentheses denote 95\% confidence intervals.}

\label{tab:f1_score_per_output_field_phase2}
\resizebox{\textwidth}{!}{%
\begin{tabular}{lccccc}
\toprule
\makecell[c]{Model} &
\makecell[c]{Diagnosis} &
\makecell[c]{Detailed\\diagnosis} &
\makecell[c]{Modality} &
\makecell[c]{Specialized\\sequence} &
\makecell[c]{Plane} \\
\midrule

\texttt{Gemini~2.0~Flash} &
\makecell{\metric{0.526876} {(\ci{0.517760} - \ci{0.538607})}} &
\makecell{\metric{0.225503} {(\ci{0.218818} - \ci{0.234093})}} &
\makecell{\metric{0.999149} {(\ci{0.998721} - \ci{0.999678})}} &
\makecell{\underline{\textbf{\metric{0.856756}}} {(\ci{0.848492} - \ci{0.864091})}} &
\makecell{\metric{0.987911} {(\ci{0.984357} - \ci{0.991396})}} \\

\texttt{Gemini~2.5~Flash} &
\makecell{\metric{0.549482} {(\ci{0.538719} - \ci{0.561245})}} &
\makecell{\metric{0.254855} {(\ci{0.240930} - \ci{0.265130})}} &
\makecell{\metric{0.998767} {(\ci{0.998295} - \ci{0.999282})}} &
\makecell{\metric{0.783426} {(\ci{0.776044} - \ci{0.790260})}} &
\makecell{\metric{0.976856} {(\ci{0.970304} - \ci{0.981130})}} \\

\texttt{Gemini~2.5~Pro} &
\makecell{\underline{\textbf{\metric{0.572630}}} {(\ci{0.565093} - \ci{0.582741})}} &
\makecell{\underline{\textbf{\metric{0.317199}}} {(\ci{0.302265} - \ci{0.327772})}} &
\makecell{\metric{0.999117} {(\ci{0.998595} - \ci{0.999598})}} &
\makecell{\metric{0.785220} {(\ci{0.776886} - \ci{0.795365})}} &
\makecell{\metric{0.967637} {(\ci{0.961826} - \ci{0.972394})}} \\

\texttt{MedGemma~1~4B} &
\makecell{\metric{0.382149} {(\ci{0.375870} - \ci{0.389581})}} &
\makecell{\metric{0.104548} {(\ci{0.098274} - \ci{0.112408})}} &
\makecell{\underline{\textbf{\metric{1.000000}}} {(\ci{1.000000} - \ci{1.000000})}} &
\makecell{\metric{0.172935} {(\ci{0.165919} - \ci{0.180135})}} &
\makecell{\underline{\textbf{\metric{1.000000}}} {(\ci{1.000000} - \ci{1.000000})}} \\

\texttt{MedGemma~1.5~4B} &
\makecell{\metric{0.391436} {(\ci{0.382782} - \ci{0.398700})}} &
\makecell{\metric{0.071375} {(\ci{0.062771} - \ci{0.079284})}} &
\makecell{\metric{0.951620} {(\ci{0.948323} - \ci{0.954847})}} &
\makecell{\metric{0.238126} {(\ci{0.228457} - \ci{0.245943})}} &
\makecell{\metric{0.497918} {(\ci{0.484018} - \ci{0.509302})}} \\

\hline

\texttt{LLaMA~4~Maverick} &
\makecell{\metric{0.451750} {(\ci{0.444314} - \ci{0.460447})}} &
\makecell{\metric{0.230175} {(\ci{0.217764} - \ci{0.243939})}} &
\makecell{\metric{0.989329} {(\ci{0.987690} - \ci{0.990855})}} &
\makecell{\metric{0.624250} {(\ci{0.613660} - \ci{0.634017})}} &
\makecell{\metric{0.980946} {(\ci{0.976507} - \ci{0.984497})}} \\

\hline

\texttt{GPT-4.1} &
\makecell{\metric{0.510509} {(\ci{0.497690} - \ci{0.523057})}} &
\makecell{\metric{0.185560} {(\ci{0.177726} - \ci{0.191960})}} &
\makecell{\metric{0.998287} {(\ci{0.997671} - \ci{0.998942})}} &
\makecell{\metric{0.752823} {(\ci{0.744118} - \ci{0.760349})}} &
\makecell{\metric{0.979963} {(\ci{0.974423} - \ci{0.986074})}} \\

\texttt{GPT-4o} &
\makecell{\metric{0.516679} {(\ci{0.508115} - \ci{0.530321})}} &
\makecell{\metric{0.199078} {(\ci{0.190411} - \ci{0.208179})}} &
\makecell{\metric{0.997196} {(\ci{0.996013} - \ci{0.997824})}} &
\makecell{\metric{0.756658} {(\ci{0.747005} - \ci{0.765691})}} &
\makecell{\metric{0.989803} {(\ci{0.987228} - \ci{0.993316})}} \\

\texttt{GPT-5~Chat} &
\makecell{\metric{0.557615} {(\ci{0.544685} - \ci{0.573805})}} &
\makecell{\metric{0.235143} {(\ci{0.226779} - \ci{0.242412})}} &
\makecell{\metric{0.997515} {(\ci{0.996786} - \ci{0.998199})}} &
\makecell{\metric{0.806697} {(\ci{0.797112} - \ci{0.813815})}} &
\makecell{\metric{0.988251} {(\ci{0.984764} - \ci{0.991581})}} \\

\texttt{GPT-5~Mini} &
\makecell{\metric{0.469601} {(\ci{0.460440} - \ci{0.476932})}} &
\makecell{\metric{0.292969} {(\ci{0.279363} - \ci{0.305757})}} &
\makecell{\metric{0.996870} {(\ci{0.996073} - \ci{0.997874})}} &
\makecell{\metric{0.802415} {(\ci{0.795176} - \ci{0.810492})}} &
\makecell{\metric{0.992110} {(\ci{0.989766} - \ci{0.994905})}} \\

\hline

\texttt{Grok~4} &
\makecell{\metric{0.465600} {(\ci{0.455955} - \ci{0.477044})}} &
\makecell{\metric{0.196368} {(\ci{0.185716} - \ci{0.204451})}} &
\makecell{\metric{0.998709} {(\ci{0.997971} - \ci{0.999208})}} &
\makecell{\metric{0.500477} {(\ci{0.491407} - \ci{0.511614})}} &
\makecell{\metric{0.998576} {(\ci{0.997597} - \ci{0.999427})}} \\

\hline

\texttt{Qwen~2.5-VL~32B~Instruct} &
\makecell{\metric{0.360398} {(\ci{0.355910} - \ci{0.365507})}} &
\makecell{\metric{0.105859} {(\ci{0.101681} - \ci{0.110715})}} &
\makecell{\underline{\textbf{\metric{1.000000}}} {(\ci{1.000000} - \ci{1.000000})}} &
\makecell{\metric{0.239574} {(\ci{0.232087} - \ci{0.246134})}} &
\makecell{\underline{\textbf{\metric{1.000000}}} {(\ci{1.000000} - \ci{1.000000})}} \\

\bottomrule
\end{tabular}
}
\end{table*}

Table~\ref{tab:f1_score_per_output_field_phase2} reports abstention-penalized Macro-F1 scores decomposed by output field on the 45\% development split, enabling a detailed assessment of model capabilities across the different structured prediction targets. Consistent with Phase~1, a clear hierarchy of difficulty emerges. Most models achieve near-perfect performance on imaging metadata extraction tasks, particularly modality and anatomical plane recognition. In contrast, primary diagnostic classification remains substantially more challenging, and detailed diagnosis (subtype) prediction emerges as the most difficult task across all evaluated models.

Among the strongest proprietary models, complementary strengths can be observed across prediction targets. Gemini~2.5~Pro achieves the highest performance for primary diagnostic classification and also the strongest results for detailed diagnosis prediction, indicating stronger diagnostic reasoning capability. Gemini~2.5~Flash and Gemini~2.0~Flash remain close competitors for primary diagnosis, while GPT-5~Chat demonstrates particularly strong performance on specialized MRI sequence recognition. These differences illustrate that high performance on metadata extraction or sequence identification does not necessarily translate into equally strong diagnostic reasoning.

A broader comparison across model families emphasizes a consistent pattern. Frontier proprietary multimodal models—including Gemini~2.5~Pro, Gemini~2.5~Flash, Gemini~2.0~Flash, GPT-5~Chat, and GPT-4o—form the leading group across diagnostic prediction tasks. In contrast, open-weight models such as LLaMA~4~Maverick and Qwen~2.5-VL~32B~Instruct achieve moderate diagnostic performance while maintaining strong metadata recognition, highlighting their value as adaptable research baselines rather than state-of-the-art diagnostic systems. Medically pretrained models (MedGemma~1~4B and MedGemma~1.5~4B) exhibit near-perfect modality recognition and generally strong plane recognition, although MedGemma~1.5~4B shows a notable degradation on the plane prediction task. Despite strong metadata extraction, both models achieve considerably weaker performance on diagnostic and subtype prediction. This pattern suggests that current domain-specific medical pretraining improves structured metadata extraction but does not yet match the diagnostic reasoning capabilities of frontier proprietary multimodal models under standardized prompting conditions.

\begin{figure}[!htbp]
\centering
\includegraphics[width=0.9\textwidth]{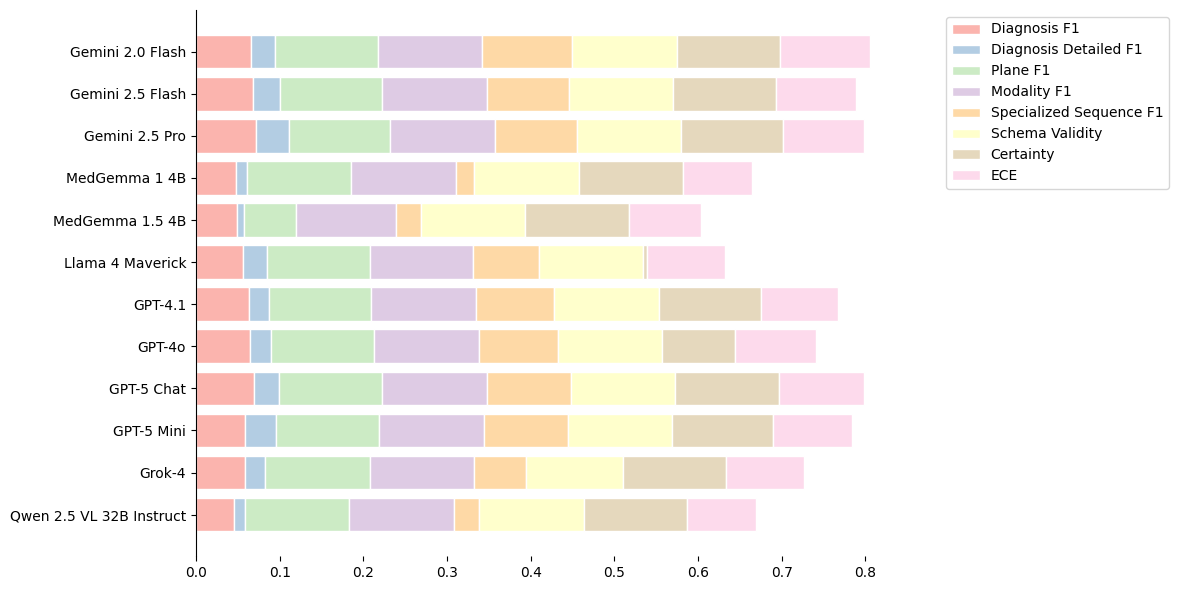}
\caption{Illustrative multi-dimensional summary of selected output fields and evaluation dimensions on the 45\% development split, including diagnosis, detailed diagnosis, metadata recognition, schema validity, and calibration-related behavior.}
\label{fig:general_stacked_phase2}
\end{figure}

Fig.~\ref{fig:general_stacked_phase2} provides a complementary visual overview by aggregating several evaluation dimensions—including diagnosis, detailed diagnosis, metadata recognition, schema validity, and calibration-related behavior—into a unified multi-dimensional summary. Ignoring the excluded LLaMA~3.2 Vision-Instruct baseline, the figure illustrates that the strongest models combine balanced performance across diagnostic reasoning, metadata extraction, and structured-output reliability. In particular, Gemini~2.5~Pro, Gemini~2.5~Flash, Gemini~2.0~Flash, GPT-5~Chat, and GPT-4o exhibit the most balanced overall profiles, although their strengths differ by dimension. Gemini~2.0~Flash stands out through particularly strong calibration-related behavior, whereas GPT-5~Chat and Gemini~2.5~Pro maintain stronger aggregate diagnostic profiles.

The figure also highlights an important limitation shared by several open-weight and medically pretrained models. Systems such as Qwen~2.5-VL~32B~Instruct, MedGemma~1~4B, and MedGemma~1.5~4B show strong contributions from metadata recognition, schema validity, and certainty-related components, yet remain significantly weaker on diagnostic classification and especially detailed diagnosis prediction. This pattern reinforces the central finding of the benchmark: current multimodal models differ far less in basic imaging metadata recognition than in clinically meaningful diagnostic reasoning, particularly when fine-grained subtype discrimination is required.

Among the open-weight systems, LLaMA~4~Maverick demonstrates a clear improvement over earlier LLaMA vision variants, achieving moderate diagnostic performance while remaining below the strongest proprietary models. MedGemma~1~4B, MedGemma~1.5~4B, and Qwen~2.5-VL~32B~Instruct maintain strong structured-output and metadata recognition capabilities but significantly weaker diagnostic reasoning. These models are therefore retained primarily as candidates for future domain-specific adaptation rather than as competitive out-of-the-box diagnostic systems. Overall, the field-wise analysis confirms that model differentiation at scale is driven primarily by diagnostic reasoning capability rather than metadata awareness, reinforcing the importance of structured multi-target evaluation for clinically meaningful benchmarking.

The final set of models for Phase~3 was selected to balance frontier performance, research extensibility, and clinical relevance. In addition to the strongest proprietary models (Gemini~2.5~Pro, Gemini~2.5~Flash, GPT-5~Chat, and GPT-4o), we retained Gemini~2.0~Flash as an efficiency-oriented variant that demonstrates competitive diagnostic performance together with particularly strong calibration behavior. 

To ensure coverage of open-weight architectures, three additional models were included. MedGemma~1~4B and MedGemma~1.5~4B were retained as a medically pretrained multimodal model with strong structured-output behavior and potential for domain-specific fine-tuning. Qwen~2.5-VL~32B~Instruct was included as a general-purpose open-weight multimodal model suitable for future adaptation and comparative research. In addition, Meta’s LLaMA~4~Maverick was retained as a frontier-scale open-weight model that demonstrates substantial improvement over earlier LLaMA vision variants while preserving adaptability and local deployment capabilities. Thus, for Phase~3 we end up with nine models selected to represent complementary trade-offs across diagnostic performance, calibration reliability, operational efficiency, and accessibility.

Retaining GPT-4o over the newer GPT-4.1 is justified by its more favorable performance–efficiency trade-off for the structured output evaluated in this benchmark. Although GPT-4.1 represents a newer architecture, GPT-4o demonstrates slightly stronger diagnostic performance in both the development-split evaluation and the Phase-1 class-level analysis given in appendix (Table A.2.~\ref{tab:f1_per_class_phase1}). In particular, GPT-4o achieves higher per-class F1 scores for several clinically important categories, including tumor (0.880 vs. 0.845), stroke (0.681 vs. 0.589), and multiple sclerosis (0.417 vs. 0.356). At the same time, GPT-4o maintains perfect structured-output reliability (100\% valid JSON) and exhibits substantially lower operational cost in our evaluation pipeline, with roughly 50\% lower average input cost and about one-third lower output cost per request. Together, these characteristics place GPT-4o closer to the practical usability, combining strong diagnostic performance with reliable structured output and lower inference cost.

GPT-5 Mini and Grok-4 were excluded due to unfavorable operational and diagnostic trade-offs relative to the retained models. GPT-5 Mini, although representing a smaller efficiency-oriented variant of the GPT-5 architecture, generates significantly longer responses in our benchmark, which increases latency and compute cost without corresponding gains in diagnostic performance. Phase-1 class-level analysis (Table A.2.~\ref{tab:f1_per_class_phase1}) further indicates weaker class discrimination compared with GPT-5 Chat, particularly for stroke detection (0.417 vs. 0.733). Grok-4, while achieving competitive performance for tumor recognition (0.888), exhibits substantially lower structured-output reliability and fails to detect the heterogeneous Other abnormalities class at all.

\subsection{Phase 3 Results: Generalization
Benchmark - Final Evaluation with Zero-shot and Few-shot Prompting}

Phase~3 represents the final held-out evaluation stage of the benchmark. In this phase, the strongest models selected through Phases~1 and~2 are evaluated un   der both zero-shot and few-shot prompting settings. The key objective of this phase is to identify and highlight which multimodal large language models provide the most favorable balance between diagnostic accuracy, reliability of structured outputs, calibration of confidence estimates, and practical operational constraints such as cost, latency, and token efficiency.

Table~\ref{tab:final_zero_shot} summarizes model performance
on the primary diagnostic task, whereas Table~\ref{tab:efficiency_cost_phase3} highlights large differences in computational efficiency across models. Under zero-shot prompting, Gemini~2.5~Pro demonstrates the strongest diagnostic performance, achieving the highest Macro-F1, weighted Macro-F1, and Micro-F1 among all evaluated models. However, Gemini~2.5~Pro is related to the highest computational cost and output token usage, as well as relatively high average input and output cost. GPT-5~Chat follows closely, with slightly lower Macro-F1 but comparable Micro-F1 and strong calibration characteristics (ECE = 0.186, Brier = 0.219), indicating reliable confidence estimation while maintaining competitive diagnostic performance. GPT-5~Chat achieves this at significantly lower latency and output cost, comparable input tokens, but the highest input cost and relatively high output cost.

It should be noted that, Gemini~2.0~Flash achieves the highest overall accuracy and the strongest calibration metrics (lowest ECE and Brier score). Additionally, it achieves the lowest latency and overall inference cost. Although its Macro-F1 remains slightly below Gemini~2.5~Pro and Gemini~2.5~Flash in terms of discriminative performance (except for the accuracy), this model exhibits very strong overall efficiency–reliability profile under zero-shot prompting, combining strong diagnostic performance with perfect structured-output reliability, minimal abstention and computational efficiency. Gemini~2.5~Flash provides a favorable balance between diagnostic performance and computational efficiency. Its Macro-F1 approaches the top-performing models while maintaining significantly lower operational cost and latency compared with Gemini~2.5~Pro. This combination positions Gemini~2.5~Flash as a practical candidate for high-throughput clinical settings where cost and inference speed are critical constraints.

\begin{table*}[htbp]
\centering
\caption{Phase 3 (final evaluation under \textbf{zero-shot} prompting) - Model performance on the primary diagnostic task evaluated for abstentions penalized discriminative performance metrics (Macro-F1, weighted Macro-F1, Micro-F1, Accuracy, AUC), calibration (ECE and Brier score), and structured output reliability (valid JSON rate, abstention rate). Values in parentheses denote 95\% confidence intervals. Bold underlined values indicate the best observed point estimate per column.}
\label{tab:final_zero_shot}
\resizebox{\textwidth}{!}{%
\begin{threeparttable}
\begin{tabular}{l | c c c c c c c c c}
\toprule
\textbf{Model} & \textbf{Macro-F1} & \textbf{Macro-F1-weighted} & \textbf{Micro-F1} & \textbf{Accuracy} & \textbf{AUC} & \textbf{ECE} & \textbf{Brier} & \textbf{Valid JSON (\%)} & \textbf{Abstention Rate (\%)} \\
\midrule

\texttt{Gemini~2.0~Flash}
 & \metric{0.537340} (\ci{0.524054}--\ci{0.552802})
 & \metric{0.707602} (\ci{0.696702}--\ci{0.722090})
 & \metric{0.705579} (\ci{0.694348}--\ci{0.717336})
 & \underline{\textbf{\metric{0.618429}}} (\ci{0.594302}--\ci{0.645080})
 & \underline{\textbf{\metric{0.756031}}}
 & \underline{\textbf{\metric{0.141528}}}
 & \underline{\textbf{\metric{0.201407}}}
 & \underline{\textbf{\metric{100}}}
 & \metric{0.040667} \\

\texttt{Gemini~2.5~Flash}
 & \metric{0.563861} (\ci{0.551675}--\ci{0.577894})
 & \metric{0.731160} (\ci{0.721402}--\ci{0.740368})
 & \metric{0.720233} (\ci{0.710306}--\ci{0.730275})
 & \metric{0.616075} (\ci{0.593375}--\ci{0.644100})
 & \metric{0.609192}
 & \metric{0.231183}
 & \metric{0.246289}
 & \metric{99.96}
 & \underline{\textbf{\metric{0}}} \\

\texttt{Gemini~2.5~Pro}
 & \underline{\textbf{\metric{0.576743}}} (\ci{0.562861}--\ci{0.589617})
 & \underline{\textbf{\metric{0.741138}}} (\ci{0.731185}--\ci{0.751152})
 & \underline{\textbf{\metric{0.740826}}} (\ci{0.732587}--\ci{0.750730})
 & \metric{0.604887} (\ci{0.580062}--\ci{0.626921})
 & \metric{0.595499}
 & \metric{0.226294}
 & \metric{0.240720}
 & \metric{99.74}
 & \underline{\textbf{\metric{0}}} \\

\texttt{MedGemma~1~4B}
 & \metric{0.379716} (\ci{0.366078}--\ci{0.392120})
 & \metric{0.555574} (\ci{0.544528}--\ci{0.566576})
 & \metric{0.585197} (\ci{0.574326}--\ci{0.594493})
 & \metric{0.388109} (\ci{0.377134}--\ci{0.398588})
 & \metric{0.592396}
 & \metric{0.344750}
 & \metric{0.353848}
 & \underline{\textbf{\metric{100}}}
 & \underline{\textbf{\metric{0}}} \\

\texttt{MedGemma~1.5~4B}
 & \metric{0.418824} (\ci{0.408453}--\ci{0.429961})
 & \metric{0.602401} (\ci{0.591439}--\ci{0.614896})
 & \metric{0.635695} (\ci{0.623764}--\ci{0.645187})
 & \metric{0.458969} (\ci{0.441839}--\ci{0.473547})
 & \metric{0.565540}
 & \metric{0.318432}
 & \metric{0.321627}
 & \metric{99.50}
 & \underline{\textbf{\metric{0}}} \\

\midrule

\texttt{LLaMA~4~Maverick}
 & \metric{0.453718} (\ci{0.441742}--\ci{0.468640})
 & \metric{0.658688} (\ci{0.647990}--\ci{0.668694})
 & \metric{0.657445} (\ci{0.645098}--\ci{0.670222})
 & \metric{0.490331} (\ci{0.467494}--\ci{0.519745})
 & \metric{0.610189}
 & \metric{0.253382}
 & \metric{0.279175}
 & \underline{\textbf{\metric{100}}}
 & \metric{0.081334} \\

\midrule

\texttt{GPT-4o}
 & \metric{0.517257} (\ci{0.500316}--\ci{0.533483})
 & \metric{0.710038} (\ci{0.698535}--\ci{0.721180})
 & \metric{0.709227} (\ci{0.700536}--\ci{0.718988})
 & \metric{0.511822} (\ci{0.492861}--\ci{0.531472})
 & \metric{0.567994}
 & \metric{0.225007}
 & \metric{0.253660}
 & \metric{99.99}
 & \metric{0.162690} \\

\texttt{GPT-5~Chat}
 & \metric{0.561478} (\ci{0.540354}--\ci{0.580738})
 & \metric{0.732510} (\ci{0.720846}--\ci{0.742168})
 & \metric{0.740018} (\ci{0.731349}--\ci{0.746682})
 & \metric{0.581647} (\ci{0.560958}--\ci{0.603515})
 & \metric{0.729277}
 & \metric{0.186117}
 & \metric{0.219159}
 & \metric{99.99}
 & \metric{0.013557} \\

\midrule

\texttt{Qwen~2.5-VL~32B~Instruct}
 & \metric{0.358283} (\ci{0.350539}--\ci{0.367519})
 & \metric{0.576429} (\ci{0.564351}--\ci{0.587468})
 & \metric{0.575341} (\ci{0.566227}--\ci{0.589073})
 & \metric{0.364368} (\ci{0.357482}--\ci{0.371848})
 & \metric{0.419604}
 & \metric{0.343527}
 & \metric{0.366681}
 & \underline{\textbf{\metric{100}}}
 & \metric{0.013556} \\

\bottomrule
\end{tabular}
\end{threeparttable}
}
\end{table*}

\begin{table*}[ht!]
\centering
\caption{
Computational efficiency and economic cost per inference across evaluated models in Phase 3 (zero-shot).
Metrics include average input tokens, output tokens, total tokens, estimated input cost, output cost, and total cost.
All values reflect per-request averages under the standardized evaluation protocol.}
\label{tab:efficiency_cost_phase3}
\resizebox{\textwidth}{!}{%
\begin{threeparttable}
\begin{tabular}{lrrrrrrr}
\toprule
\makecell[c]{Model}
& \makecell[c]{Avg In. \\ Tokens}
& \makecell[c]{Avg Out. \\ Tokens}
& \makecell[c]{Total Tokens}
& \makecell[c]{Avg Latency \\ (ms)}
& \makecell[c]{Avg In. \\ Cost (est.)\tnote{**}}
& \makecell[c]{Avg Out. \\ Cost (est.)\tnote{**}}
& \makecell[c]{Avg Cost} \\
\midrule

\texttt{\iffalse google/ \fi Gemini~2.0~Flash}
& 2430.91 & 105.07 & 2535.98 & 1607.12
& \$0.000243 & \$0.000042 & \$0.000285 \\

\texttt{\iffalse google/ \fi Gemini~2.5~Flash}
& 1475.00 & 103.57 & 1578.57 & 1775.04
& \$0.000443 & \$0.000259 & \$0.000701 \\

\texttt{\iffalse google/ \fi Gemini~2.5~Pro}
& 1475.00 & 1134.59 & 2609.59 & 13423.05
& \$0.001844 & \$0.011346 & \$0.013190 \\

\texttt{\iffalse google/ \fi MedGemma~1.5~4B \tnote{*}}
& 1493.00 & 519.31 & 2012.31 & N/A
& \$0.000184 & \$0.000237 & \$0.000420 \\

\texttt{MedGemma~1~4B \tnote{*}}
& 300.00 & 1000.00 & 1300.00 & N/A
& \$0.000037 & \$0.000456 & \$0.000493 \\

\hline

\texttt{\iffalse meta-llama/ \fi LLaMA~4~Maverick}
& 1827.77 & 86.05 & 1913.82 & 937.65
& \$0.000274 & \$0.000052 & \$0.000326 \\

\hline

\texttt{\iffalse openai/ \fi GPT-4o}
& 1583.87 & 91.88 & 1675.75 & 2182.30
& \$0.003960 & \$0.000919 & \$0.004878 \\

\texttt{\iffalse openai/ \fi GPT-5~Chat}
& 1512.81 & 92.68 & 1605.49 & 1777.00
& \$0.001891 & \$0.000927 & \$0.002818 \\

\hline

\texttt{\iffalse alibaba/ \fi Qwen~2.5-VL~32B~Instruct}
& 1556.28 & 97.97 & 1654.26 & N/A
& \$0.000015 & \$0.000220 & \$0.000235 \\

\bottomrule
\end{tabular}
 \begin{tablenotes}
  \item[*]MedGemma was deployed within a Google Colab Pro environment. Due to the dynamic nature of resource allocation, latency and computational costs may exhibit variance between execution runs.
  \item[**] Average input and output costs are computed from the per-request input and output costs under the pricing schedule active at the time of benchmarking.
  \end{tablenotes}
  \end{threeparttable}
}
\end{table*}

Open-weight models demonstrate more heterogeneous behavior. LLaMA~4~Maverick achieves moderate diagnostic performance, outperforming other open-weight systems across most discriminative metrics while maintaining perfect structured-output validity. In contrast, medically pretrained MedGemma models and the general-purpose Qwen~2.5-VL~32B~Instruct show weaker diagnostic performance despite producing perfectly valid structured outputs and requiring very low computational cost. These results highlight a persistent gap between reliable structured-output generation and clinically meaningful diagnostic reasoning. 

To summarize, the zero-shot results indicate that diagnostic reasoning performance, calibration quality, and operational efficiency do not necessarily coincide within a single model, highlighting the importance of multi-dimensional benchmarking that must be considered when deploying multimodal diagnostic systems in real-world neuroimaging workflows.

\begin{table*}[htbp]
\centering
\caption{Phase 3: final evaluation under \textbf{few-shot} prompting - Model performance on the primary diagnostic task evaluated for abstentions penalized discriminative performance metrics (Macro-F1, weighted Macro-F1, Micro-F1, Accuracy, AUC), calibration (ECE and Brier score), and structured output reliability (valid JSON rate, abstention rate). Values in parentheses denote 95\% confidence intervals. Bold underlined values indicate the best observed point estimate per column.}
\label{tab:final_few_shot}
\resizebox{\textwidth}{!}{%
\begin{threeparttable}
\begin{tabular}{l | c c c c c c c c c}
\toprule
\textbf{Model} & \textbf{Macro-F1} & \textbf{Macro-F1-weighted} & \textbf{Micro-F1} & \textbf{Accuracy} & \textbf{AUC} & \textbf{ECE} & \textbf{Brier} & \textbf{Valid JSON (\%)} & \textbf{Abstention Rate (\%)} \\
\midrule

\texttt{Gemini~2.0~Flash}
 & \metric{0.577833} (\ci{0.565670}--\ci{0.590448})
 & \metric{0.739939} (\ci{0.732649}--\ci{0.749215})
 & \metric{0.721996} (\ci{0.713058}--\ci{0.731351})
 & \metric{0.679066} (\ci{0.648518}--\ci{0.707649})
 & \metric{0.719120}
 & \underline{\textbf{\metric{0.173139}}}
 & \metric{0.213237}
 & \metric{99.97}
 & \metric{0.027118644} \\

\texttt{Gemini~2.5~Flash}
 & \underline{\textbf{\metric{0.612046}}} (\ci{0.596073}--\ci{0.629174})
 & \underline{\textbf{\metric{0.769355}}} (\ci{0.760352}--\ci{0.776717})
 & \underline{\textbf{\metric{0.758083}}} (\ci{0.749387}--\ci{0.767663})
 & \metric{0.683918} (\ci{0.659943}--\ci{0.708097})
 & \metric{0.601556}
 & \metric{0.178346}
 & \underline{\textbf{\metric{0.211461}}}
 & \underline{\textbf{\metric{100.00}}}
 & \metric{0.013555646} \\

\texttt{Gemini~2.5~Pro}
 & \metric{0.594458} (\ci{0.575786}--\ci{0.610713})
 & \metric{0.753350} (\ci{0.744725}--\ci{0.763074})
 & \metric{0.747117} (\ci{0.738176}--\ci{0.757258})
 & \metric{0.698083} (\ci{0.674420}--\ci{0.722946})
 & \metric{0.654758}
 & \metric{0.212496}
 & \metric{0.228542}
 & \metric{99.97}
 & \metric{0.108474576} \\

\texttt{MedGemma~1.5~4B}
 & \metric{0.557444} (\ci{0.542215}--\ci{0.572004})
 & \metric{0.723397} (\ci{0.712873}--\ci{0.733629})
 & \metric{0.724549} (\ci{0.715453}--\ci{0.732839})
 & \metric{0.591174} (\ci{0.568623}--\ci{0.616303})
 & \metric{0.610520}
 & \metric{0.260842}
 & \metric{0.281554}
 & \underline{\textbf{\metric{100.00}}}
 & \underline{\textbf{\metric{0.0000}}} \\

\texttt{MedGemma~1~4B}
 & \metric{0.408694} (\ci{0.399107}--\ci{0.418075})
 & \metric{0.598584} (\ci{0.590701}--\ci{0.609730})
 & \metric{0.608177} (\ci{0.600034}--\ci{0.617884})
 & \metric{0.444607} (\ci{0.433912}--\ci{0.455936})
 & \metric{0.492620}
 & \metric{0.357827}
 & \metric{0.365184}
 & \underline{\textbf{\metric{100.00}}}
 & \metric{0.067778230} \\

\hline

\texttt{LLaMA~4~Maverick}
 & \metric{0.520425} (\ci{0.507162}--\ci{0.535534})
 & \metric{0.708752} (\ci{0.699067}--\ci{0.720006})
 & \metric{0.700229} (\ci{0.689034}--\ci{0.710503})
 & \metric{0.573159} (\ci{0.541754}--\ci{0.592668})
 & \metric{0.716861}
 & \metric{0.228507}
 & \metric{0.248076}
 & \metric{97.76}
 & \metric{0.041597338} \\

\hline

\texttt{GPT-4o}
 & \metric{0.594577} (\ci{0.573820}--\ci{0.608566})
 & \metric{0.728369} (\ci{0.717283}--\ci{0.737043})
 & \metric{0.728901} (\ci{0.718033}--\ci{0.737263})
 & \metric{0.701092} (\ci{0.680282}--\ci{0.722861})
 & \metric{0.645300}
 & \metric{0.204722}
 & \metric{0.233606}
 & \metric{99.91}
 & \metric{0.0000} \\

\texttt{GPT-5~Chat}
 & \metric{0.580309} (\ci{0.564484}--\ci{0.595444})
 & \metric{0.718468} (\ci{0.710532}--\ci{0.727210})
 & \metric{0.711646} (\ci{0.701199}--\ci{0.721593})
 & \underline{\textbf{\metric{0.725009}}} (\ci{0.698322}--\ci{0.752150})
 & \underline{\textbf{\metric{0.733728}}}
 & \metric{0.217035}
 & \metric{0.244619}
 & \metric{99.76}
 & \underline{\textbf{\metric{0.0000}}} \\

\hline

\texttt{Qwen~2.5-VL~32B~Instruct}
 & \metric{0.204519} (\ci{0.199283}--\ci{0.208978})
 & \metric{0.379587} (\ci{0.368018}--\ci{0.389861})
 & \metric{0.481052} (\ci{0.468106}--\ci{0.494117})
 & \metric{0.249569} (\ci{0.245127}--\ci{0.253524})
 & \metric{0.605410}
 & \metric{0.393394}
 & \metric{0.398675}
 & \underline{\textbf{\metric{100.00}}}
 & \metric{0.121951220} \\

\bottomrule
\end{tabular}
\end{threeparttable}
}
\end{table*}

\begin{table*}[ht!]
\centering
\caption{
Computational efficiency and economic cost per inference across evaluated models in Phase 3 (few-shot prompting).
Metrics include average input tokens, output tokens, total tokens, estimated input cost, output cost, and total cost.
All values reflect per-request averages under the standardized evaluation protocol.}
\label{tab:efficiency_cost_phase3_few_shot}

\resizebox{\textwidth}{!}{%
\begin{threeparttable}
\begin{tabular}{lrrrrrrr}
\toprule

\makecell[c]{Model}
& \makecell[c]{Avg In. \\ Tokens}
& \makecell[c]{Avg Out. \\ Tokens}
& \makecell[c]{Total Tokens}
& \makecell[c]{Avg Latency \\ (ms)}
& \makecell[c]{Avg In. \tnote{*} \\ Cost (est.)}
& \makecell[c]{Avg Out. \tnote{*} \\ Cost (est.)}
& \makecell[c]{Avg Cost} \\
\midrule

\texttt{\iffalse google/ \fi Gemini~2.0~Flash}
& 6869.00 & 104.95 & 6973.95 & 1752.21
& \$0.000687 & \$0.000042 & \$0.000729 \\

\texttt{\iffalse google/ \fi Gemini~2.5~Flash}
& 6868.57 & 103.68 & 6972.25 & 2091.15
& \$0.002061 & \$0.000259 & \$0.002320 \\

\texttt{\iffalse google/ \fi Gemini~2.5~Pro}
& 6867.03 & 1018.23 & 7885.26 & 11689.04
& \$0.008584 & \$0.010182 & \$0.018766 \\

\texttt{\iffalse google/ \fi MedGemma~1.5~4B \tnote{*}}
& 8081.00 & 57.07 & 8138.07 & N/A
& \$0.000994 & \$0.000026 & \$0.001020 \\

\texttt{MedGemma~1~4B \tnote{*}}
& 1068.00 & 1000.00 & 1300.00 & N/A
& \$0.000037 & \$0.000456 & \$0.000493 \\

\hline

\texttt{\iffalse meta-llama/ \fi LLaMA~4~Maverick}
& 16595.01 & 92.87 & 16687.88 & 1696.96
& \$0.002489 & \$0.000056 & \$0.002545 \\

\hline

\texttt{\iffalse openai/ \fi GPT-4o}
& 11661.72 & 92.92 & 11754.64 & 2527.89
& \$0.029154 & \$0.000929 & \$0.030084 \\

\texttt{\iffalse openai/ \fi GPT-5~Chat}
& 9861.97 & 92.43 & 9954.41 & 1582.68
& \$0.012327 & \$0.000924 & \$0.013252 \\

\hline

\texttt{\iffalse alibaba/ \fi Qwen~2.5-VL~32B~Instruct}
& 9243.28 & 95.56 & 9338.85 & N/A
& \$0.000015 & \$0.000220 & \$0.000235 \\

\bottomrule
\end{tabular}
 \begin{tablenotes}
  \item[*]MedGemma was deployed within a Google Colab Pro environment. Due to the dynamic nature of resource allocation, latency and computational costs may exhibit variance between execution runs.
  \item[**] Average input and output costs are computed from the per-request input and output costs under the pricing schedule active at the time of benchmarking.
  \end{tablenotes}
  \end{threeparttable}
}
\end{table*}

\begin{figure}[!htbp]
\centering
\includegraphics[width=0.9\textwidth]{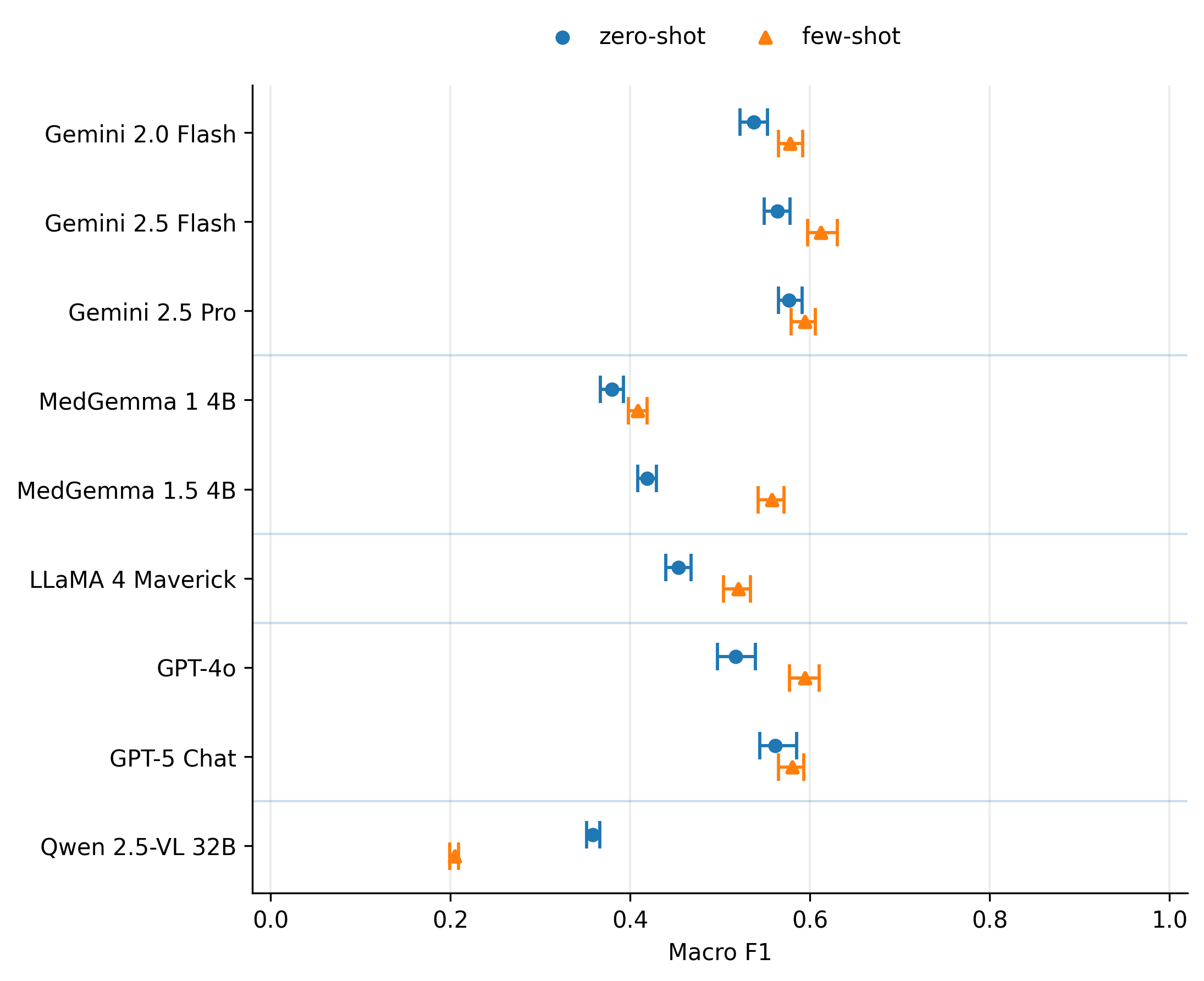}
\caption{Comparison between zero-shot and few-shot Macro-F1 performance with confidence intervals for the primary diagnostic task. Points indicate the model’s Macro-F1 score under zero-shot and few-shot prompting, and horizontal whiskers represent 95\% confidence intervals. Models are grouped by provider to facilitate comparison across model families.}
\label{fig:phase3_diagnosis_forest_zero_vs_few_shot}
\end{figure}

Few-shot prompting (Table~\ref{tab:final_few_shot}) improves performance for several models, although the magnitude of these gains varies across architectures and is accompanied by higher operational overhead (Table~\ref{tab:efficiency_cost_phase3_few_shot}). This is additionally demonstrated with Fig.~\ref{fig:phase3_diagnosis_forest_zero_vs_few_shot}, on which zero-shot and few-shot Macro-F1 performance with 95\% confidence intervals are depicted. While most models benefit (in terms of Macro F1) from exemplar-based prompting, particularly Gemini~2.5~Flash and MedGemma~1.5~4B—the effect is architecture-dependent, with Qwen~2.5-VL~32B showing a substantial decline under few-shot conditions. Confidence intervals remain relatively narrow, indicating stable performance differences rather than variability driven by sampling noise.

Among proprietary frontier models, Gemini~2.5~Flash achieves the strongest balanced diagnostic performance, obtaining the highest Macro-F1 (0.612), weighted Macro-F1 (0.769), and Micro-F1 (0.758). At the same time, it maintains excellent structured-output reliability and competitive calibration. These results indicate that Gemini~2.5~Flash benefits significantly from few-shot prompting while preserving stable output behavior. Gemini~2.5~Pro also performs strongly under few-shot prompting. While its diagnostic metrics remain competitive, this improvement comes at a higher operational cost. The model is characterized by over 1000 output tokens on average, significantly higher latency and the highest inference cost among all evaluated models. 

GPT-family models exhibit a different pattern. GPT-5~Chat achieves the highest overall accuracy and the strongest AUC, indicating strong class separability and effective ranking of diagnostic probabilities. However, its Macro-F1 remains slightly below the leading Gemini models, yet still comparable. GPT-4o performs comparably in balanced metrics (Macro-F1 = 0.595) and maintains excellent structured-output reliability with zero abstention, although its operational cost is significantly higher than that of the Gemini models due to much larger prompt sizes in the few-shot setting. Moreover, under few-shot prompting, GPT-4o slightly surpasses GPT-5~Chat in discriminative classification metrics, reversing their relative ordering observed in the zero-shot setting. However, this improvement is accompanied by higher input tokens inference cost, resulting in the highest average total cost among the evaluated proprietary models.

Open-weight models demonstrate improvements but remain below the proprietary frontier systems. MedGemma~1.5~4B benefits noticeably from few-shot prompting, achieving the strongest performance among open-weight architectures while maintaining perfect JSON validity and zero abstention. LLaMA~4~Maverick comes next from the open-weight models while maintaining relatively low latency. However, its structured-output reliability is the lowest among the evaluated models. On the other side, MedGemma~1~4B remains weaker, indicating that few-shot prompting cannot fully overcome limitations in model capacity. Finally, the general-purpose open-weight model Qwen~2.5-VL~32B shows a significant decline in discriminative performance under few-shot prompting, despite maintaining perfect structured-output validity and low computational cost. This result suggests that additional context examples may not consistently improve reasoning for all multimodal architectures and may sometimes destabilize classification behavior.

It is very important to emphasize that few-shot prompting partially narrows the performance gap between open-weight and proprietary models. In particular, the medically pretrained MedGemma 1.5 4B achieves a Macro-F1 of approximately 0.557, which approaches the best zero-shot performance of the proprietary frontier models, such as Gemini 2.5 Flash (0.612) and Gemini 2.5 Pro (0.577) and outperforms Gemini 2.0 Flash (0.537) and GPT-4o (0.517). These findings suggest that domain-specialized multimodal models, in particular MedGemma 1.5 4B, represent a promising direction for future research, although large-scale multimodal training currently provides a clear advantage for complex diagnostic reasoning tasks.

Few-shot prompting increases the number of tokens processed per request, leading to higher latency and inference cost, particularly for proprietary API-based models where pricing scales directly with token usage. In contrast, open-weight models can be locally hosted without per-request API charges, although they still require computational resources for inference. The results highlight an important practical trade-off: while few-shot prompting can improve diagnostic performance for most of the models, this overhead may limit scalability in real-world clinical deployments.

\begin{table*}[ht!]
\centering
\caption{Phase 3 — Detailed abstention-aware Macro-F1 performance by model across output fields under zero-shot prompting. 
Performance is reported for the primary diagnosis, detailed diagnosis, imaging modality, specialized imaging sequence, and anatomical plane. 
Values in parentheses denote 95\% confidence intervals. 
Bold underlined values indicate the best observed point estimate per column.}
\label{tab:f1_zero_shot_phase3}
\resizebox{\textwidth}{!}{%
\begin{tabular}{lccccc}
\toprule
\makecell[c]{Model} &
\makecell[c]{Diagnosis} &
\makecell[c]{Detailed\\diagnosis} &
\makecell[c]{Modality} &
\makecell[c]{Specialized\\sequence} &
\makecell[c]{Plane} \\
\midrule

\texttt{Gemini~2.0~Flash}  
& \metric{0.537340} (\ci{0.522433}--\ci{0.553087})
& \metric{0.223363} (\ci{0.213574}--\ci{0.234630})
& \metric{0.999368} (\ci{0.998793}--\ci{0.999802})
& \metric{0.851496} (\ci{0.841198}--\ci{0.862019})
& \metric{0.986122} (\ci{0.979958}--\ci{0.990671}) \\

\texttt{Gemini~2.5~Flash}  
& \metric{0.563861} (\ci{0.549318}--\ci{0.577596})
& \metric{0.258661} (\ci{0.242944}--\ci{0.270849})
& \metric{0.999278} (\ci{0.998623}--\ci{0.999856})
& \metric{0.783816} (\ci{0.775571}--\ci{0.794867})
& \metric{0.976080} (\ci{0.969695}--\ci{0.982821}) \\

\texttt{Gemini~2.5~Pro}  
& \underline{\textbf{\metric{0.576743}}} (\ci{0.564799}--\ci{0.591176})
& \underline{\textbf{\metric{0.322712}}} (\ci{0.305275}--\ci{0.337935})
& \metric{0.999422} (\ci{0.998914}--\ci{0.999931})
& \metric{0.792360} (\ci{0.780181}--\ci{0.802454})
& \metric{0.970052} (\ci{0.963191}--\ci{0.977253}) \\

\texttt{MedGemma~1~4B} 
& \metric{0.379716} (\ci{0.366975}--\ci{0.392677})
& \metric{0.101338} (\ci{0.091438}--\ci{0.111571})
& \underline{\textbf{\metric{1.000000}}} (\ci{1.000000}--\ci{1.000000})
& \metric{0.169878} (\ci{0.159937}--\ci{0.183787})
& \underline{\textbf{\metric{1.000000}}} (\ci{1.000000}--\ci{1.000000}) \\

\texttt{MedGemma~1.5~4B} 
& \metric{0.418824} (\ci{0.408301}--\ci{0.429015})
& \metric{0.071423} (\ci{0.063003}--\ci{0.079391})
& \metric{0.947443} (\ci{0.942685}--\ci{0.951259})
& \metric{0.251111} (\ci{0.238627}--\ci{0.264367})
& \metric{0.502978} (\ci{0.488302}--\ci{0.518245}) \\

\midrule

\texttt{LLaMA~4~Maverick}  
& \metric{0.453718} (\ci{0.439437}--\ci{0.467487})
& \metric{0.225677} (\ci{0.210270}--\ci{0.240989})
& \metric{0.996611} (\ci{0.995264}--\ci{0.997969})
& \metric{0.641746} (\ci{0.629325}--\ci{0.653629})
& \metric{0.984528} (\ci{0.978004}--\ci{0.990612}) \\

\midrule

\texttt{GPT-4o}  
& \metric{0.517257} (\ci{0.497038}--\ci{0.539441})
& \metric{0.197652} (\ci{0.188236}--\ci{0.205686})
& \metric{0.997653} (\ci{0.996634}--\ci{0.998416})
& \metric{0.753107} (\ci{0.742713}--\ci{0.762626})
& \metric{0.994830} (\ci{0.992131}--\ci{0.997612}) \\

\texttt{GPT-5~Chat}  
& \metric{0.561478} (\ci{0.544015}--\ci{0.585442})
& \metric{0.234484} (\ci{0.223872}--\ci{0.247264})
& \metric{0.998180} (\ci{0.997122}--\ci{0.998988})
& \underline{\textbf{\metric{0.816257}}} (\ci{0.807140}--\ci{0.825738})
& \metric{0.994804} (\ci{0.991342}--\ci{0.997149}) \\

\midrule

\texttt{Qwen~2.5-VL~32B~Instruct}  
& \metric{0.358283} (\ci{0.351528}--\ci{0.366222})
& \metric{0.109475} (\ci{0.102306}--\ci{0.116286})
& \underline{\textbf{\metric{1.000000}}} (\ci{1.000000}--\ci{1.000000})
& \metric{0.243461} (\ci{0.233549}--\ci{0.253696})
& \underline{\textbf{\metric{1.000000}}} (\ci{1.000000}--\ci{1.000000}) \\

\bottomrule
\end{tabular}
}
\end{table*}

\begin{table*}[ht!]
\centering
\caption{Phase 3 — Detailed abstention-aware Macro-F1 performance by model across structured output fields under \textbf{few-shot} prompting (4 examples per class). 
Reported are Macro-F1 scores with abstention for the primary diagnosis, detailed diagnosis, imaging modality, specialized imaging sequence, and anatomical plane. 
Values in parentheses denote 95\% confidence intervals. Bold underlined values indicate the best observed point estimate per column (ties highlighted equally).}
\label{tab:f1_few_shot_phase3}

\resizebox{\textwidth}{!}{%
\begin{tabular}{lccccc}
\toprule
\makecell[c]{Model} &
\makecell[c]{Diagnosis} &
\makecell[c]{Detailed\\diagnosis} &
\makecell[c]{Modality} &
\makecell[c]{Specialized\\sequence} &
\makecell[c]{Plane} \\
\midrule

\texttt{Gemini~2.0~Flash}  
& \metric{0.577833} (\ci{0.564862} - \ci{0.591770})
& \metric{0.241156} (\ci{0.228556} - \ci{0.253517})
& \metric{0.999279} (\ci{0.998627} - \ci{0.999730})
& \metric{0.867514} (\ci{0.853118} - \ci{0.878966})
& \metric{0.990003} (\ci{0.984686} - \ci{0.995080}) \\

\texttt{Gemini~2.5~Flash}  
& \textbf{\underline{\metric{0.612046}}} (\ci{0.597689} - \ci{0.630406})
& \metric{0.284182} (\ci{0.266233} - \ci{0.307984})
& \metric{0.999567} (\ci{0.999133} - \ci{1.000000})
& \metric{0.786989} (\ci{0.776063} - \ci{0.797023})
& \metric{0.977916} (\ci{0.971668} - \ci{0.984430}) \\

\texttt{Gemini~2.5~Pro}  
& \metric{0.594458} (\ci{0.578917} - \ci{0.606029})
& \textbf{\underline{\metric{0.329969}}} (\ci{0.309375} - \ci{0.346206})
& \metric{0.999513} (\ci{0.998981} - \ci{1.000000})
& \metric{0.806477} (\ci{0.796803} - \ci{0.817593})
& \metric{0.982505} (\ci{0.977767} - \ci{0.987512}) \\

\texttt{MedGemma~1~4B} 
& \metric{0.408694} (\ci{0.398150} - \ci{0.418592})
& \metric{0.093754} (\ci{0.084582} - \ci{0.101230})
& \textbf{\underline{\metric{1.000000}}} (\ci{1.000000} - \ci{1.000000})
& \metric{0.273237} (\ci{0.261182} - \ci{0.289666})
& \textbf{\underline{\metric{1.000000}}} (\ci{1.000000} - \ci{1.000000}) \\

\texttt{MedGemma~1.5~4B} 
& \metric{0.557444} (\ci{0.542496} - \ci{0.570977})
& \metric{0.137000} (\ci{0.126743} - \ci{0.149338})
& \metric{0.987522} (\ci{0.984421} - \ci{0.990213})
& \metric{0.405988} (\ci{0.387703} - \ci{0.419932})
& \metric{0.643121} (\ci{0.617029} - \ci{0.669541}) \\

\hline

\texttt{LLaMA~4~Maverick}  
& \metric{0.520425} (\ci{0.503816} - \ci{0.533775})
& \metric{0.251110} (\ci{0.235960} - \ci{0.262830})
& \metric{0.995727} (\ci{0.993954} - \ci{0.997099})
& \metric{0.692319} (\ci{0.679338} - \ci{0.705375})
& \metric{0.994857} (\ci{0.990883} - \ci{0.997956}) \\

\hline

\texttt{GPT-4o}  
& \metric{0.594577} (\ci{0.576952} - \ci{0.610086})
& \metric{0.169312} (\ci{0.157918} - \ci{0.176765})
& \metric{0.996827} (\ci{0.995536} - \ci{0.997906})
& \metric{0.783862} (\ci{0.771547} - \ci{0.793697})
& \metric{0.988745} (\ci{0.983728} - \ci{0.993575}) \\

\texttt{GPT-5~Chat}  
& \metric{0.580309} (\ci{0.564867} - \ci{0.592970})
& \metric{0.212017} (\ci{0.201731} - \ci{0.220542})
& \metric{0.996678} (\ci{0.995438} - \ci{0.997907})
& \textbf{\underline{\metric{0.835036}}} (\ci{0.826869} - \ci{0.843965})
& \metric{0.996380} (\ci{0.994197} - \ci{0.997876}) \\

\hline

\texttt{Qwen~2.5-VL~32B~Instruct}  
& \metric{0.204519} (\ci{0.199489} - \ci{0.208890})
& \metric{0.038319} (\ci{0.035693} - \ci{0.040878})
& \textbf{\underline{\metric{1.000000}}} (\ci{1.000000} - \ci{1.000000})
& \metric{0.197014} (\ci{0.186842} - \ci{0.206499})
& \textbf{\underline{\metric{1.000000}}} (\ci{1.000000} - \ci{1.000000}) \\

\bottomrule
\end{tabular}
}
\end{table*}

Table~\ref{tab:f1_zero_shot_phase3} summarizes Phase~3 performance (Macro-F1) of multimodal large language models on the structured radiology prompting task, reported separately for each output field, namely modality identification, anatomical plane recognition, MRI specialized sequence classification, primary diagnosis category, and diagnosis subtype.

The results reveal a clear hierarchy of task difficulty. Low-level metadata fields such as imaging modality are predicted nearly perfectly by most models, including Gemini~2.5~Pro, GPT-5~Chat, and GPT-4o. Anatomical plane recognition is also generally very strong, with several models achieving near-perfect scores (e.g., GPT-4o and GPT-5~Chat), whereas MedGemma~1.5~4B shows significantly lower performance (Macro-F1 = 0.503), indicating instability in extracting even basic metadata.

In contrast, MRI specialized sequence classification exhibits substantial variability across models. GPT-5~Chat achieves the strongest performance on this task, while several smaller or domain-specialized models, including MedGemma~1~4B and Qwen~2.5-VL~32B, perform considerably worse. This suggests that sequence-specific imaging characteristics are not uniformly captured across architectures.

Clinically semantic tasks remain the most challenging. Primary diagnostic categorization shows moderate performance with clear inter-model differences, with Gemini~2.5~Pro and Gemini~2.5~Flash achieving the strongest results. Diagnostic subtype prediction remains the most difficult task overall, even for the best-performing models. For example, Gemini~2.5~Pro achieves the highest detailed diagnosis score, while LLaMA~4~Maverick achieve competitive performance that approaches Gemini~2.5~Flash, outperforming several other systems. In contrast, domain-specialized models such as MedGemma and general open-weight systems such as Qwen~2.5-VL~32B exhibit very low performance on this task.

Overall, larger general-purpose multimodal models such as Gemini~2.5~Pro and GPT-5~Chat demonstrate the most balanced performance across all output fields. In contrast, smaller or domain-specialized models often show strong results on specific technical attributes—for example, MedGemma~1~4B achieving perfect scores for modality and plane—but exhibit reduced robustness on clinically meaningful diagnostic reasoning tasks.



Similarly, Table~\ref{tab:f1_few_shot_phase3} reports Phase~3 performance under structured radiology prompting in a few-shot setting. Compared with zero-shot evaluation, the effect of few-shot prompting is task-specific rather than uniform. The most consistent improvement is observed for primary diagnostic categorization (Diagnosis), where almost all evaluated models demonstrate higher Macro-F1 scores.

\begin{figure}[htbp]
    \centering
    \begin{subfigure}[b]{0.48\textwidth}
        \centering
        \includegraphics[width=\textwidth]{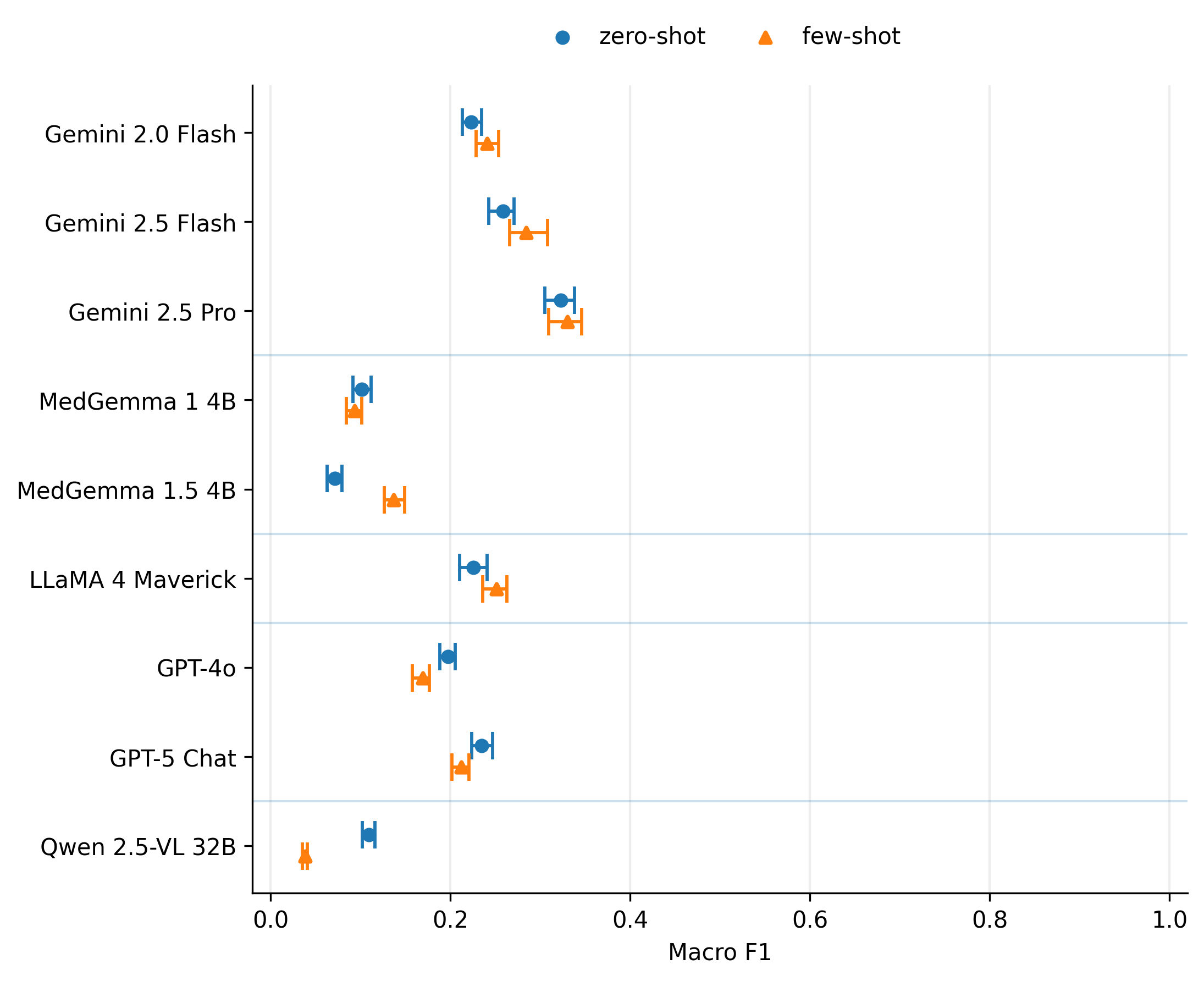}
        \caption{Detailed diagnosis}
        \label{fig:diagnosis_detailed}
    \end{subfigure}
    \hfill
    \begin{subfigure}[b]{0.48\textwidth}
        \centering
        \includegraphics[width=\textwidth]{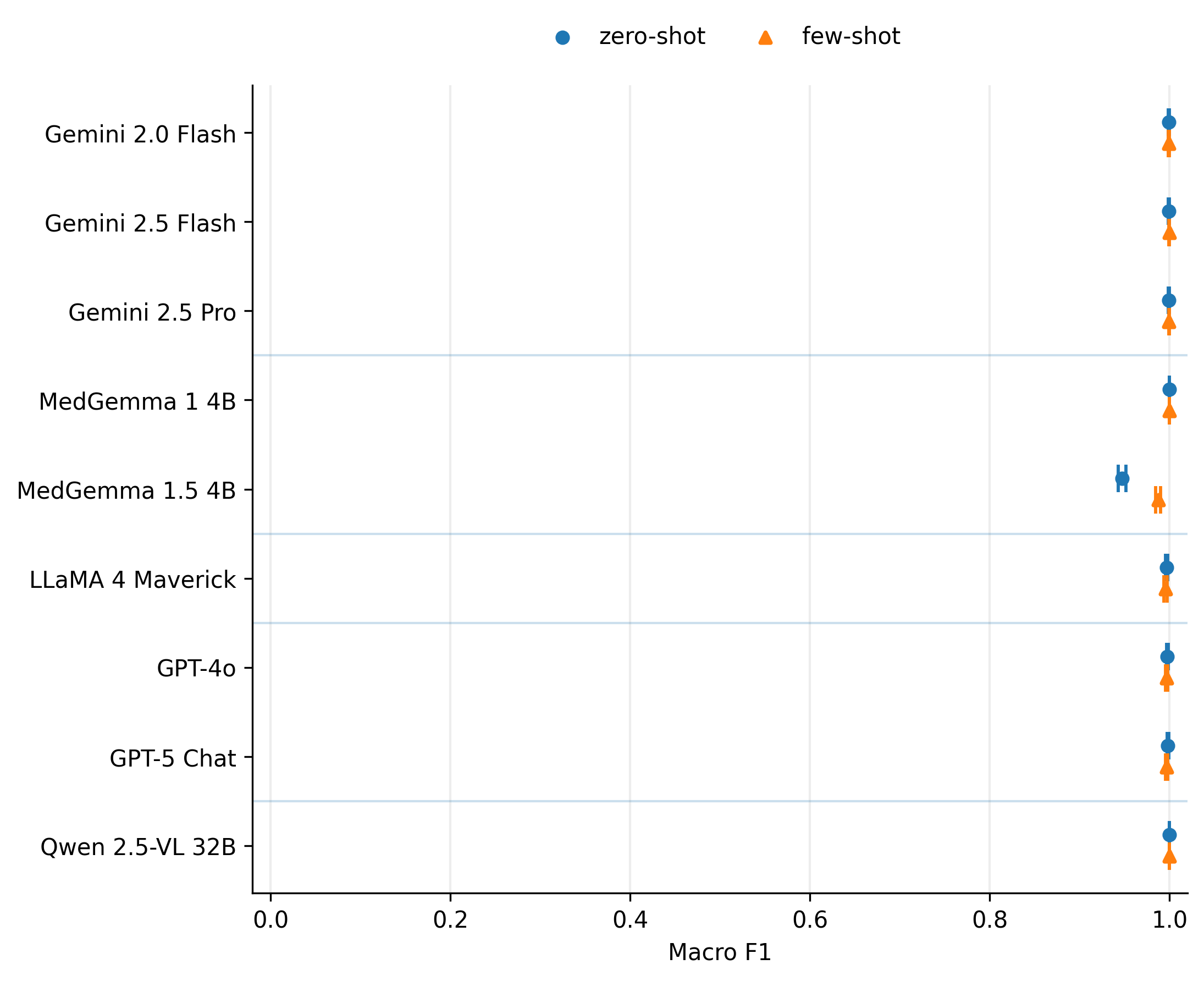}
        \caption{Imaging modality.}
        \label{fig:modality}
    \end{subfigure}
    \hfill
    \begin{subfigure}[b]{0.48\textwidth}
        \centering
        \includegraphics[width=\textwidth]{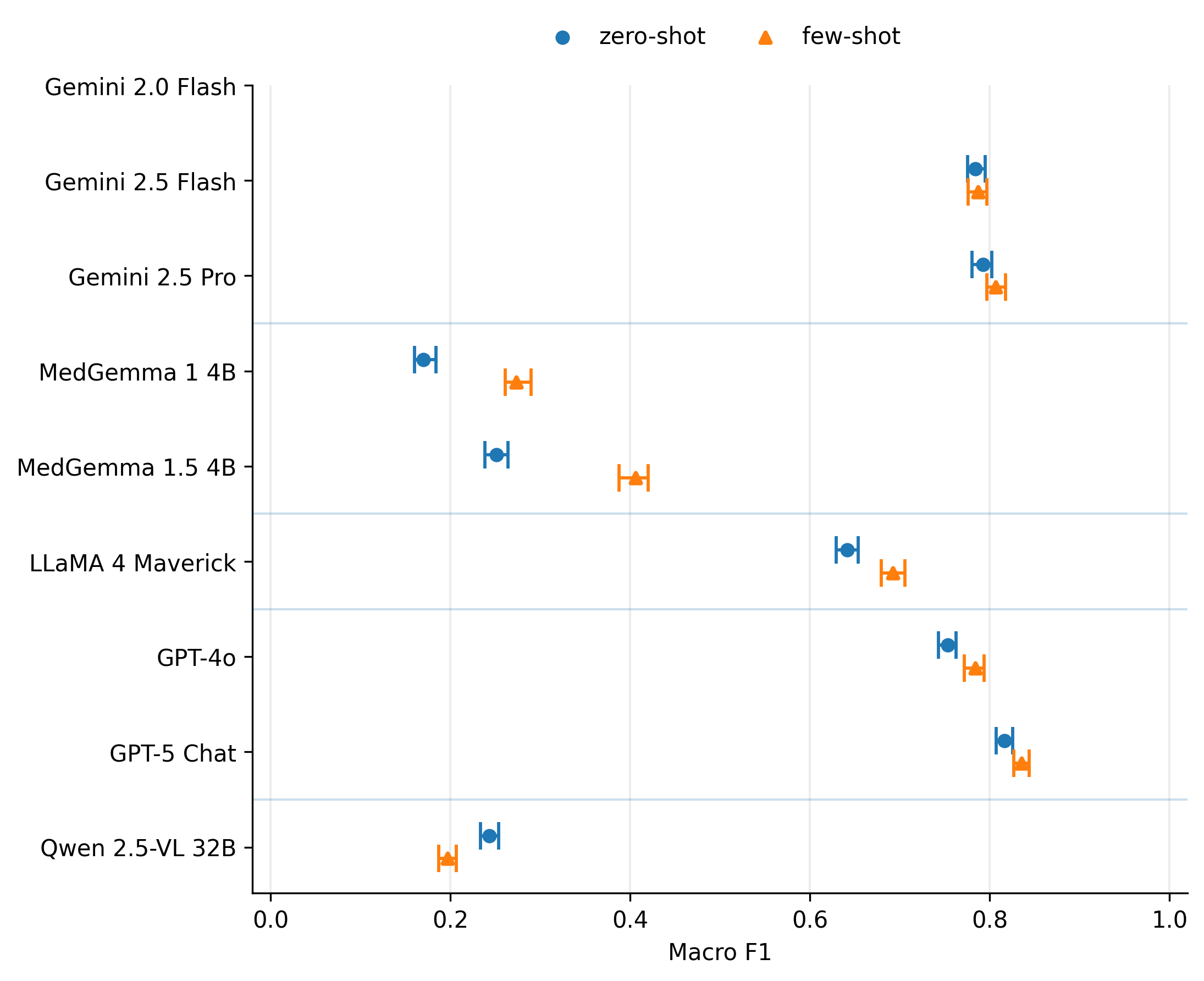}
        \caption{Specialized imaging sequence.}
        \label{fig:plane}
    \end{subfigure}
    \hfill
    \begin{subfigure}[b]{0.48\textwidth}
        \centering
        \includegraphics[width=\textwidth]{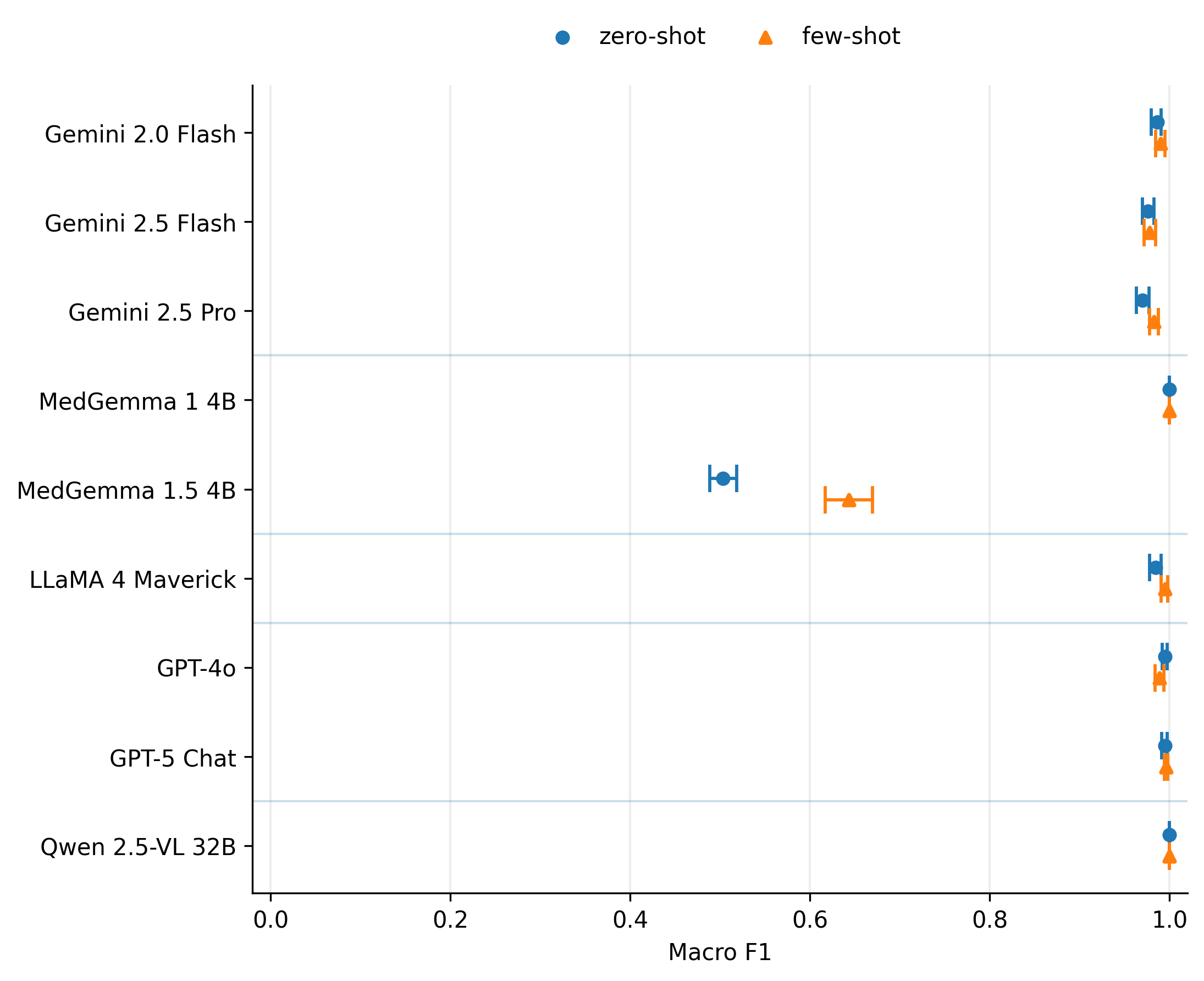}
        \caption{Anatomical plane.}
        \label{fig:specialized_sequence}
    \end{subfigure}
    \caption{Comparison between zero-shot and few-shot Macro-F1 performance with confidence intervals across additional structured output fields: (a) detailed diagnosis, (b) imaging modality, (c) specialized imaging sequence, and (d) anatomical plane. Points represent the model Macro-F1 scores under zero-shot and few-shot prompting, while horizontal whiskers denote 95\% confidence intervals. Models are grouped by provider to facilitate comparison across model families.}
    \label{fig:output_fields_zeroS_fewS}
\end{figure}

On the other side, other structured output fields exhibit more heterogeneous responses. Imaging modality identification remains near ceiling across models, with several systems achieving perfect or near-perfect scores, indicating saturation effects. Anatomical plane recognition also remains highly stable across prompting settings, although some domain-specialized models, such as MedGemma~1.5~4B, continue to show lower performance.

A detailed comparison between zero-shot and few-shot prompting across the remaining structured output fields is illustrated in Fig. \ref{fig:output_fields_zeroS_fewS}, which reports Macro-F1 scores with confidence intervals for detailed diagnosis, modality, specialized sequence, and anatomical plane recognition.

MRI specialized sequence classification displays architecture-dependent behavior. GPT-5~Chat achieves the strongest performance on this task, suggesting strong sensitivity to sequence-level imaging features, while other models show more moderate improvements. These results indicate that sequence recognition benefits non uniformly from in-context examples.

Diagnostic subtype prediction (Detailed Diagnosis) remains the most difficult task. Even with few-shot prompting, the highest score is achieved by Gemini~2.5~Pro (Macro-F1 = 0.33), while several models remain far below this level. Although some improvement in comparison to zero-shot is visible for the leading models, more detailed and subtle clinical reasoning remains a major challenge.

The response of open-weight and domain-specialized models further illustrates this variability. MedGemma~1.5~4B benefits significantly from few-shot prompting in primary diagnosis prediction, approaching the performance of larger proprietary models. However, its performance on sequence classification and plane recognition remains non uniform. LLaMA~4~Maverick achieves moderate and balanced performance across most tasks, while Qwen~2.5-VL~32B shows a clear performance decline under few-shot prompting despite perfect metadata extraction. This highlights that additional in-context examples may destabilize reasoning in some architectures.

Overall, the results indicate that few-shot prompting reliably improves coarse diagnostic categorization but does not consistently enhance other clinically relevant outputs. While larger general-purpose multimodal models tend to utilize in-context examples more effectively, the benefits remain task-dependent and do not eliminate the challenges associated with more specific clinical reasoning.

\begin{figure}[!htbp]
\centering
\includegraphics[width=0.9\textwidth]{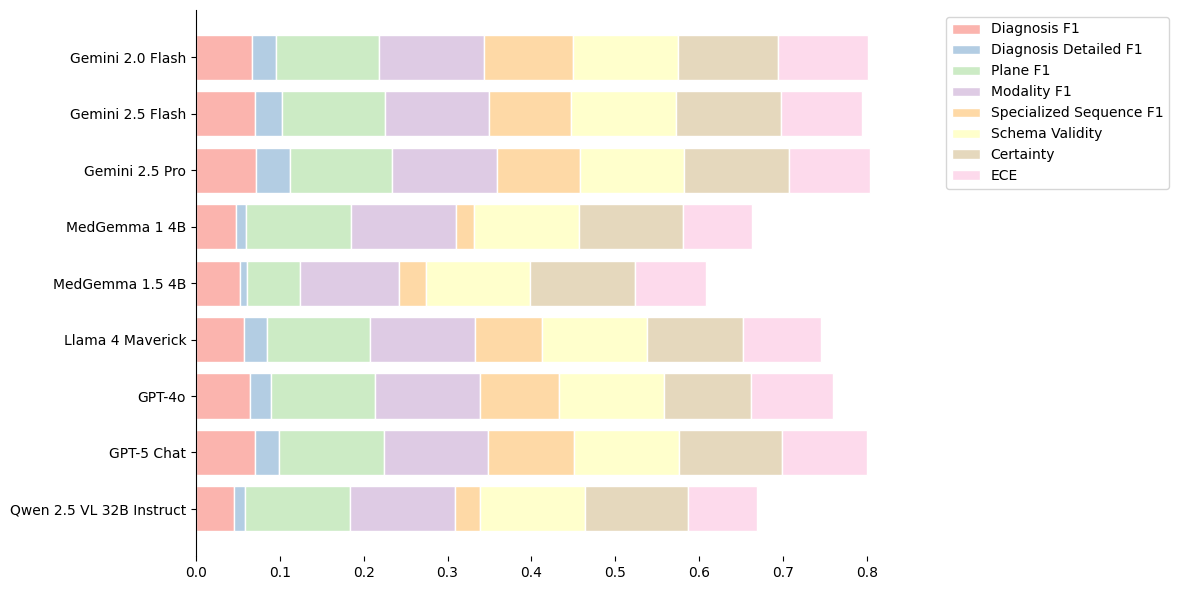}
\caption{Illustrative multidimensional overview of model performance across selected output fields and evaluation dimensions in Phase~3 of the benchmark with zero-shot prompting. Each horizontal bar summarizes diagnostic performance, detailed diagnosis, modality prediction, specialized sequence recognition, anatomical plane detection, structured schema validity, model certainty, and calibration (ECE), enabling qualitative comparison of model behaviour across tasks.}
\label{fig:general_stacked_phase3_zero_shot}
\end{figure}

\begin{figure}[!htbp]
\centering
\includegraphics[width=0.9\textwidth]{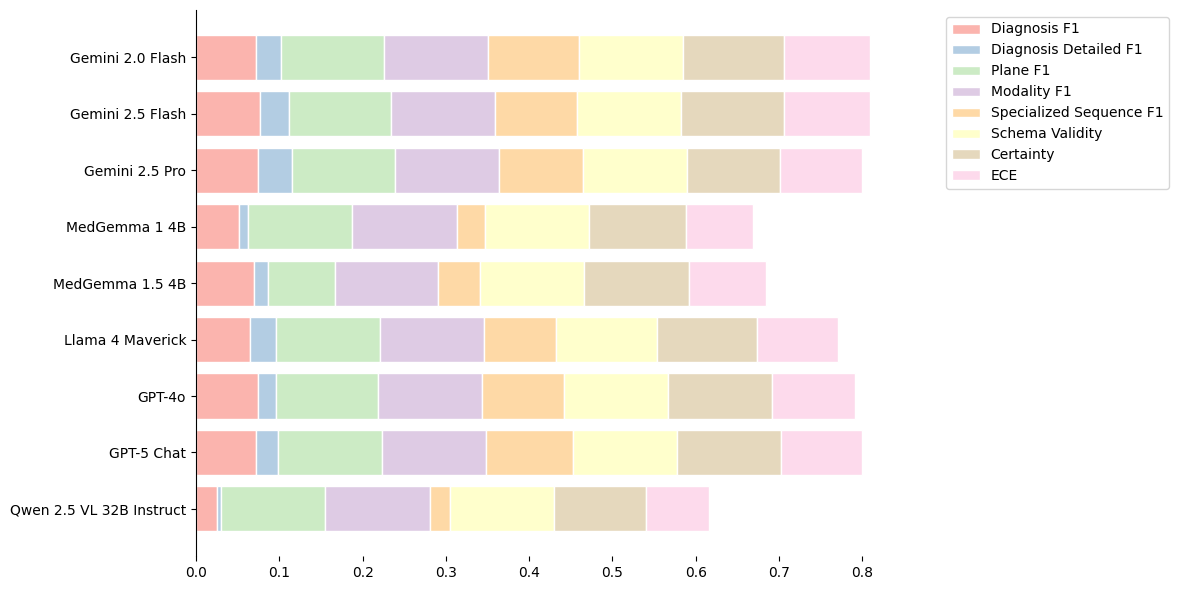}
\caption{Illustrative multidimensional overview of model performance across selected output fields and evaluation dimensions in Phase~3 of the benchmark with few-shot prompting. Each horizontal bar summarizes diagnostic performance, detailed diagnosis, modality prediction, specialized sequence recognition, anatomical plane detection, structured schema validity, model certainty, and calibration (ECE), enabling qualitative comparison of model behaviour across tasks.}
\label{fig:general_stacked_phase3_few_shot}
\end{figure}

Fig.~\ref{fig:general_stacked_phase3_zero_shot} and Fig.~\ref{fig:general_stacked_phase3_few_shot} presents an illustrative multidimensional overview of model performance across several output fields and evaluation dimensions assessed in Phase~3 of the benchmark, including discriminative diagnostic classification performance, imaging attribute prediction, structured output validity, and calibration under both zero-shot and few-shot prompting. Each horizontal bar aggregates multiple evaluation metrics for a given model, enabling a compact visual comparison of overall capability across tasks.

Tables~\ref{tab:f1_per_class_zero_shot} and \ref{tab:f1_per_class_few_shot} report per-class F1-scores (with abstention) for the final evaluation, highlighting substantial heterogeneity in model behavior across diagnostic categories.

\begin{table*}[ht!]
\centering
\caption{Per-class F1-scores with abstention per class (Diagnosis name) in the case of zero-shot prompting.}
\label{tab:f1_per_class_zero_shot}
\resizebox{\textwidth}{!}{%
\begin{tabular}{lccccc}
\toprule
\makecell[l]{Model} &
\makecell[c]{Tumor} &
\makecell[c]{Stroke} &
\makecell[c]{Multiple sclerosis} &
\makecell[c]{Normal} &
\makecell[c]{Other abnormalities} \\
\midrule
\texttt{Gemini~2.0~Flash}  & \metric{0.908219} (\ci{0.900553} - \ci{0.914694})  & \metric{0.471398} (\ci{0.442212} - \ci{0.495413})  & \metric{0.536941} (\ci{0.501028} - \ci{0.568862})  & \metric{0.654589} (\ci{0.638119} - \ci{0.666007})  & \metric{0.115556} (\ci{0.074838} - \ci{0.153718}) \\
\texttt{Gemini~2.5~Flash}  & \metric{0.897496} (\ci{0.887594} - \ci{0.905480})  & \metric{0.525682} (\ci{0.498729} - \ci{0.547017})  & \metric{0.601527} (\ci{0.563463} - \ci{0.638285})  & \textbf{\underline{\metric{0.696400}}} (\ci{0.680566} - \ci{0.709531}) & \metric{0.098200} (\ci{0.072488} - \ci{0.125956}) \\
\texttt{Gemini~2.5~Pro}  & \textbf{\underline{\metric{0.912138}}} (\ci{0.904834} - \ci{0.917815}) & \metric{0.642009} (\ci{0.622978} - \ci{0.657167})  & \textbf{\underline{\metric{0.623244}}} (\ci{0.588971} - \ci{0.658554}) &  \metric{0.622989} (\ci{0.609343} - \ci{0.637637}) & \metric{0.083333} (\ci{0.040738} - \ci{0.130400}) \\
\texttt{MedGemma~1~4B} & \metric{0.771649} (\ci{0.761765} - \ci{0.780759})  & \metric{0.348624} (\ci{0.321967} - \ci{0.370727})  & \metric{0.305913} (\ci{0.273853} - \ci{0.337421})  & \metric{0.472393} (\ci{0.452272} - \ci{0.493033})  & \metric{0.000000} (\ci{0.000000} - \ci{0.000000}) \\
\texttt{MedGemma~1.5~4B} & \metric{0.822663} (\ci{0.813278} - \ci{0.835302})  & \metric{0.152486} (\ci{0.130070} - \ci{0.176538})  & \metric{0.415385} (\ci{0.374093} - \ci{0.454666})  & \metric{0.679195} (\ci{0.665810} - \ci{0.695915})  & \metric{0.024390} (\ci{0.007622} - \ci{0.045744}) \\
\hline
\texttt{LLaMA~4~Maverick}  & \metric{0.877331} (\ci{0.868646} - \ci{0.886318})  & \metric{0.347092} (\ci{0.324999} - \ci{0.371123})  & \metric{0.327986} (\ci{0.267886} - \ci{0.367615})  & \metric{0.660917} (\ci{0.650511} - \ci{0.676694})  & \metric{0.055263} (\ci{0.033897} - \ci{0.077939}) \\
\hline
\texttt{GPT-4o}  & \metric{0.875595} (\ci{0.865584} - \ci{0.884394})  & \metric{0.672936} (\ci{0.656212} - \ci{0.692704})  & \metric{0.366917} (\ci{0.326556} - \ci{0.413219})  & \metric{0.588361} (\ci{0.575099} - \ci{0.602206})  & \metric{0.082474} (\ci{0.020096} - \ci{0.150538}) \\
\texttt{GPT-5~Chat}  & \metric{0.900299} (\ci{0.893288} - \ci{0.908439})  & \textbf{\underline{\metric{0.736842}}} (\ci{0.723786} - \ci{0.753107}) & \metric{0.461343} (\ci{0.411937} - \ci{0.499836})  & \metric{0.566048} (\ci{0.548476} - \ci{0.583730})  & \textbf{\underline{\metric{0.142857}}} (\ci{0.071083} - \ci{0.231638}) \\
\hline
\texttt{Qwen~2.5-VL~32B~Instruct}  & \metric{0.710042} (\ci{0.697070} - \ci{0.722077})  & \metric{0.483652} (\ci{0.462529} - \ci{0.502064})  & \metric{0.031169} (\ci{0.005545} - \ci{0.055665})  & \metric{0.566553} (\ci{0.548021} - \ci{0.578955})  & \metric{0.000000} (\ci{0.000000} - \ci{0.000000}) \\

\bottomrule
\end{tabular}%
}
\end{table*}

\begin{table*}[htbp]
\centering
\caption{Per-class F1-scores with abstention per class (Diagnosis name) in the case of few-shot prompting (4 examples per class).}
\resizebox{\textwidth}{!}{%
\begin{tabular}{lccccc}
\toprule
Model & Tumor & Stroke & Multiple Sclerosis & Normal & Other Abnormalities \\
\midrule
\texttt{Gemini~2.0~Flash}  & \metric{0.896709} (\ci{0.888945} - \ci{0.904404})  & \metric{0.630435} (\ci{0.608170} - \ci{0.648687})  & \metric{0.550839} (\ci{0.524096} - \ci{0.589630})  & \metric{0.656636} (\ci{0.643297} - \ci{0.671855})  & \metric{0.154545} (\ci{0.114341} - \ci{0.197080}) \\
\texttt{Gemini~2.5~Flash}  & \metric{0.915019} (\ci{0.907731} - \ci{0.922671})  & \metric{0.684354} (\ci{0.662001} - \ci{0.701114})  & \textbf{\underline{\metric{0.560088}}} (\ci{0.520866} - \ci{0.588889}) & \textbf{\underline{\metric{0.685491}}} (\ci{0.667954} - \ci{0.700792}) & \metric{0.215278} (\ci{0.161149} - \ci{0.293269}) \\
\texttt{Gemini~2.5~Pro}  & \textbf{\underline{\metric{0.936805}}} (\ci{0.928676} - \ci{0.942683}) & \metric{0.704992} (\ci{0.683908} - \ci{0.721429})  & \metric{0.549738} (\ci{0.515886} - \ci{0.573715})  & \metric{0.592836} (\ci{0.577359} - \ci{0.609845})  & \metric{0.187919} (\ci{0.128329} - \ci{0.243217}) \\
\texttt{MedGemma~1~4B} & \metric{0.807424} (\ci{0.795881} - \ci{0.816121})  & \metric{0.287776} (\ci{0.259758} - \ci{0.311961})  & \metric{0.347511} (\ci{0.314240} - \ci{0.380330})  & \metric{0.600759} (\ci{0.585374} - \ci{0.615285})  & \metric{0.000000} (\ci{0.000000} - \ci{0.000000}) \\
\texttt{MedGemma~1.5~4B} & \metric{0.869508} (\ci{0.861415} - \ci{0.878810})  & \metric{0.568455} (\ci{0.542565} - \ci{0.590776})  & \metric{0.531868} (\ci{0.495637} - \ci{0.575416})  & \metric{0.687390} (\ci{0.673213} - \ci{0.698854})  & \metric{0.130000} (\ci{0.065190} - \ci{0.177645}) \\
\hline
\texttt{LLaMA~4~Maverick}  & \metric{0.904655} (\ci{0.898227} - \ci{0.911171})  & \metric{0.541299} (\ci{0.519670} - \ci{0.563907})  & \metric{0.390501} (\ci{0.348263} - \ci{0.437405})  & \metric{0.635671} (\ci{0.620125} - \ci{0.652771})  & \metric{0.130000} (\ci{0.090852} - \ci{0.176391}) \\
\hline
\texttt{GPT-4o}  & \metric{0.912383} (\ci{0.903377} - \ci{0.919597})  & \metric{0.739909} (\ci{0.727776} - \ci{0.754384})  & \metric{0.538328} (\ci{0.505266} - \ci{0.569185})  & \metric{0.519108} (\ci{0.500869} - \ci{0.536527})  & \textbf{\underline{\metric{0.263158}}} (\ci{0.193520} - \ci{0.337057}) \\
\texttt{GPT-5~Chat}  & \metric{0.875947} (\ci{0.868164} - \ci{0.886926})  & \textbf{\underline{\metric{0.757693}}} (\ci{0.741787} - \ci{0.768005}) & \metric{0.525452} (\ci{0.492503} - \ci{0.561665})  & \metric{0.526709} (\ci{0.507957} - \ci{0.545507})  & \metric{0.215743} (\ci{0.156684} - \ci{0.274348}) \\
\hline
\texttt{Qwen~2.5-VL~32B~Instruct}  & \metric{0.621144} (\ci{0.610077} - \ci{0.632709})  & \metric{0.000000} (\ci{0.000000} - \ci{0.000000})  & \metric{0.000000} (\ci{0.000000} - \ci{0.000000})  & \metric{0.401452} (\ci{0.383361} - \ci{0.424151})  & \metric{0.000000} (\ci{0.000000} - \ci{0.000000}) \\

\bottomrule
\end{tabular}
}
\label{tab:f1_per_class_few_shot}
\end{table*}

Tumor detection emerges as the most robust class. Under zero-shot prompting, proprietary models, including Gemini~2.5~Pro, Gemini~2.5~Flash, GPT-5~Chat, and GPT-4o already achieve very high tumor F1-scores (0.87–0.91), with Gemini~2.5~Pro reaching the highest value (0.912). Few-shot prompting further improves tumor recognition for several models, most notably Gemini~2.5~Pro (0.937) and Gemini~2.5~Flash (0.915). However, this improvement is not uniform: GPT-5~Chat shows a decrease from 0.900 to 0.876, and Gemini~2.0~Flash decreases slightly from 0.908 to 0.897. These mixed results indicate that few-shot examples do not consistently enhance visual recognition of tumor oriented pathologies and may sometimes alter decision boundaries in ways that slightly reduce performance.

Stroke classification shows the largest gains from few-shot prompting but also shows model-specific differences. In the zero-shot setting, GPT-5~Chat achieves the highest stroke F1 (0.737), followed by GPT-4o and Gemini~2.5~Pro. Few-shot prompting improves stroke recognition for nearly all proprietary models, with GPT-5~Chat increasing to 0.758, GPT-4o to 0.740, and Gemini~2.5~Pro to 0.705. Gemini~2.0~Flash also benefits considerably (0.47 → 0.63). In contrast, some smaller or open-weight systems show weaker responses: MedGemma~1~4B decreases from 0.349 to 0.288, indicating that few-shot conditioning may not reliably improve performance for smaller models.

Multiple sclerosis (MS) remains the most difficult major diagnostic category. In the zero-shot setting, the strongest models (Gemini~2.5~Pro and Gemini~2.5~Flash) achieve F1 values slightly above 0.60. Few-shot prompting leads to only modest improvements and in several cases slight degradation: Gemini~2.5~Flash decreases from 0.602 to 0.560, and GPT-5~Chat increases only marginally. Open-weight models remain weaker overall, and Qwen~2.5-VL collapses completely in the few-shot setting, dropping from a small but non-zero F1 (0.03) to zero. These results suggest that few-shot prompting provides limited benefit for conditions characterized by small and diffuse lesions that require fine spatial reasoning.

For normal controls, performance is moderate and again, model-dependent. In the zero-shot prompting, Gemini~2.5~Flash achieves the highest F1 (0.696), followed by MedGemma~1.5~4B and LLaMA~4~Maverick. Few-shot prompting improves performance for some models. In fact, MedGemma~1.5~4B benefits at most (0.679 → 0.687), making it a model with best Macro F1 value. MedGemma~1~4B provides improved performance with the few-shot prompting as well (0.472 → 0.601). On the other hand, several frontier models show slight declines, most notably GPT-4o (0.588 → 0.519) and Gemini~2.5~Pro (0.623 → 0.593), suggesting that additional examples may shift model predictions toward pathological classes.

The Other abnormalities class remains the most unstable category across models and prompting regimes. In the zero-shot setting, GPT-5~Chat achieves the highest F1 (0.143), although scores remain low overall due to the small and heterogeneous nature of this category. Few-shot prompting improves performance for several proprietary models, most notably GPT-4o (0.082 → 0.263) and Gemini~2.5~Flash (0.098 → 0.215). Nevertheless, many models remain unable to capture this category reliably, and some systems—including MedGemma~1~4B and Qwen~2.5-VL continue to produce near-zero scores.

The per-class analysis reveals a clear hierarchy of diagnostic difficulty. Tumor recognition is consistently strong across architectures, stroke detection shows the largest improvements from few-shot prompting, and multiple sclerosis remains challenging even for the strongest systems. It is important to emphasize that few-shot prompting does not uniformly improve performance and in several cases slightly degrades the results, indicating that example based setting can alter decision thresholds and interact differently with each model. These findings highlight both the promise and the limitations of few-shot prompting for clinical neuroimaging interpretation.

\subsection{Reporting Strategy for Rare Classes}

In addition to brain tumors, multiple sclerosis, stroke, and normal controls, the dataset includes a heterogeneous other abnormalities category (abscesses, cysts, and miscellaneous encephalopathies), comprising 257 samples. These conditions are clinically relevant and were intentionally retained to reflect the diagnostic diversity encountered in real-world neuroimaging practice. All reported evaluations include this category, without exclusion or reweighting.

Due to the limited sample size and heterogeneous composition, performance estimates for this category are inherently unstable. As expected in highly imbalanced settings, models frequently failed to produce correct predictions for these cases, resulting in very low or near-zero F1 scores. In this regime, individual prediction errors exert a disproportionate influence on per-class metrics, which can highly affect aggregate performance measures.

We explicitly retain this category in all reported results to ensure transparency and to avoid masking model failure modes on rare but clinically significant conditions. Consequently, aggregate metrics should be interpreted with the understanding that performance is influenced by both well-represented diagnostic classes and underrepresented edge cases. This reporting choice aligns with clinical AI evaluation principles emphasizing full disclosure of model limitations, risk characterization, and avoidance of selective reporting, particularly in settings where rare conditions may carry disproportionate clinical risk.

\subsection{Discussion}
Results from all phases and computed evaluation metrics are displayed on the publicly available leaderboard \footnote{https://neurovlm-bench.chatmed-project.eu}.

This study provides a systematic, multi-dimensional evaluation of frontier vision-enabled multimodal large language models in neuroimaging, revealing both encouraging capabilities and persistent limitations that are highly relevant for clinical translation. Across evaluation phases, no single model demonstrates uniformly strong performance across all diagnostic categories, imaging attributes, and operational constraints. Instead, model behavior is best understood as field- and context-dependent, with distinct strengths emerging for specific diagnostic targets.

\subsubsection{Efficiency–Performance Trade-offs and Deployment Implications}

Because the benchmark splits were constructed in a stratified manner, the held-out Phase~3 evaluation confirms the main patterns observed in the initial screening phase and the development phase. In particular, the relative strength of the leading proprietary models, the heterogeneous behavior of open-weight architectures, and the task-specific effects of prompting remain consistent across phases, indicating stable benchmark conclusions.

Beyond diagnostic accuracy, operational factors clearly differentiate models. Among proprietary systems, Gemini~2.5~Flash offers the most favorable balance between performance and efficiency, achieving the strongest balanced diagnostic performance under few-shot prompting while maintaining reliable structured outputs and competitive calibration. Gemini~2.5~Pro also performs strongly but at a considerably higher operational cost, largely due to longer generated outputs and increased latency.

The GPT-family models exhibit a different trade-off profile. under few-shot prompting GPT-4o slightly surpasses GPT-5~Chat in balanced discriminative performance, reversing their relative ordering from the zero-shot setting. This difference highlights the sensitivity of closely matched models to prompting strategy. At the same time, GPT-4o incurs the highest average total inference cost among proprietary models due to increased token usage in the few-shot setting.

Few-shot prompting improves performance for several models but introduces higher computational overhead. Because exemplar-based prompts increase the number of processed tokens, both latency and inference cost rise accordingly, particularly for proprietary API-based systems where pricing scales with token usage. Consequently, improvements in balanced diagnostic performance must be considered alongside scalability and operational constraints.

Open-weight models show more heterogeneous behavior. MedGemma~1.5~4B represents the most promising medically specialized open-weight result, with few-shot prompting significantly improving its balanced diagnostic performance while preserving perfect JSON validity and zero abstention. Additionally, LLaMA~4~Maverick demonstrates notable strengths in certain detailed diagnostic tasks. In contrast, MedGemma~1~4B remains weaker, and the general-purpose Qwen~2.5-VL~32B shows a decline under few-shot prompting despite perfect structured-output validity and low computational cost.

The results indicate that diagnostic performance, calibration, reliability, and computational efficiency do not converge within a single model. These trade-offs suggest that future clinical decision-support systems may benefit from specialist-aware routing strategies, directing cases to models according to task requirements, uncertainty, or operational constraints rather than relying on a single general-purpose multimodal model.

\subsubsection{Class-Selective Strengths and Emerging “Specialist” Behavior}

Per-class analysis highlights that current multimodal large language models do not fail or succeed uniformly across diagnostic categories. Instead, several models present \emph{class-selective strengths}, suggesting an emerging form of specialist behavior. This is particularly evident when comparing tumors, stroke, multiple sclerosis, and other abnormalities across zero-shot and few-shot prompting settings.

Tumor detection is the most consistently strong class across nearly all competitive models. Gemini~2.5~Pro, Gemini~2.5~Flash, GPT-5~Chat, and GPT-4o all achieve high tumor F1-scores, with Gemini~2.5~Pro reaching the strongest performance under few-shot prompting. This likely reflects the fact that tumors often present as visually specific, mass-like abnormalities with relatively strong structural contrast. In practical terms, this suggests that tumor-oriented assistive triage may be one of the most realistic near-future applications of multimodal LLMs, especially when the task is limited to coarse detection rather than detailed subtype assignment.

Stroke exhibits a different and clinically important pattern. Compared with tumors, stroke performance is more variable across models. However, some models, especially GPT-5~Chat and GPT-4o, show strength to some extent for this class, with GPT-5~Chat achieving the strongest stroke Macro F1 under both prompting settings. This indicates that some models may be better suited to vascular pathology than to other neurological categories. This emphasizes that rather than expecting one general model to perform equally well across all classes, it may be more realistic to consider \emph{routing strategies}, in which cases suspected of acute vascular pathology are directed to models that empirically demonstrate stronger stroke sensitivity.

Multiple sclerosis remains the most difficult major diagnostic category. Even the strongest models achieve only moderate MS performance, and improvements under few-shot prompting are limited. MS recognition often depends on subtle lesion morphology, small lesion burden, anatomical distribution, and broader contextual interpretation, none of which are fully represented in a single 2D slice. Therefore, although some models show partial capability, no evaluated system can currently be considered reliable for MS diagnosis in isolation. This class highlights a key limitation of current slice-based multimodal benchmarking and points to the need for volumetric reasoning, richer clinical context, or disease-specific adaptation.

The Other abnormalities category, representing mimicking cases, further emphasizes the limits of current models. Performance remains low across all systems in both prompting settings, although GPT-4o and GPT-5~Chat show relatively better results than the remaining models. Because this class is both rare and heterogeneous, it represents exactly the type of category where errors are likely to be clinically costly. This finding has direct safety implications. Namely, models that appear strong on common classes may still fail on rare, but important alternatives. Thus, any practical deployment should include explicit safeguards for low-confidence or out-of-distribution cases.

These results suggest that multimodal LLMs may be more useful as \emph{specialized assistants} than as fully general neuroimaging classifiers. A promising future direction is therefore not only model improvement, but also \emph{specialist-aware orchestration} - routing cases to models that show empirically stronger behavior for particular diagnostic patterns, while escalating uncertain, rare, or diagnostically diffuse cases to expert review. Such a design would better align deployment with the actual strengths and weaknesses observed in this benchmark.

\subsubsection{Clinical Interpretation and Limitations}

The results obtained from this work indicate that current MLLMs should not be considered as general-purpose neuroimaging diagnosticians. Instead, their potential clinical value lies in narrowly defined assistive roles, where model strengths align with specific tasks such as tumor triage. However, it is important to emphasize that even the strongest models, exhibit clinically relevant failure modes, particularly for subtle or rare conditions.

This study has several limitations. First, the benchmark is based on 2D neuroimaging slices rather than full volumetric data. Many neurological conditions, especially multiple sclerosis and subtle stroke, require 3D spatial context and longitudinal comparison for reliable interpretation. As a result, the reported performance may overestimate real-world capabilities.

Second, the evaluation focuses on image-only inputs, without access to structured clinical metadata such as patient history, symptoms, laboratory results, or previous imaging. In clinical practice, such information is critical for diagnostic decision-making and uncertainty resolution. The absence of this context limits direct clinical applicability.

Third, although uncertainty handling is evaluated through abstention behavior and calibration metrics, these measures primarily reflect model-internal uncertainty under fixed and complete inputs. In clinical practice, uncertainty also arises from ambiguous findings, missing imaging sequences, incomplete clinical context, or the need for longitudinal and/or multi-slice reasoning, which cannot be fully captured by single-image prompts. Taking this into account, model confidence estimates and calibration metrics should not be interpreted as direct substitutes for clinical judgment or as a guaranty of decision safety.

Fourth, the benchmark evaluates general-purpose multimodal models without domain-specific fine-tuning. While this reflects realistic zero-shot and few-shot usage, it does not capture the full potential of models adapted to neurological imaging through supervised or self-supervised training.

\subsubsection{Implications for Future Research}

Several directions emerge for future research. A key direction is the evaluation of volumetric (3D) neuroimaging data, including multi-slice reasoning and spatial consistency across entire scans. This is essential to assess clinical readiness for neurological imaging.

Future benchmarks should also integrate structured clinical metadata, enabling evaluation of multimodal reasoning that more closely reflects real diagnostic workflows. Another important direction is domain-specific adaptation, including fine-tuning or instruction tuning on curated neuroimaging datasets, followed by controlled re-evaluation of calibration and safety.

Longitudinal evaluation across multi-timepoint imaging studies would allow assessment of disease progression and treatment response, which are central to many neurological conditions.

Finally, future work should explore interaction between humans and AI, including human-in-the-loop evaluation, error analysis by specialists, and usability studies to determine how such systems may safely assist, rather than replace, clinical decision-making.


\section{Conclusions}

We presented NeuroVLM-Bench, a comprehensive and clinically grounded benchmark for evaluating multimodal large language models in neuroimaging across multiple sclerosis, stroke, and brain tumors using diverse 2D MRI and CT data. Through a rigorous, multi-phase evaluation of 20 frontier models, performance was assessed along four complementary evaluation directions: discriminative classification performance, calibration, structured-output reliability, and computational efficiency. Under a structured unified prompting protocol, the benchmark evaluates multiple structured output fields simultaneously, including diagnosis, diagnosis subtype, imaging modality, specialized MRI sequence, and anatomical plane. The results show that technical imaging attributes such as modality identification, anatomical plane recognition, and to a large extent specialized sequence classification are nearly solved by most models, whereas clinically meaningful reasoning tasks remain substantially more challenging. In particular, diagnostic reasoning—and especially fine-grained diagnosis subtype prediction—continues to represent the main limitation of current multimodal systems. Consistent with this pattern, tumor classification is the most reliable diagnostic category, stroke remains moderately solvable, and multiple sclerosis and rare abnormalities remain challenging. Because the evaluation splits were constructed in a stratified manner, the final Phase~3 results confirm the patterns observed in the earlier benchmark phases.

Among the evaluated systems, Gemini~2.5~Pro and GPT-5~Chat achieved the strongest overall diagnostic performance, while Gemini~2.5~Flash offered the most favorable balance between diagnostic accuracy, calibration, and computational efficiency. Few-shot prompting improved performance for many models but also increased token usage, latency, and inference cost. These results highlight that performance gains must be considered together with operational scalability when evaluating models for real-world clinical deployment.

Our findings emphasize both the promise and the current limitations of multimodal LLMs for neuroimaging. Diagnostic accuracy, reliability, calibration, and computational efficiency are not covered under a single model, underscoring the importance of multi-dimensional evaluation when developing clinically usable systems. At the same time, the progress observed in both proprietary and domain-specialized open-weight models suggests that continued architectural improvements and targeted domain adaptation may further narrow current performance gaps.

Within neuroimaging, this benchmarking provides value on several levels. For healthcare institutions, it offers a systematic framework for assessing how reliably vision-enabled LLMs can assist diagnostic workflows and analyze large repositories of imaging data already present in clinical archives and the scientific literature. Such large-scale automated classification could accelerate retrospective studies, support secondary analyses, and enable more efficient use of historical datasets. For developers of medical AI systems, the benchmark reveals concrete performance gaps across neurological conditions and imaging tasks, guiding model refinement and domain adaptation. For the research community, it establishes a reproducible and transparent basis for comparing emerging multimodal models and promoting cumulative scientific progress. The benchmark also opens opportunities in medical education by enabling structured categorization of imaging cases for training purposes. Finally, policymakers and regulatory bodies may benefit from systematic benchmarking evidence when defining guidelines for the safe integration of AI into clinical practice. Together, these contributions extend beyond technical evaluation, supporting responsible development, deployment, and governance of multimodal AI systems in neuroimaging.

\clearpage


\bmhead{Ethics and Consent to Participate declarations} 
Not applicable.

\bmhead{Competing Interests} 
The authors declare no competing interests.

\bmhead{Acknowledgment} 
This research was funded by the European Union under Horizon Europe project ChatMED grant agreement ID: 101159214.

\bmhead{Supplementary Information}
Supplementary material for this article is available alongside the submitted manuscript.
\clearpage

\bookmarksetup{startatroot}
\appendix

\renewcommand{\thesection}{\Alph{section}}

\renewcommand{\thesubsection}{\thesection.\arabic{subsection}}

\renewcommand{\thesubsubsection}{\thesubsection.\arabic{subsubsection}}

\clearpage

\bibliography{references}

@article{knowles2024comparing,
  title={Comparing the pathology, clinical, and demographic characteristics of younger and older-onset multiple sclerosis},
  author={Knowles, Sarah and Middleton, Rod and Cooze, Benjamin and Farkas, Ildiko and Leung, Yeung Yeung and Allen, Kelsey and Winslade, Molly and Owen, David RJ and Magliozzi, Roberta and Reynolds, Richard and others},
  journal={Annals of Neurology},
  volume={95},
  number={3},
  pages={471--486},
  year={2024},
  publisher={Wiley Online Library}
}

@article{louis20212021,
  title={The 2021 WHO classification of tumors of the central nervous system: a summary},
  author={Louis, David N and Perry, Arie and Wesseling, Pieter and Brat, Daniel J and Cree, Ian A and Figarella-Branger, Dominique and Hawkins, Cynthia and Ng, HK and Pfister, Stefan M and Reifenberger, Guido and others},
  journal={Neuro-oncology},
  volume={23},
  number={8},
  pages={1231--1251},
  year={2021},
  publisher={Oxford University Press US}
}

@article{gbd2021global,
  title={Global, regional, and national burden of stroke and its risk factors, 1990--2019: a systematic analysis for the Global Burden of Disease Study 2019},
  author={GBD 2019 Stroke Collaborators and others},
  journal={The Lancet. Neurology},
  volume={20},
  number={10},
  pages={795},
  year={2021}
}

@article{thompson2018diagnosis,
  title={Diagnosis of multiple sclerosis: 2017 revisions of the McDonald criteria},
  author={Thompson, Alan J and Banwell, Brenda L and Barkhof, Frederik and Carroll, William M and Coetzee, Timothy and Comi, Giancarlo and Correale, Jorge and Fazekas, Franz and Filippi, Massimo and Freedman, Mark S and others},
  journal={The Lancet Neurology},
  volume={17},
  number={2},
  pages={162--173},
  year={2018},
  publisher={Elsevier}
}

@article{powers2019update,
  title={Update to the 2018 guidelines for the early management of acute ischemic stroke: a guideline for healthcare professionals from the American Heart Association/American Stroke Association},
  author={Powers, WJ and Rabinstein, AA and Ackerson, T and Adeoye, OM and Bambakidis, NC and Becker, K and Biller, J and Brown, M and Demaerschalk, BM and Hoh, B and others},
  journal={Stroke},
  volume={50},
  number={12},
  pages={e344--e418},
  year={2019}
}

@article{isensee2021nnu,
  title={nnU-Net: a self-configuring method for deep learning-based biomedical image segmentation},
  author={Isensee, Fabian and Jaeger, Paul F and Kohl, Simon AA and Petersen, Jens and Maier-Hein, Klaus H},
  journal={Nature methods},
  volume={18},
  number={2},
  pages={203--211},
  year={2021},
  publisher={Nature Publishing Group}
}

@article{baid2021rsna,
  title={The rsna-asnr-miccai brats 2021 benchmark on brain tumor segmentation and radiogenomic classification},
  author={Baid, Ujjwal and Ghodasara, Satyam and Mohan, Suyash and Bilello, Michel and Calabrese, Evan and Colak, Errol and Farahani, Keyvan and Kalpathy-Cramer, Jayashree and Kitamura, Felipe C and Pati, Sarthak and others},
  journal={arXiv preprint arXiv:2107.02314},
  year={2021}
}

@article{hernandez2022isles,
  title={ISLES 2022: A multi-center magnetic resonance imaging stroke lesion segmentation dataset},
  author={Hernandez Petzsche, Moritz R and De La Rosa, Ezequiel and Hanning, Uta and Wiest, Roland and Valenzuela, Waldo and Reyes, Mauricio and Meyer, Maria and Liew, Sook-Lei and Kofler, Florian and Ezhov, Ivan and others},
  journal={Scientific data},
  volume={9},
  number={1},
  pages={762},
  year={2022},
  publisher={Nature Publishing Group UK London}
}

@article{rocca2024current,
  title={Current and future role of MRI in the diagnosis and prognosis of multiple sclerosis},
  author={Rocca, Maria A and Preziosa, Paolo and Barkhof, Frederik and Brownlee, Wallace and Calabrese, Massimiliano and De Stefano, Nicola and Granziera, Cristina and Ropele, Stefan and Toosy, Ahmed T and Vidal-Jordana, {\`A}ngela and others},
  journal={The Lancet Regional Health--Europe},
  volume={44},
  year={2024},
  publisher={Elsevier}
}

@article{martucci2023magnetic,
  title={Magnetic resonance imaging of primary adult brain tumors: state of the art and future perspectives},
  author={Martucci, Matia and Russo, Rosellina and Schimperna, Francesco and D’Apolito, Gabriella and Panfili, Marco and Grimaldi, Alessandro and Perna, Alessandro and Ferranti, Andrea Maurizio and Varcasia, Giuseppe and Giordano, Carolina and others},
  journal={Biomedicines},
  volume={11},
  number={2},
  pages={364},
  year={2023},
  publisher={MDPI}
}

@techreport{christensen2021opportunities,
  title={Opportunities and obstacles for deep learning in biology and medicine [update in progress]},
  author={Christensen, Brock C and Gitter, Anthony and Himmelstein, Daniel S and Titus, Alexander J and Levy, Joshua J and Greene, Casey S and Elton, Daniel C},
  year={2021},
  institution={Manubot}
}

@article{huang2020fusion,
  title={Fusion of medical imaging and electronic health records using deep learning: a systematic review and implementation guidelines},
  author={Huang, Shih-Cheng and Pareek, Anuj and Seyyedi, Saeed and Banerjee, Imon and Lungren, Matthew P},
  journal={NPJ digital medicine},
  volume={3},
  number={1},
  pages={136},
  year={2020},
  publisher={Nature Publishing Group UK London}
}

@article{kline2022multimodal,
  title={Multimodal machine learning in precision health: A scoping review},
  author={Kline, Adrienne and Wang, Hanyin and Li, Yikuan and Dennis, Saya and Hutch, Meghan and Xu, Zhenxing and Wang, Fei and Cheng, Feixiong and Luo, Yuan},
  journal={NPJ digital medicine},
  volume={5},
  number={1},
  pages={171},
  year={2022},
  publisher={Nature Publishing Group UK London}
}

@article{kelly2019key,
  title={Key challenges for delivering clinical impact with artificial intelligence},
  author={Kelly, Christopher J and Karthikesalingam, Alan and Suleyman, Mustafa and Corrado, Greg and King, Dominic},
  journal={BMC medicine},
  volume={17},
  number={1},
  pages={195},
  year={2019},
  publisher={Springer}
}

@article{alayrac2022flamingo,
  title={Flamingo: a visual language model for few-shot learning},
  author={Alayrac, Jean-Baptiste and Donahue, Jeff and Luc, Pauline and Miech, Antoine and Barr, Iain and Hasson, Yana and Lenc, Karel and Mensch, Arthur and Millican, Katherine and Reynolds, Malcolm and others},
  journal={Advances in neural information processing systems},
  volume={35},
  pages={23716--23736},
  year={2022}
}

@article{wang2025capabilities,
  title={Capabilities of gpt-5 on multimodal medical reasoning},
  author={Wang, Shansong and Hu, Mingzhe and Li, Qiang and Safari, Mojtaba and Yang, Xiaofeng},
  journal={arXiv preprint arXiv:2508.08224},
  year={2025}
}

@article{hu2025benchmarking,
  title={Benchmarking GPT-5 for Zero-Shot Multimodal Medical Reasoning in Radiology and Radiation Oncology},
  author={Hu, Mingzhe and Eidex, Zach and Wang, Shansong and Safari, Mojtaba and Li, Qiang and Yang, Xiaofeng},
  journal={arXiv preprint arXiv:2508.13192},
  year={2025}
}

@article{achiam2023gpt,
  title={Gpt-4 technical report},
  author={Achiam, Josh and Adler, Steven and Agarwal, Sandhini and Ahmad, Lama and Akkaya, Ilge and Aleman, Florencia Leoni and Almeida, Diogo and Altenschmidt, Janko and Altman, Sam and Anadkat, Shyamal and others},
  journal={arXiv preprint arXiv:2303.08774},
  year={2023}
}

@article{li2023llava,
  title={Llava-med: Training a large language-and-vision assistant for biomedicine in one day},
  author={Li, Chunyuan and Wong, Cliff and Zhang, Sheng and Usuyama, Naoto and Liu, Haotian and Yang, Jianwei and Naumann, Tristan and Poon, Hoifung and Gao, Jianfeng},
  journal={Advances in Neural Information Processing Systems},
  volume={36},
  pages={28541--28564},
  year={2023}
}

@article{zhang2023large,
  title={Large-scale domain-specific pretraining for biomedical vision-language processing},
  author={Zhang, Sheng and Xu, Yanbo and Usuyama, Naoto and Bagga, Jaspreet and Tinn, Robert and Preston, Sam and Rao, Rajesh and Wei, Mu and Valluri, Naveen and Wong, Cliff and others},
  journal={arXiv preprint arXiv:2303.00915},
  volume={2},
  number={3},
  pages={6},
  year={2023}
}

@article{tanno2025collaboration,
  title={Collaboration between clinicians and vision--language models in radiology report generation},
  author={Tanno, Ryutaro and Barrett, David GT and Sellergren, Andrew and Ghaisas, Sumedh and Dathathri, Sumanth and See, Abigail and Welbl, Johannes and Lau, Charles and Tu, Tao and Azizi, Shekoofeh and others},
  journal={Nature Medicine},
  volume={31},
  number={2},
  pages={599--608},
  year={2025},
  publisher={Nature Publishing Group US New York}
}

@article{wang2025zero,
  title={Zero-Shot Multi-modal Large Language Model vs Supervised Deep Learning: A Comparative Study on CT-Based Intracranial Hemorrhage Subtyping},
  author={Wang, Yinuo and Zeng, Yue and Chen, Kai and Meng, Cai and Pan, Chao and Tang, Zhouping},
  journal={arXiv preprint arXiv:2505.09252},
  year={2025}
}

@article{cheng2025understanding,
  title={Understanding the robustness of vision-language models to medical image artefacts},
  author={Cheng, Zijie and Ong, Ariel Yuhan and Wagner, Siegfried K and Merle, David A and Ju, Lie and Li, Boxuan and He, Tiantian and Ran, An Ran and Jiang, Hongyang and Yang, Dawei and others},
  journal={medRxiv},
  pages={2025--05},
  year={2025},
  publisher={Cold Spring Harbor Laboratory Press}
}

@article{nan2025beyond,
  title={Beyond the Hype: A Dispassionate Look at Vision--Language Models in Medical Scenario},
  author={Nan, Yang and Zhou, Huichi and Xing, Xiaodan and Yang, Guang},
  journal={IEEE Transactions on Neural Networks and Learning Systems},
  year={2025},
  publisher={IEEE}
}

@article{zhou2025drvd,
  title={DrVD-Bench: Do Vision-Language Models Reason Like Human Doctors in Medical Image Diagnosis?},
  author={Zhou, Tianhong and Xu, Yin and Zhu, Yingtao and Xiao, Chuxi and Bian, Haiyang and Wei, Lei and Zhang, Xuegong},
  journal={arXiv preprint arXiv:2505.24173},
  year={2025}
}

@inproceedings{hu2024omnimedvqa,
  title={Omnimedvqa: A new large-scale comprehensive evaluation benchmark for medical lvlm},
  author={Hu, Yutao and Li, Tianbin and Lu, Quanfeng and Shao, Wenqi and He, Junjun and Qiao, Yu and Luo, Ping},
  booktitle={Proceedings of the IEEE/CVF Conference on Computer Vision and Pattern Recognition},
  pages={22170--22183},
  year={2024}
}

@article{ye2024gmai,
  title={Gmai-mmbench: A comprehensive multimodal evaluation benchmark towards general medical ai},
  author={Ye, Jin and Wang, Guoan and Li, Yanjun and Deng, Zhongying and Li, Wei and Li, Tianbin and Duan, Haodong and Huang, Ziyan and Su, Yanzhou and Wang, Benyou and others},
  journal={Advances in Neural Information Processing Systems},
  volume={37},
  pages={94327--94427},
  year={2024}
}

@article{das2025trustworthy,
  title={Trustworthy Medical Imaging with Large Language Models: A Study of Hallucinations Across Modalities},
  author={Das, Anindya Bijoy and Sakib, Shahnewaz Karim and Ahmed, Shibbir},
  journal={arXiv preprint arXiv:2508.07031},
  year={2025}
}

@article{sozer2025llms,
  title={Do LLMs Have ‘the Eye’for MRI? Evaluating GPT-4o, Grok, and Gemini on Brain MRI Performance: First Evaluation of Grok in Medical Imaging and a Comparative Analysis},
  author={Sozer, Alperen and Sahin, Mustafa Caglar and Sozer, Batuhan and Erol, Gokberk and Tufek, Ozan Yavuz and Nernekli, Kerem and Demirtas, Zuhal and Celtikci, Emrah},
  journal={Diagnostics},
  volume={15},
  number={11},
  pages={1320},
  year={2025},
  publisher={MDPI}
}

@article{toh2014differentiation,
  title={Differentiation of brain abscesses from glioblastomas and metastatic brain tumors: comparisons of diagnostic performance of dynamic susceptibility contrast-enhanced perfusion MR imaging before and after mathematic contrast leakage correction},
  author={Toh, Cheng Hong and Wei, Kuo-Chen and Chang, Chen-Nen and Ng, Shu-Hang and Wong, Ho-Fai and Lin, Ching-Po},
  journal={PLoS One},
  volume={9},
  number={10},
  pages={e109172},
  year={2014},
  publisher={Public Library of Science San Francisco, USA}
}

@article{toh2011differentiation,
  title={Differentiation of brain abscesses from necrotic glioblastomas and cystic metastatic brain tumors with diffusion tensor imaging},
  author={Toh, CH and Wei, K-C and Ng, S-H and Wan, Y-L and Lin, C-P and Castillo, Mauricio},
  journal={American journal of neuroradiology},
  volume={32},
  number={9},
  pages={1646--1651},
  year={2011},
  publisher={American Journal of Neuroradiology}
}

@article{cui2024diffusion,
  title={Diffusion-weighted imaging-based radiomics model using automatic machine learning to differentiate cerebral cystic metastases from brain abscesses},
  author={Cui, Linyang and Qin, Zheng and Sun, Siyuan and Feng, Weihua and Hou, Mingyuan and Yu, Dexin},
  journal={Journal of Cancer Research and Clinical Oncology},
  volume={150},
  number={3},
  pages={132},
  year={2024},
  publisher={Springer}
}

@misc{gaillard2025tumefactive,
  author       = {Gaillard, Frank and Sharma, R and Hooi Hooi, T and others},
  title        = {Tumefactive demyelinating lesion},
  howpublished = {Reference article, Radiopaedia.org},
  year         = {2025},
  note         = {Accessed: 27 Aug 2025},
  doi          = {10.53347/rID-2225},
  url          = {https://radiopaedia.org/articles/2225}
}

@article{cheng2017brain,
  title={Brain tumor dataset},
  author={Cheng, Jun},
  journal={Figshare},
  year={2017},
  doi={10.6084/m9.figshare.1512427},
  url={https://figshare.com/articles/dataset/brain_tumor_dataset/1512427}
}

@misc{fernando2022brain44,
  title={Brain Tumor MRI Images 44 Classes},
  author={Fernando, Feltrin},
  howpublished={Kaggle},
  year={2022},
  url={https://www.kaggle.com/datasets/fernando2rad/brain-tumor-mri-images-44c}
}

@misc{fernando2022brain17,
  title={Brain Tumor MRI Images 17 Classes},
  author={Fernando, Feltrin},
  howpublished={Kaggle},
  year={2022},
  url={https://www.kaggle.com/datasets/fernando2rad/brain-tumor-mri-images-17-classes}
}

@misc{hamada2020br35h,
  author       = {Ahmed Hamada},
  title        = {Br35H: Brain Tumor Detection (Kaggle Dataset)},
  howpublished = {\url{https://www.kaggle.com/datasets/ahmedhamada0/brain-tumor-detection/data}},
  year         = {2020},
  note         = {Accessed: 2025-08-27}
}

@misc{buraktaci2022ms,
  title={Multiple Sclerosis MRI Dataset},
  author={Burak Taci},
  howpublished={Kaggle},
  year={2022},
  url={https://www.kaggle.com/datasets/buraktaci/multiple-sclerosis}
}

@misc{liang2023aisd,
  title={AISD: Acute Ischemic Stroke Dataset},
  author={Liang, Griffin and others},
  howpublished={GitHub},
  year={2023},
  url={https://github.com/GriffinLiang/AISD}
}

@misc{ozgur2022stroke,
  title={Brain Stroke CT Dataset},
  author={Ozgur, Aslank},
  howpublished={Kaggle},
  year={2022},
  url={https://www.kaggle.com/datasets/ozguraslank/brain-stroke-ct-dataset}
}

@article{filippi2018lancet,
  title={Multiple sclerosis},
  author={Filippi, Massimo and Bar-Or, Amit and Piehl, Fredrik and Preziosa, Paolo and Solari, Alessandra and Vukusic, Sandra and Rocca, Maria A},
  journal={The Lancet},
  volume={391},
  number={10130},
  pages={1622--1636},
  year={2018},
  doi={10.1016/S0140-6736(18)30481-1}
}

@article{jauch2013stroke,
  title={Guidelines for the early management of patients with acute ischemic stroke: 2013 update},
  author={Jauch, Edward C and Saver, Jeffrey L and Adams, Harold P and Bruno, Antony and Connors, J Joseph and Demaerschalk, Bart M and others},
  journal={Stroke},
  volume={44},
  number={3},
  pages={870--947},
  year={2013},
  doi={10.1161/STR.0b013e318284056a}
}

@article{ferber2024context,
  title={In-context learning enables multimodal large language models to classify cancer pathology images},
  author={Ferber, Dyke and W{\"o}lflein, Georg and Wiest, Isabella C and Ligero, Marta and Sainath, Srividhya and Ghaffari Laleh, Narmin and El Nahhas, Omar SM and M{\"u}ller-Franzes, Gustav and J{\"a}ger, Dirk and Truhn, Daniel and others},
  journal={Nature Communications},
  volume={15},
  number={1},
  pages={10104},
  year={2024},
  publisher={Nature Publishing Group UK London}
}

@inproceedings{shakeri2024few,
  title={Few-shot adaptation of medical vision-language models},
  author={Shakeri, Fereshteh and Huang, Yunshi and Silva-Rodr{\'\i}guez, Julio and Bahig, Houda and Tang, An and Dolz, Jose and Ben Ayed, Ismail},
  booktitle={International Conference on Medical Image Computing and Computer-Assisted Intervention},
  pages={553--563},
  year={2024},
  organization={Springer}
}

@article{wang2023real,
  title={A real-world dataset and benchmark for foundation model adaptation in medical image classification},
  author={Wang, Dequan and Wang, Xiaosong and Wang, Lilong and Li, Mengzhang and Da, Qian and Liu, Xiaoqiang and Gao, Xiangyu and Shen, Jun and He, Junjun and Shen, Tian and others},
  journal={Scientific Data},
  volume={10},
  number={1},
  pages={574},
  year={2023},
  publisher={Nature Publishing Group UK London}
}

@article{bedi2505medhelm,
  title={MedHELM: Holistic Evaluation of Large Language Models for Medical Tasks. Published online June 2, 2025. doi: 10.48550},
  author={Bedi, S and Cui, H and Fuentes, M and others},
  journal={arXiv preprint arXiv.2505.23802},
  year={2025}
}

@article{pal2023med,
  title={Med-HALT: medical domain hallucination test for large language models. arXiv},
  author={Pal, A and Umapathi, LK and Sankarasubbu, M},
  journal={arXiv preprint arXiv:2307.15343},
  year={2023}
}

@article{wen2024art,
  title={The art of refusal: A survey of abstention in large language models},
  author={Wen, Bingbing and Yao, Jihan and Feng, Shangbin and Xu, Chenjun and Tsvetkov, Yulia and Howe, Bill and Wang, Lucy Lu},
  journal={arXiv e-prints},
  pages={arXiv--2407},
  year={2024}
}

@article{jing2025beyond,
  title={Beyond Classification Accuracy: Neural-MedBench and the Need for Deeper Reasoning Benchmarks},
  author={Jing, Miao and Jia, Mengting and Lin, Junling and Shen, Zhongxia and Wang, Lijun and Peng, Yuanyuan and Gao, Huan and Xu, Mingkun and Li, Shangyang},
  journal={arXiv preprint arXiv:2509.22258},
  year={2025}
}

@article{aali2025structured,
  title={Structured Prompting Enables More Robust, Holistic Evaluation of Language Models},
  author={Aali, Asad and Mohsin, Muhammad Ahmed and Bikia, Vasiliki and Singhvi, Arnav and Gaus, Richard and Bedi, Suhana and Cui, Hejie and Fuentes, Miguel and Unell, Alyssa and Mai, Yifan and others},
  journal={arXiv preprint arXiv:2511.20836},
  year={2025}
}

@article{fraile2025measuring,
  title={On Measuring Large Language Models Performance with Inferential Statistics},
  author={Fraile-Hern{\'a}ndez, Jes{\'u}s M and Pe{\~n}as, Anselmo},
  journal={Information},
  volume={16},
  number={9},
  pages={817},
  year={2025},
  publisher={MDPI}
}

@misc{peng2025omnibrainbenchcomprehensivemultimodalbenchmark,
      title={OmniBrainBench: A Comprehensive Multimodal Benchmark for Brain Imaging Analysis Across Multi-stage Clinical Tasks}, 
      author={Zhihao Peng and Cheng Wang and Shengyuan Liu and Zhiying Liang and Zanting Ye and Minjie Ju and PeterYM Woo and Yixuan Yuan},
      year={2025},
      eprint={2511.00846},
      archivePrefix={arXiv},
      primaryClass={cs.CV},
      url={https://arxiv.org/abs/2511.00846}, 
}

@inproceedings{singh2025crossmed,
  title={CrossMed: A Multimodal Cross-Task Benchmark for Compositional Generalization in Medical Imaging},
  author={Singh, Pooja and Ujjain, Siddhant and Gandhi, Tapan Kumar and Kumar, Sandeep},
  booktitle={2025 IEEE EMBS International Conference on Biomedical and Health Informatics (BHI)},
  pages={1--7},
  year={2025},
  organization={IEEE}
}

@inproceedings{ruan2025comprehensive,
  title={Comprehensive evaluation of multimodal ai models in medical imaging diagnosis: From data augmentation to preference-based comparison},
  author={Ruan, Cailian and Huang, Chengyue and Yang, Yahe},
  booktitle={2025 13th International Conference on Bioinformatics and Computational Biology (ICBCB)},
  pages={58--64},
  year={2025},
  organization={IEEE}
}

@article{nam2025multimodal,
  title={Multimodal large language models in medical imaging: current state and future directions},
  author={Nam, Yoojin and Kim, Dong Yeong and Kyung, Sunggu and Seo, Jinyoung and Song, Jeong Min and Kwon, Jimin and Kim, Jihyun and Jo, Wooyoung and Park, Hyungbin and Sung, Jimin and others},
  journal={Korean Journal of Radiology},
  volume={26},
  number={10},
  pages={900},
  year={2025}
}

@inproceedings{pan2025medvlm,
  title={Medvlm-r1: Incentivizing medical reasoning capability of vision-language models (vlms) via reinforcement learning},
  author={Pan, Jiazhen and Liu, Che and Wu, Junde and Liu, Fenglin and Zhu, Jiayuan and Li, Hongwei Bran and Chen, Chen and Ouyang, Cheng and Rueckert, Daniel},
  booktitle={International Conference on Medical Image Computing and Computer-Assisted Intervention},
  pages={337--347},
  year={2025},
  organization={Springer}
}

@inproceedings{shi2025medm,
  title={Medm-vl: What makes a good medical lvlm?},
  author={Shi, Yiming and Yang, Shaoshuai and Zhu, Xun and Wang, Haoyu and Fu, Xiangling and Li, Miao and Wu, Ji},
  booktitle={International Workshop on Agentic AI for Medicine},
  pages={290--299},
  year={2025},
  organization={Springer}
}

@inproceedings{nguyen2025localizing,
  title={Localizing Before Answering: A Benchmark for Grounded Medical Visual Question Answering},
  author={Nguyen, Dung and Ho, Minh Khoi and Ta, Huy and Nguyen, Thanh Tam and Chen, Qi and Rav, Kumar and Dang, Quy Duong and Ramchandre, Satwik and Phung, Son Lam and Liao, Zhibin and others},
  booktitle={Proceedings of the Thirty-Fourth International Joint Conference on Artificial Intelligence},
  pages={7670--7678},
  year={2025}
}

@article{yue2025medsg,
  title={MedSG-Bench: A Benchmark for Medical Image Sequences Grounding},
  author={Yue, Jingkun and Zhang, Siqi and Jia, Zinan and Xu, Huihuan and Han, Zongbo and Liu, Xiaohong and Wang, Guangyu},
  journal={arXiv preprint arXiv:2505.11852},
  year={2025}
}

@article{mei2022radimagenet,
  title={RadImageNet: an open radiologic deep learning research dataset for effective transfer learning},
  author={Mei, Xueyan and Liu, Zelong and Robson, Philip M and Marinelli, Brett and Huang, Mingqian and Doshi, Amish and Jacobi, Adam and Cao, Chendi and Link, Katherine E and Yang, Thomas and others},
  journal={Radiology: Artificial Intelligence},
  volume={4},
  number={5},
  pages={e210315},
  year={2022},
  publisher={Radiological Society of North America}
}

@article{tak2024brainiac,
    title={BrainIAC: A Foundation Model for Generalized Brain MRI Analysis},
    author={Tak, Divyanshu and others},
    journal={medRxiv},
    year={2024},
    doi={10.1101/2024.12.02.24317992}
}

@article{cox2024brainsegfounder,
  title={BrainSegFounder: Towards 3D foundation models for neuroimage segmentation},
  author={Cox, Joseph and Liu, Peng and Stolte, Skylar E and Yang, Yunchao and Liu, Kang and See, Kyle B and Ju, Huiwen and Fang, Ruogu},
  journal={Medical Image Analysis},
  volume={97},
  pages={103301},
  year={2024},
  publisher={Elsevier}
}

@misc{zhengzhengbraingpt,
  title        = {BrainGPT: A Brain-Inspired SNN-Based Large Language Model},
  author       = {Tang, Zhengzheng and Zhu, Eva},
  year         = {2025},
  howpublished = {\url{https://github.com/braingpt-lovelab}},
  note         = {Github repository}
}

@article{qiu2024multimodal_robustness,
  title={Benchmarking Robustness of Multimodal Image-Text Models under Distribution Shift},
  author={Qiu, Jielin and Zhu, Yi and Shi, Xingjian and others},
  journal={Journal of Data-centric Machine Learning Research},
  year={2024}
}

@article{oh2025understanding,
  title={Understanding multimodal llms under distribution shifts: An information-theoretic approach},
  author={Oh, Changdae and Fang, Zhen and Im, Shawn and Du, Xuefeng and Li, Yixuan},
  journal={arXiv preprint arXiv:2502.00577},
  year={2025}
}

@inproceedings{imam2025robustness,
  title={On the robustness of medical vision-language models: Are they truly generalizable?},
  author={Imam, Raza and Marew, Rufael and Yaqub, Mohammad},
  booktitle={Annual Conference on Medical Image Understanding and Analysis},
  pages={233--256},
  year={2025},
  organization={Springer}
}

@article{macin2022accurate,
  title={An accurate multiple sclerosis detection model based on exemplar multiple parameters local phase quantization: ExMPLPQ},
  author={Macin, Gulay and Tasci, Burak and Tasci, Irem and Faust, Oliver and Barua, Prabal Datta and Dogan, Sengul and Tuncer, Turker and Tan, Ru-San and Acharya, U Rajendra},
  journal={Applied Sciences},
  volume={12},
  number={10},
  pages={4920},
  year={2022},
  publisher={Mdpi}
}

\end{document}